\documentclass[10pt,compsoc]{IEEEtran}

\makeatletter \def\mdseries@tt{m} \makeatother 

\usepackage{wrapfig}
\usepackage{graphicx}
\usepackage{indentfirst}
\usepackage{booktabs} 
\usepackage{graphicx}
\graphicspath{./figures}
\usepackage{colortbl}
\usepackage{float}
\usepackage{amsmath}
\usepackage{makecell}

\usepackage{rotating}

\newcommand{\Rev}[1]{\textcolor{black}{{#1}}}
\newcommand{\Revv}[1]{\textcolor{black}{{#1}}}

\usepackage[backgroundcolor=white,bordercolor=black,linecolor=black]{todonotes}

\usepackage[singlespacing]{setspace}
\usepackage{threeparttable}
\usepackage{multirow}
\usepackage{subcaption}
\usepackage[utf8]{inputenc}
\usepackage[english]{babel}
\usepackage{amsthm}
\usepackage{changepage} 
\usepackage{mathtools}
\usepackage{textgreek}
\usepackage{tcolorbox}
\usepackage[compact]{titlesec}     
\usepackage{amsmath}
\usepackage{tikz}

\usepackage{float}
\usepackage{wrapfig}
\usetikzlibrary{shapes,positioning,arrows,calc}

\DeclareFontFamily{OT1}{pzc}{}
\DeclareFontShape{OT1}{pzc}{m}{it}{<-> s * [1.1] pzcmi7t}{}
\DeclareMathAlphabet{\mathpzc}{OT1}{pzc}{m}{it}
\theoremstyle{definition}
\DeclareMathAlphabet\mathbfcal{OMS}{cmsy}{b}{n}
\usepackage{centernot}
\usepackage{enumitem}
\makeatletter
\def\algbackskip{\hskip-\ALG@thistlm}

\newcommand{\ygg@basicalert}[2]{\fbox{\bfseries\sffamily\scriptsize#1}{\sf\small$\blacktriangleright$\textit{#2}$\blacktriangleleft$}}
\newcommand{\manel}[1]{\ygg@basicalert{Manel}{#1}}

\usepackage{xcolor}
\usepackage[ruled, vlined, linesnumbered]{algorithm2e}
\newcommand{\nonl}{\renewcommand{\nl}{\let\nl\oldnl}}% Remove line number for one line

\SetKwInput{KwInput}{Input}                % Set the Input
\SetKwInput{KwOutput}{Output} 

\SetAlFnt{\small\sffamily}

\SetCommentSty{mycommfont}

\usepackage{amsmath, listings, amsthm, amssymb, xspace}

\usepackage{hyperref}
\hypersetup{colorlinks,allcolors=black}

\theoremstyle{remark}

\author{Zohreh Aghababaeyan, Manel Abdellatif, Lionel Briand, and Ramesh S

\IEEEcompsocitemizethanks{\IEEEcompsocthanksitem Zohreh Aghababaeyan is with the School of EECS, University of Ottawa, Ottawa, Canada. \protect\\
E-mail: zagha052@uottawa.ca
\IEEEcompsocthanksitem Manel Abdellatif is with the Software and Information Technology Engineering Department, École de Technologie Supérieure, Montreal, Canada. \protect\\
E-mail: Manel.abdellatif@etsmtl.ca
\IEEEcompsocthanksitem Lionel Briand is with the School of Electrical Engineering and Computer Science (EECS), University of Ottawa, Ottawa, Canada, and also with the Lero SFI Research Center and University of Limerick, Ireland.\protect\\
E-mail: Lbriand@uottawa.ca
\IEEEcompsocthanksitem Ramesh S is with the Department of Research and Development, General Motors, Warren, MI, USA.\protect\\
E-mail: Ramesh.s@gm.com
}}

\begin{document}

\title{ DiffGAN: A Test Generation Approach for Differential Testing of Deep Neural Networks for Image Analysis}

\IEEEtitleabstractindextext{%

\begin{abstract}
 
Deep Neural Networks (DNNs) are increasingly deployed across a wide range of applications, from image classification to autonomous driving. However, ensuring their reliability remains a challenge, and in many situations, alternative models with similar functionality and accuracy levels are available. Traditional accuracy-based evaluations often fail to capture behavioral differences between such models, particularly when testing datasets are limited, making it challenging to select or optimally combine models.

Differential testing addresses this limitation by generating test inputs that expose discrepancies in the behavior of DNN models. However, existing differential testing approaches face significant limitations: many rely on access to model internals or are constrained by the availability of seed inputs, limiting their generalizability and effectiveness.
\Revv{In response to these challenges, we propose \textit{DiffGAN}, a black-box test generation approach for differential testing of DNN models. Our approach, though adaptable to other domains, is specific to DNN models for image classification tasks, a highly prevalent application area.}
%In response to these challenges, we propose \textit{DiffGAN}, a black-box test generation approach for differential testing of DNN models. 
Our method relies on a Generative Adversarial Network (GAN) and the Non-dominated Sorting Genetic Algorithm II (NSGA-II) to generate diverse and valid triggering inputs that effectively reveal behavioral discrepancies between models. 
Our method employs two custom fitness functions, one focused on diversity and the other on divergence, to guide the exploration of the GAN input space and identify discrepancies between the models' outputs.
By strategically searching the GAN input space, we show that \textit{DiffGAN} can effectively generate inputs with specific features that trigger differences in behavior for the models under test.
Unlike traditional white-box methods, \textit{DiffGAN} does not require access to the internal structure of the models, which makes it applicable to a wider range of situations. We evaluate \textit{DiffGAN} on a benchmark comprising eight pairs of DNN models trained on two widely used image classification datasets. Our results demonstrate that \textit{DiffGAN} significantly outperforms a state-of-the-art (SOTA) baseline, generating four times more triggering inputs, with higher diversity and validity, within the same testing budget. Furthermore, we show that the generated input can be used to improve the accuracy of a machine learning-based model selection mechanism, which dynamically selects the best-performing model based on input characteristics and can thus be used as a smart model output voting mechanism when using alternative models together.
% Model selection is essential, as models may perform differently under varying conditions. For instance, one model might be more accurate but expensive, while another is more efficient. Some models may excel in specific environments, such as rainy or snowy weather. Real-world testing to uncover these differences can be risky and time-consuming, making DiffGAN’s ability to generate such inputs critical for improving model selection and accuracy, especially in safety-critical applications like autonomous driving and healthcare.

\end{abstract}

%\keywords{Deep Neural Network, Differential Testing, Test Generation, GAN, NSGA-II.}

}

\maketitle 
\section{Introduction} 
\label{Sec:Introduction}
% Further justification for the use of GANs in enhancing DNN testing comes from their application in biomedical image analysis. Research has shown that GANs can significantly improve classification accuracy by producing images that closely mimic real ones (Research \cite{FRIDADAR2018321}). Through visual Turing tests, where experts attempted to distinguish between GAN-generated and real biomedical images, the accuracies achieved were only 62.5% for identifying real images as real and 58.6% for identifying GAN images as fake. These results underscore the remarkable ability of GANs to create highly realistic images that are indistinguishable from genuine ones, demonstrating their substantial potential in data augmentation and model performance enhancement across fields.

Deep Neural Networks (DNNs) have become essential and widely used across various application domains such as image classification, natural language processing, and autonomous driving~\cite{aghababaeyan2021black,alam2020survey,Tonella2023}. Despite their success, ensuring the accuracy of these models remains a challenge, for example when alternative models with similar functionality and accuracy levels are available. Traditional accuracy-based evaluations of such models often fail to capture behavioral differences between them, particularly when testing datasets are limited, and thus selecting or optimally combining models is challenging.

These challenges underscore the need for differential testing, an approach that focuses on generating triggering inputs---test inputs specifically designed to expose behavioral discrepancies between DNN models under test.  
Differential testing becomes particularly crucial in scenarios such as model selection, model compression,  regression analysis, and the use of model ensembles. In the case of model selection, organizations often need to choose from several models that have been trained for the same task, possibly with different architectures, training data, or optimization techniques. 

Although these models may achieve similar accuracy on a given test dataset, their performance can vary significantly under different operational conditions. Differential testing helps identify these variations by generating inputs that expose where one model might outperform the others, providing deeper insights that go beyond simple accuracy metrics. Similarly, in model compression, large DNN models are often compressed using techniques such as quantization, pruning, or knowledge distillation to make them suitable for deployment in resource-constrained environments, such as mobile devices or embedded systems. Differential testing is essential here to ensure that the compressed models retain the accuracy of the original uncompressed versions. By generating inputs that highlight behavioral differences between the compressed and original models, differential testing allows developers to better understand the trade-offs involved in compression and to ensure that the model’s critical performance characteristics are preserved. For regression analysis, when DNN models are updated or retrained, differential testing identifies and generates inputs where the new version performs worse than the original or vice versa, ensuring that updates do not compromise the model's accuracy. Last, when using ensembles of models, a voting mechanism is required, and one needs to be able to learn when to trust which model's output. 

However, the differential testing process presents several challenges. First, the input subspace where highly accurate models mispredict is often extremely limited, within a huge input space. This makes it challenging to find the specific inputs that will trigger behavioral differences, especially in models designed to perform well across a wide range of scenarios. Efficient and effective techniques are thus required to reveal meaningful such discrepancies. Second, many models are treated as black boxes, where only the inputs and outputs are visible, without access to the internal workings of the model. This lack of access complicates the generation of test inputs since the internal decision-making processes remain hidden. Third, limited testing budgets pose another constraint, particularly in scenarios where collecting, labeling, and validating test inputs are time-consuming and resource-intensive. In cases where models are deployed in environments that require expensive simulators or manual validation, testing costs can quickly escalate, further limiting the resources available for comprehensive testing.

Finally, ensuring the validity of the generated inputs in differential testing poses a significant challenge. Invalid inputs can lead to mispredictions and skew models' performance assessments. Human feedback is often required to verify the contextual relevance and validity of test inputs, adding complexity and making the process more resource-intensive. This manual validation step is difficult to scale, further complicating efforts to evaluate true performance discrepancies between models effectively.

Several test generation approaches for the differential testing of DNN models have been proposed in the literature~\cite{pei2017deepxplore,xie2019deephunter,yahmed2022diverget,braiek2019deepevolution,xie2019diffchaser,tian2023finding,you2023regression}. However, these approaches often come with key limitations. Many of these methods are white-box~\cite{pei2017deepxplore,xie2019deephunter,yahmed2022diverget,braiek2019deepevolution}, requiring access to the internal structure and parameters of the DNN models under test. This restricts their applicability to proprietary models, where such internal access is typically unavailable due to privacy or intellectual property concerns. Some other approaches~\cite{yahmed2022diverget,braiek2019deepevolution} assume structural similarities between the models under test, which limits their applicability to DNN models of different architectures. Black-box differential testing methods~\cite{xie2019diffchaser,tian2023finding,you2023regression}, on the other hand, often rely on customized mutation operators applied to seed inputs to generate triggering inputs. However, these methods are highly dependent on the availability and quality of the seed data, restricting the generation process to variations of existing inputs.

In response to these challenges, we propose \textit{DiffGAN}, a black-box test generation approach tailored for the differential testing of DNN models with similar accuracy levels. \Revv{Our approach is specific to DNN models used in image classification, a domain in which such models are widely applied.} \textit{DiffGAN} leverages Generative Adversarial Networks (GANs) and the Non-dominated Sorting Genetic Algorithm II (NSGA-II) to generate diverse and valid triggering inputs that effectively reveal behavioral disagreements between the models under test.  
The use of a GAN model for this purpose stems from its proven ability to generate realistic and novel images that conform to the original input domain. This is achieved by exploring the latent space of possible inputs. To effectively guide the generation process, we rely on NSGA-II to effectively search the GAN’s latent space and generate diverse triggering images. The genetic search employs two fitness functions, focusing on maximizing (1) diversity to avoid the generation of redundant triggering images, and (2) divergence to maximize the behavioral disagreements between the DNN models. Since \textit{DiffGAN} operates as a black-box method, it does not require access to the internal structure of the models or assume structural similarities between them, making it suitable for testing any pair of DNN models. \textit{DiffGAN} is focused on image classification problems, for which Deep Learning is widely applied across many applications.

%In response to these challenges, we propose \textit{DiffGAN}, a black-box test generation approach tailored for the differential testing of DNN models with similar accuracy levels on a given test dataset. By leveraging Generative Adversarial Networks (GANs) and the Non-dominated Sorting Genetic Algorithm II (NSGA-II), \textit{DiffGAN} generates diverse and valid triggering inputs that effectively reveal behavioral disagreements between the models under test. Since \textit{DiffGAN} operates as a black-box method, it does not require access to the internal structure of the models or assume structural similarities between them, making it suitable for testing any pair of DNN models. \Rev{\textit{DiffGAN} is specific to image classification problems, for which Deep Learning is widely applied across many applications.}

To evaluate \textit{DiffGAN}, we constructed a benchmark consisting of eight pairs of models (16 DNN models) trained on two widely used image classification datasets. We compared its performance to \textit{DRfuzz}~\cite{you2023regression}, a SOTA differential testing baseline assuming the same testing budget. Our experimental results demonstrate that \textit{DiffGAN} significantly outperforms the baseline, generating on average four times more triggering inputs. Moreover, these inputs include a greater number of valid images and exhibit higher diversity, enabling a more comprehensive assessment of the models' behaviors. 
% To further demonstrate the utility of the triggering inputs generated by \textit{DiffGAN}, we trained a Random Forest (RF) model on these inputs to distinguish between two classifiers. The RF model predicts, based on the characteristics of the incoming input, which of the two classifiers (model 1 or model 2) will perform better, without the need to run both models in an online setting. By leveraging the triggering images generated by \textit{DiffGAN}, the RF model learns when each classifier is more suitable, effectively serving as a decision mechanism for selecting the optimal classifier for a given input. It is important to note that this use of RF is not intended as a novel ensemble method or voting mechanism. Instead, it serves as an example to demonstrate the value and efficiency of the \textit{DiffGAN}-generated inputs in highlighting differences between models and improving model selection processes.
We further demonstrate an application of the triggering inputs generated by \textit{DiffGAN} through the development of a machine learning-based selection mechanism. This system uses a machine learning model to dynamically select between two classifiers, predicting which one will perform better based on the input characteristics. Our results show that training the selection model with inputs generated by \textit{DiffGAN} leads to significantly more accurate selections compared to those trained with inputs from the baseline approach.

% We further provide an application scenario of the triggering inputs generated by \textit{DiffGAN} through the development of a machine learning-based voting mechanism. This voting system consists of two models responsible for classification, and a third model that dynamically selects which of the two classifiers to use based on the input characteristics.
% Our results show that training the selection model with inputs generated by \textit{DiffGAN} leads to a significantly more accurate voting mechanism compared to those trained with inputs from the baseline approach.

The main contributions of our paper are as follows: 

\begin{enumerate}

    \item We propose \textit{DiffGAN}, a novel, black-box test generation approach that leverages GANs and NSGA-II to identify triggering inputs for differential testing of DNN models, enabling an exhaustive analysis of behavioral disagreements between models with similar accuracy levels.
    %To the best of our knowledge, this is the first work addressing black-box differential testing for model selection.
    \item We perform a comprehensive evaluation on a benchmark of eight image classification model pairs, showing that \textit{DiffGAN} significantly outperforms the state-of-the-art baseline in generating more diverse and valid triggering inputs, given the same testing budget.
    \item We show that the triggering inputs generated by \textit{DiffGAN} can be used to train a machine learning-based mechanism to select the most accurate model based on input characteristics, leading to improved results when compared to models trained on data generated by the baseline. Model selection constitutes a very important application of differential testing. 
    \item To support future research and reproducibility, we have made our source code and benchmark publicly available\footnote{\url{https://github.com/zohreh-aaa/Diff_testing_GAN}}.

\end{enumerate}

The remainder of this paper is structured as follows. 
In Section~\ref{Sec:Prob}, we define the research problem and outline the key assumptions underlying our approach.
Section~\ref{Sec:Approach} details the methodology and design of our proposed solution.
In Section~\ref{Sec:Evaluation}, we present the empirical evaluation. 
In Section~\ref{Sec:Results}, we report our results.
Section~\ref{sec:Threats} discusses potential threats to the validity of our findings.
Section~\ref{Sec:RW} reviews related works.
Finally, Section~\ref{Sec:Conclusion} concludes our work.

\section{Problem Definition} \label{Sec:Prob}
%In this paper, we propose a test generation approach for the differential testing of DNN models to reveal behavioral disagreements between DNN models that exhibit similar accuracy levels. 

\Revv{In this paper, we propose a test generation approach for the differential testing of DNN models. The goal is to generate triggering inputs---inputs that expose behavioral discrepancies between DNN models with similar accuracy levels. Though it is adaptable to other domains, our approach focuses on image classification tasks, a prominent application domain for DNN models. Throughout the paper, we define a triggering input as an image that causes the DNN models under test to produce different outputs.}

\subsection{Motivations}

Differential testing is crucial across various practical applications, including model selection, voting systems, model compression, model regression, and detecting model reuse~\cite{xie2019diffchaser, tian2023finding, wang2022bet, you2023regression, li2021modeldiff}. 
For example, consider the context of model selection.
In practice, organizations often face the challenge of choosing the most suitable model during the training phase, prior to inference time. This decision process involves choosing from multiple DNN models, each one possibly trained by different providers for the same task, using different model architectures or training datasets, and demonstrating similar accuracy on a testing dataset. In such a scenario, the ability to generate inputs that clearly indicate differences in model behavior is invaluable to guide model selection.
 
Differential testing is also crucial for model output selection at inference time, in the context of ensemble learning or voting systems, where multiple DNN models are combined to obtain a better prediction than with individual models. This type of testing involves revealing triggering inputs that can not only help characterize the conditions under which one model may outperform the others but also enhance our understanding of each model's operational boundaries. By understanding the conditions under which one model fares better, one can determine at inference time which model's output to trust.

Differential testing is also crucial in the context of model compression, a process where large DNN models are transformed into more compact versions through techniques such as quantization, pruning, and knowledge distillation~\cite{9523139}. This transformation is aimed at deploying these models in resource-constrained environments while preserving as much of their original accuracy as possible. In these scenarios, differential testing is invaluable as it helps identify how the behavior of the compressed model deviates from the original. By generating triggering inputs that expose these differences, developers can gain a better understanding of the trade-offs involved in the compression process. 
%This is crucial for ensuring that the compressed models still meet the required accuracy levels before they are deployed in real-world applications.

Differential testing also plays a pivotal role in identifying model regression~\cite{you2023regression}, which entails detecting discrepancies between iterations of a deep learning model (original versus updated versions). 
For instance, an update aimed at enhancing the overall accuracy of a DNN model might inadvertently introduce inaccuracies under specific input conditions. To mitigate this, triggering test inputs are essential to systematically assess changes in accuracy across updates and thus ensure they do not lead to poorer accuracy in critical conditions. 
%This proactive strategy strengthens the model's reliability during modifications.

% Another application of Differential testing is on Model Regression, the process of identifying regression faults between versions of a deep learning model (for example original and updated version). Regression faults occur when updates to a model unintentionally degrade functionalities that were previously correct, potentially affecting user satisfaction and causing economic and reputational damage.\ref{}

% Take, for instance, an update to a DNN model—referred to here as Updated model—designed to improve overall accuracy of the original model. While the update enhances general performance, it introduces mispredictions in some scenarios previously handled correctly.

% In this context, triggering test inputs are essential. These inputs specifically challenge the updates in Updated model, ensuring that enhancements do not undermine the model’s accuracy in specific cases. By employing these inputs, developers can proactively identify and correct new vulnerabilities, thus maintaining the model's reliability post-update.

Additionally, differential testing helps detect model reuse~\cite{li2021modeldiff}. In industries where proprietary models are crucial assets, it is essential to verify that deployed models are original. Differential testing can identify if a model presented as new is actually derived from or modified from an existing model, thus protecting intellectual property and ensuring the authenticity of models.

%add sth here: we focus on model selection scenarios

In conclusion, while differential testing has broad applications, our focus in this work is specifically on model selection. Unlike other contexts, such as model compression, regression detection, or model reuse, the model selection scenario presents unique challenges. Here, the goal is not to evaluate how a single model evolves or changes, but rather to compare multiple models developed independently for the same task. These models may exhibit similar accuracy levels but differ in training data, structure, and optimization methods. Focusing on model selection allows us to explore how differential testing can expose behavioral differences between DNN models. This focus is critical because, in practice, selecting the most appropriate model, whether after training or at inference time, can directly impact the systems using these models. %By narrowing our scope to this application, we aim to provide deeper insights into model behavior, which can significantly enhance decision-making in real-world scenarios.

\subsection{Challenges}

Generating triggering inputs to reveal behavioral disagreements between DNN models under test introduces three main challenges. First, uncovering discrepancies in highly accurate models is inherently difficult due to the very small input subspace, leading to mispredictions. %their robustness against typical input variations and perturbations. 
Because they are designed to perform well under a wide range of conditions, we require effective and efficient techniques to identify behavioral disagreements. Second, in many real-world scenarios, we do not have access to the internal information of DNN models. Indeed, such models are often proprietary and deployed on servers, accessible only through cloud APIs. This limitation constrains us to using black-box techniques to generate triggering inputs, as we only have access to the inputs and outputs of the DNN models. Finally, the testing budget is often limited, particularly when models (1) require extensive manual labor to label and validate triggering inputs, or (2) operate on expensive simulators or physical devices. These simulators usually provide a controlled environment that replicates true operational scenarios, allowing for rigorous testing without the costs and risks associated with testing models deployed on expensive physical devices.  Within a realistic testing budget, we must effectively generate triggering inputs that reveal diverse behavioral disagreements between the models under test. 

%Together, these challenges demand a strategic approach to differential testing that balances efficacy in uncovering descrepencies in models behaviors with efficiency in resource use.

%Generating triggering inputs to reveal behavioral disagreements between DNN models under test presents three main challenges. First, Generating triggering inputs that effectively reveal discrepancies in highly-accurate models can be challenging since they, by design, are robust against typical variations and perturbations in input data. Second, in many practical contexts, access to the internals of the DNN models is restricted because these models are often proprietary. They may be deployed on servers and made available to customers exclusively through cloud APIs, limiting access to their internal layers. Third, in many cases we testing budget (input generation budget) is limited because of the models are deployed on  costly simulators and because of the high manual labeling cost of the triggering inputs to determine the accuracy levels of the models or understand the conditions under which one model fares better.

\subsection{Assumptions}

In this work, we propose a differential testing approach for DNN models considering the following assumptions: 
% BB
% :
% It is preferred to find triggering inputs quickly so that developers can obtain in-time feedback
% to assess and facilitate the entire dissemination workflow. However, this is a challenging task.
% Specifically, to accelerate the inference speed and reduce storage consumption, compressed models
% usually do not expose their architectures or intermediate computation results via APIs

\textbf{Black-Box Models:} We assume that our DNN models are black-box. The internal workings of the DNN, including intermediate computations and gradients, are inaccessible. This assumption is crucial, especially in scenarios involving models received from third-party providers or models trained via federated learning, where internal model details are naturally obscured to protect privacy and intellectual property. Moreover, frameworks such as TensorFlow Lite and ONNX Inference, which are commonly used for model quantization, typically offer only end-to-end inference APIs. These frameworks do not provide access to intermediate model data, emphasizing efficiency in inference~\cite{tensorflowliteguide}.
    
\textbf{Pairwise Differential Testing:} Our test generation approach primarily focuses on the pairwise comparison of DNN models. This assumption means that our analysis of behavioral differences is based on comparisons between two DNN models at a time, allowing for the generation of triggering inputs for a pair of DNN models.
    
\textbf{No Access to the Training Data:} We assume that we do not have access to the original training datasets of the models under test. This is a common scenario in many real-world situations where models are proprietary or when model testing is provided as a service, for example. Consequently, our ability to generate inputs and interpret model behavior is based solely on observations from models' outputs and the currently available testing data, rather than insights from the training process. %This assumption aligns with real-world practice, as described by Thompson et al\cite{NNN}, where testing without training data often warrants alternative strategies for input synthesis and model evaluation.
    
\textbf{Similar Accuracy Levels:} We assume that each pair of models under test yields similar and high accuracy on the currently available testing dataset. Such generation warrants the effective generation of triggering inputs to reveal behavioral disagreement between the DNN models, to better assess their respective capabilities, and thus enable their use in the various practical applications mentioned above.  
    
    %This similarity in performance sets a baseline from which deviations in behavior---triggered by the inputs generated during differential testing---can be further observed and analyzed.
    %This similarity assumption sets a baseline, highlighting the need for generating triggering inputs for thorough differential testing and model selection. 
    
\textbf{Limited Testing Budget:} We assume that a limited testing budget constrains our test generation approach, a valid assumption in many real-world situations when (1) DNNs are deployed on physical devices or with costly simulators in the loop, or (2) when the cost of the manual labeling or the validation of triggering inputs is high. In our study, we determine the testing budget based on the feasible execution time for test generation. We thus restrict the number of inputs to generate and analyze. Furthermore, a fixed and identical testing budget allows for a fair evaluation of alternative differential test generation approaches.

\subsection{Differences from Adversarial Inputs}

Adversarial samples are inputs subtly modified to mislead a specific single model by incorporating nearly imperceptible changes that may lead to incorrect predictions. Conversely, triggering inputs are designed to expose discrepancies between two versions of a model---for example, the original and its compressed counterpart---by demonstrating that they produce divergent outcomes when fed with the same input. This distinction is crucial for understanding how models behave after modification or for comparing the behavior of two distinct models designed for the same task. 
Tian \textit{et al.}~\cite{tian2023finding} demonstrated that, on the MNIST dataset, only 18 out of 10,000 adversarial samples generated using various adversarial attacks served as triggering inputs for differential testing. Similarly, prior studies \cite{8854377, Chen2021_AD} indicate that adversarial inputs can be limited in their ability to expose differences between compressed models.
Given that adversarial inputs target single model weaknesses and are not suitable for revealing differences in models' behavior, they are excluded from our differential testing analysis.

% adversarial samples of the original model rarely act as triggering inputs for compressed models. By using FGSM and CW adversarial methods~\cite{tian2023finding} 
% Also, several recent studies have shown that adversarial attacks are generally less effective at causing mispredictions in compressed models, as these models may exhibit increased robustness against such inputs manipulations

% From DFLARE paper:
% Note that adversarial samples of the original model are often not triggering inputs for compressed
% models. In our preliminary exploration, we have leveraged FGSM [20] and CW [7] to generate
% adversarial samples for three compressed models using MNIST. On average, only 18.6 out of 10,000
% adversarial samples are triggering inputs. Recent studies pointed out that compressed models can
% be an effective approach to defend against adversarial samples [12, 31].

% \\\\
\section{Approach: Reformulation as an NSGA-II Search Problem }
\label{Sec:Approach}

%A central problem in differential testing, especially when the DNN models under test have similar accuracy levels is to generate triggering inputs that reveal behavioural disagreements between models. Such inputs will enable to accurately learn when each of the models can be expected to fare better

A central problem in differential testing, particularly when evaluating DNN models that exhibit similar accuracy levels, is to generate triggering inputs that reveal behavioral disagreements between models. The generation and analysis of such inputs are crucial, as they not only reveal the specific conditions under which one model may outperform another but also enhance our understanding of the models' operational boundaries. This insight is essential for extracting the precise operational contexts in which each model is likely to fare best, thereby informing more targeted and effective application of these models in real-world scenarios.
Traditional approaches, such as transformation filters and adversarial techniques, often fall short in producing inputs that introduce novel characteristics or maintain realism, thus limiting their effectiveness in the comprehensive DNN assessment\cite{Bowles2018GANAA,articleSur} and differential testing of DNNs. To address this challenge, we introduce \textit{DiffGAN}, a differential testing approach for DNN models that leverages the capabilities of GANs and NSGA-II to drive the generation of diverse triggering inputs. The use of a GAN model for this purpose results from its proven efficacy in generating realistic and new test inputs across different DNN testing scenarios~\cite{zhang2018deeproad, 10074689, 8100115, FRIDADAR2018321}. When adequately trained, GANs can produce novel inputs that are distinct from the training data and belong to the same input domain---inputs that can be classified within the existing training labels.

%GANs, known for their ability to produce high-quality test inputs across various DNN testing scenarios, consist of a generator and a discriminator working in concert to explore the GAN's latent space—a high-dimensional domain encoding a compact representation of the training data. This latent space is key to the GAN's capacity for generating novel inputs that, while reflective of the training input domain, are distinct from the actual training data.

In a typical GAN architecture, two deep neural networks operate in tandem: a generator, to create new images from a random latent vector exploring the GAN's latent space, and a discriminator, whose purpose is to determine whether a given image is produced by the generator or a direct sample from the training dataset. The ultimate goal is for the generator to produce new inputs to ensure realism and compliance with the training distribution.
%so that the discriminator is unable to differentiate them from the inputs found in the training data.
Note that the latent space in the context of GANs refers to a high-dimensional space that encodes a compressed representation of the data on which the GAN is trained~\cite{radford2015unsupervised, shen2020interpreting}.

%We should note that the latent space in the context of GANs refers to a high-dimensional space that encodes a compressed representation of the data on which the GAN is trained. This space is essentially an abstract and compact domain, where points within this space can be mapped to realistic outputs by the GAN's generator network. The latent space is pivotal for the GAN's ability to generate diverse and new inputs because it serves as the source of variability in the generation process.
%The random vector, often referred to as a noise vector, plays a crucial role in exploring this latent space. It is a randomly generated set of numbers with a specified dimensionality that serves as the input to the generator network. This random vector is sampled from a predefined distribution (commonly a Gaussian distribution) and is used to traverse the latent space. By varying this random vector, the generator can produce a wide variety of outputs, even though it has never seen exactly these inputs during training.

%In essence, the latent space and the random vector together enable the GAN to generate new inputs that belong to the training inputs domain but are not identical to the training data. 

Since the models under test often exhibit similar high accuracy levels, GANs alone might not be effective in generating diverse triggering inputs that reveal behavioural disagreements between highly accurate models. To overcome these limitations, we leverage NSGA-II to effectively explore the large GAN's latent space, thereby enhancing the generation of diverse triggering inputs. More specifically, our genetic search aims at finding latent vectors that maximize the generation of inputs produced by the GAN that are both diverse and reveal behavioral disagreements between pairs of DNN models under test. 

\begin{figure*}[ht]
\centering
\includegraphics[scale=0.48]{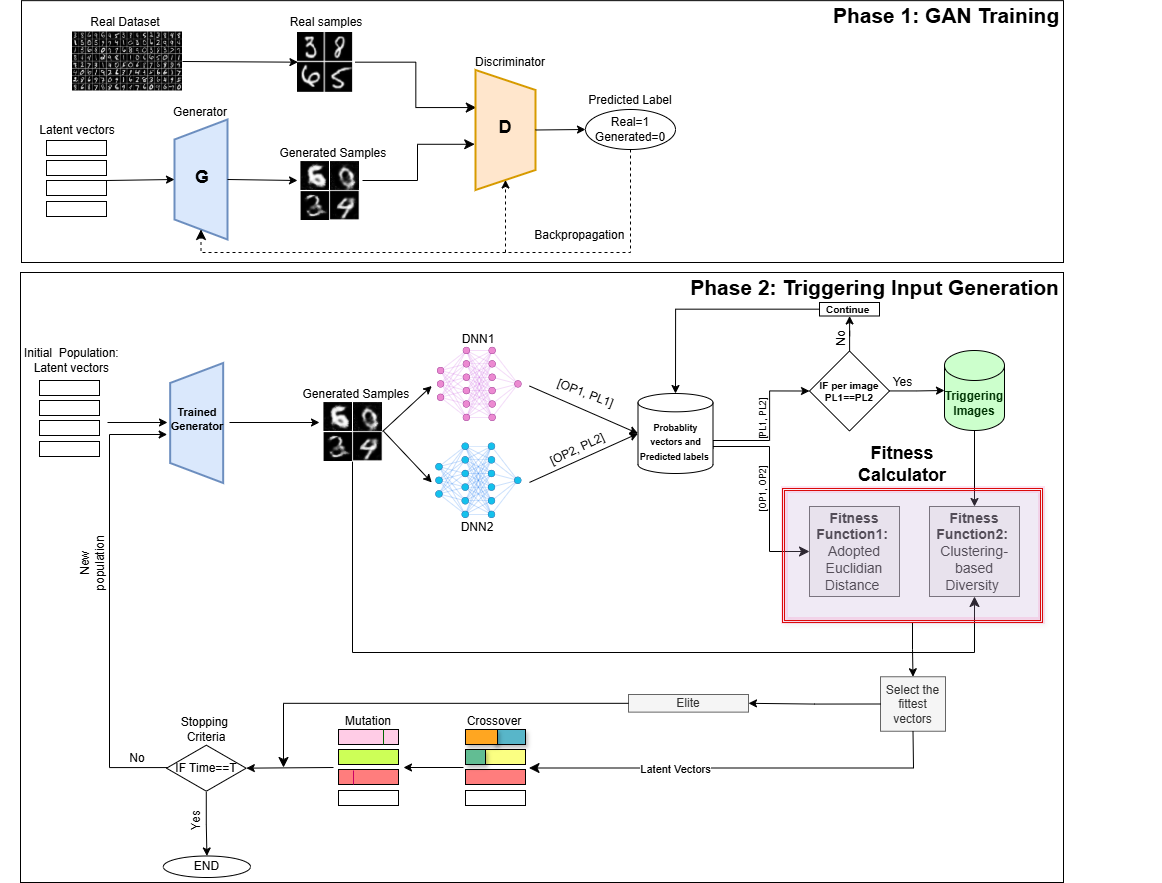}

\centering
\caption{\textbf{DiffGAN Framework Overview.} The figure illustrates the two-phase process of DiffGAN. \textbf{Phase 1} involves training a GAN using augmented test data to produce a realistic generator. After this, a transition occurs where the \textbf{trained generator} is carried forward into \textbf{Phase 2}, where NSGA-II is applied to search the generator’s latent space. Mutation and crossover operations, guided by divergence- and diversity-based fitness functions, iteratively evolve latent vectors to generate diverse and triggering inputs that expose behavioral discrepancies between DNN models.}

\label{fig:Approach}
\end{figure*}

We also rely on NSGA-II since it is specifically adapted to our multi-objective search problem. It is widely used in the literature and has shown good performance in solving many search-based problems in DNN testing~\cite{hemmati2013achieving,wang2015cost, Aghababaeyan2024}. Many existing works in the literature have demonstrated the effectiveness of NSGA-II in navigating the complex latent space of GANs and solving multi-objective optimization problems~\cite{9896140, wang2022nsgan, hou2022aerodynamic}.  Unlike algorithms such as MOSA, the Many Objective Sorting Algorithm~\cite{MOSA7102604}, which approaches each fitness function in isolation, NSGA-II tries to find solutions that offer diverse trade-offs among multiple objectives. More specifically, our search is driven by two objectives: (1) maximizing the difference between the output probability vectors of the inputs to generate triggering inputs, and (2) maximizing the diversity of generated inputs by the GAN model.
We illustrate in Figure~\ref{fig:Approach} our test generation approach and describe in the following the different steps of our approach, along with the different components of our search algorithm.

\subsection{GAN Training Dataset} \label{sec:GAN_Dataset}

To effectively train our GAN model, we use the testing dataset available for the models under test. Indeed, access to the original training datasets of these models is often restricted in many real-world scenarios, particularly with proprietary models, hence our reliance on testing datasets. 
%From this dataset, we select a subset of NNN inputs as our starting point. 
To enrich the diversity of the GAN’s training data, we enhance the testing dataset of the models under test with five realistic image transformations, encompassing operations such as rotation, shifting, shearing, and adjustments to the images' brightness and blurriness. These transformations are applied to every image in the dataset, following the same procedures and parameters established by Shorten \textit{et al.}~\cite{shorten2019survey}, which serve to ensure (1) the validity of the generated inputs and (2) label preservation by these transformations. Consequently, the augmented dataset feeding into the GAN's training process is expanded to five times its initial size, due to the application of five distinct image transformations on each original test input. Finally, we merge the original testing dataset with the augmented data to build the final GAN training dataset.
We should note that training a separate GAN (or its generator) is not needed for each pair of models under test. The GAN is trained on the shared test dataset of the models under test, augmented with data transformations. Since the test dataset remains consistent across model pairs operating within the same domain, a single GAN can be reused for all model pairs tested within that domain.

\subsection{GAN Model Training}\label{Sec:GAN_Traning}

We use a GAN model~\cite{radford2015unsupervised} in our approach, which encompasses a generator to produce new images and a discriminator whose role is to distinguish between the generated images and real images from the GAN's training dataset. %\Rev{We should note that }
% We should consider the unique challenges in evaluating GANs, which differ fundamentally from standard DNNs. Unlike classical DNN models focused on data classifications, for example, GANs are assessed based on their ability to generate diverse images which should be interpretable by human experts and belong to the targeted input domain.
There are unique challenges in evaluating generators in GANs, which differ fundamentally from standard DNNs.
In the realm of traditional DNNs, models' predictions across various tasks are compared against a ground truth, which consists of the actual outcomes or labels. However, generative models, such as the generator component of a GAN, follow a different paradigm. Their output is not a label but a new, synthetic image that did not previously exist in the dataset. As a result, traditional metrics based on accuracy against a ground truth are not applicable.

To evaluate the generator’s performance effectively, we employed two SOTA metrics designed for evaluating GAN’s performance: the \textit{Fréchet Inception Distance} (FID) and the \textit{Kernel Inception Distance} (KID)~\cite{xu2018empirical}. The FID metric quantifies the difference between the distribution of GAN-generated images and the distribution of actual images in the training dataset. Note that in the literature~\cite{10.5555/3295222.3295408}, FID scores typically range from zero to a few hundred, with lower scores indicating higher quality in image generation, suggesting better GAN performance.

The KID  measures the similarity between the distributions of real and generated images by computing the maximum mean discrepancy between their feature representations~\cite{xu2018empirical}. 
Unlike FID, KID is calculated using a kernel function, making it more robust to variations in sample size. KID scores range from 0 to 1, with the former reflecting superior quality in the images generated by the model.

%Note that in the literature\cite{NNN}, FID scores typically range from 0 to a few hundred, with lower scores indicating higher quality in image generation, suggesting better performance of the GAN. KID scores, on the other hand, operate on a more constrained scale, often ranging from 0 to 1. Scores nearing 0 r KID similarly reflect superior quality in the images generated by the model.
%These metrics were chosen for their strengths in assessing the quality and the diversity of the images generated by GANs. FID is sensitive to differences in terms of image distributions whereas KID is robust against varying sample sizes~\cite{xu2018empirical}.% Moreover, the FID showed a good correlation with human visual perception~\cite{greenspan2019uncertainty}.

These metrics were selected for their ability to assess both the quality and diversity of images generated by GANs. FID is sensitive to differences in image distributions, making it effective for evaluating the overall alignment between generated and real data distributions. On the other hand, KID is robust against varying sample sizes, providing accurate results even when only a few images are available~\cite{xu2018empirical}. Together, FID ensures that the generated images closely match the real distribution, while KID maintains robustness across different dataset sizes, making the two metrics complementary in their ability to provide a thorough evaluation of GAN-generated images~\cite{xu2018empirical}.

We train the GAN model using our training dataset and perform hyperparameter tuning using the Optuna framework~\footnote{https://optuna.readthedocs.io/en/stable/}, which is known for its efficiency in automating the optimization process for machine learning models~\cite{akiba2019optuna}. Optuna was selected for its user-friendly interface, capability to efficiently search ML models hyperparameter spaces, and support for a variety of search strategies. We also employ Optuna's pruner feature for stopping the evaluation of unpromising trials, which allows for expediting the optimization process. Our tuning process targets hyperparameters, learning rates, epochs, and batch sizes, and aims to minimize the FID and KID scores.

During each iteration of the Optuna optimization process, a GAN model is trained, and 10,000 images are generated by feeding 10,000 randomly generated latent vectors into the trained GAN. These generated images are compared with a randomly selected subset of 10,000 real images from the GAN’s training dataset. To ensure the reliability of our results and reduce the impact of randomness, this process is repeated ten times per iteration. The average FID and KID scores across these ten runs are computed for the currently trained GAN model, and Optuna uses these scores to guide the search for the next set of hyperparameters, aiming to minimize FID and KID. Finally, the GAN model with the lowest average scores is selected as the optimal model.

% For each GAN model trained in our study, we generate a set of 10,000 images by inputting 10,000 randomly chosen latent vectors into the GAN. These generated images are then compared to a randomly selected subset of 10,000 real images from the GAN training dataset. To ensure the reliability of our results and reduce the impact of randomness, this procedure is conducted ten times for each GAN model. We then compute the average FID and KID scores across the ten iterations for each trained GAN model. 
% The trained GAN that obtains the lowest average scores for both FID and KID is selected as the optimal model.
%with an average FID score of 0.33 and a KID score of 0.002 and will be used in the processing phase which we will describe in the following.

% The trained GAN that demonstrates the lowest average scores for both FID and KID is selected as the optimal model and will be used in the processing phase which we will describe in the following.

%For the empirical evaluation, we generated 1,000 images using 1,000 random input vectors to the GAN and compared these with a randomly selected set of real images from the test dataset. This process was repeated ten times to mitigate the effects of randomness, and we calculated the average scores for both FID and KID. The results showed an average FID of 0.33 and a KID of 0.0028, which, according to the literature, indicate excellent GAN performance. These scores underscore our model's high efficacy in image generation.

\subsection{Generation of Triggering Inputs}

After training the GAN, we leverage its trained generator to produce triggering images. We employ the NSGA-II search algorithm to effectively explore the GAN's latent space, guiding the search towards the generation of diverse triggering input images. In this approach, a genetic algorithm manipulates and evolves a population of latent vectors through successive crossover and mutation operations. This evolutionary procedure iteratively refines these latent vectors towards the objectives. The selection process is guided by specific fitness functions, which, in our context, prioritize the selection of latent vectors that maximize the generation of diverse triggering input images. In other words, through successive iterations, we expect to increasingly produce test inputs that reveal a high number of diverse behavioral disagreements between the DNN models under test. To properly translate the generation process into a search problem using NSGA-II, we need to define the following elements.

\subsubsection{\textbf{Individuals and Initial Population}}

In genetic search, the initial population contains individuals consisting of a set of elements called genes. These genes are connected and form an individual, also referred to as a solution. 
In our context, the initial population contains 100 randomly generated latent vectors, each of size 100, as recommended in the literature~\cite{fernandes2020evolutionary}. Each dimension of a latent vector in the initial population is independently generated from a uniform distribution over the range $[-1,1]$, thereby ensuring a diverse and unbiased initialization in the search space.

% We use random selection to build our initial population of individuals, thereby ensuring a diverse starting point for our genetic search.

\subsubsection{\textbf{Fitness Functions}} \label{subSec:Fitness}

The main objective of our approach is to generate test inputs that reveal diverse behavioral disagreements between the DNN models under test. To achieve this, our search process is guided by two fitness functions. The first function aims to maximize the difference between the output probability vectors of the DNN models under test when fed with the generated input images. By maximizing the difference between the output probability vectors, we are more likely to trigger more pronounced behavioral disagreements between the DNN models under test. 
The second function aims to maximize the diversity of the generated inputs since generating redundant inputs is a waste of computational resources, particularly when faced with constraints such as limited testing budgets and high labeling costs for testing data.
% (to accurately measure the performance of the models or extract the conditions under which one model fares better than the other for example). 
This strategy ensures that each input contributes to assessing the models' accuracy and identifying conditions where one model outperforms another.
By combining both fitness functions, we aim to generate diverse test input images that maximize the chances of revealing distinct behavioral disagreements between the DNN models under test.

\paragraph{\textbf{Divergence-based Fitness Function}}

This fitness function estimates the divergence between the output probability vectors of the models under test when fed with the same input image generated by the trained GAN. More specifically, the fitness function takes as input a GAN latent vector $Z$ responsible for generating the candidate test input image. We feed the models under test with such an image and compute the Euclidean distance between the resulting output probability vectors $Y_{m1}$ and $Y_{m2}$. The goal of maximizing this distance is to increase the probability of selecting latent vectors that lead to the generation of input images revealing significant behavioral disagreements between the models under test.
% However, since the Euclidean distance may yield similar scores for inputs revealing discrepancies or similarities in the predicted labels by the models under test, an adjustment of the fitness function is necessary. 
The Euclidean distance serves as a straightforward measure of the magnitude of discrepancy between model predictions, capturing how far apart the models' outputs are in the probability space. 
The choice of Euclidean distance over other metrics, such as KL divergence~\cite{hershey2007approximating}, is motivated by its efficiency in terms of computation cost.

%Unlike KL divergence, which quantifies the divergence between two probability distributions and can exhibit asymmetry (Cover, T. M., Thomas, J. A. (2006). Elements of Information Theory. Wiley-Interscience), Euclidean distance provides a symmetric, intuitive measure of the magnitude of discrepancy directly in the probability space. This makes Euclidean distance particularly suitable for a wide array of applications

However, the use of Euclidean distance or any other distance metric alone has a limitation: they do not account for the differences in predicted labels across models. This means that while a high Euclidean distance indicates a discrepancy in the output probabilities, it does not necessarily reflect a disagreement in the models' final predicted labels. This is particularly critical when the primary objective lies in searching for triggering inputs that lead to different label predictions by the models under test.

% To address this limitation and enhance the fitness function's sensitivity to discrepancies that are more consequential from a differential testing perspective, we introduced an adjustment by incrementing the Euclidean distance by one whenever the models predict different labels for the same input image.

% This increment acts as a penalty for label disagreement, elevating the importance of inputs that not only produce divergent probability vectors but also result in distinct classifications by the models.
%To illustrate the practical impact of this limitation and the effectiveness of our adjustment, consider these two specific examples:

We add the following example to illustrate the above limitation. Let us consider that for a generated image (image 1), the Euclidean distance calculated between the output probability vectors P1 = [0.05,0.05,0.05,0.85] from the first model under test (Model 1) and Q1 = [0.15,0.15,0.15,0.55] from the second model under test (Model 2) is 0.3. Further, both models predict the same class for image 1.
Then, let us consider another image (image 2), where the calculated Euclidean distance between the output probability vectors P2 = [0.1,0.2,0.3,0.4] from Model 1 and Q2 = [0.15,0.25,0.35,0.05] from Model 2 is also 0.3, despite the fact that these models predict different classes for image 2. We obtain the same Euclidean distances for both images, though clearly image 2 is more interesting given our objectives. 

Using a standard Euclidean distance in the context of differential testing is therefore not optimal and indicates the need to refine the fitness function, to ensure it not only considers the magnitude of input differences but also changes in the predicted labels, indicating different models' predictions.

%To enhance the accuracy of our fitness function in reflecting discrepancies in models predictions, we have adjusted it to increment the Euclidean distance by one whenever the models predict different labels for an image generated by the GAN. This adjustment specifically targets scenarios where the output labels from the models diverge, thereby ensuring that such cases are given higher priority in our analysis.

%This approach operates on the principle that, upon receiving a GAN latent vector, if the evaluated models generate differing labels for the produced image, the divergence-based fitness function augments the calculated Euclidean distance by one. This increment elevates the importance of these images in our testing, recognizing their potential to uncover significant behavioral disagreements between the models. Conversely, if the models agree on the label, the fitness function simply relies on the original Euclidean distance between their output probability vectors as illustrated in Equation~\ref{Eq:Diveregence_fitness}:

To address this limitation, we modified the fitness function to reflect differences in model predictions. More specifically, given a GAN latent vector, if the model's output different labels for a GAN-generated image, the fitness function increments the Euclidean distance by one to always assign a higher score for triggering input images. Otherwise, it returns the Euclidean distance between the two output probability vectors as illustrated in Equation~\ref{Eq:Diveregence_fitness}:
\begin{equation}\label{Eq:Diveregence_fitness}
\small
D(z)= 
\begin{cases}
      \sqrt{\sum_{i=1}^{n} (Y(i)_{m1}- Y(i)_{m2})^2}+1 & \text{if the models output}\\
      & \text{different labels}\\
      \sqrt{\sum_{i=1}^{n} (Y(i)_{m1}- Y(i)_{m2})^2} & \text{otherwise}
    \end{cases} 
\end{equation}

Where $z$ represents a GAN latent vector, $D(z)$ denotes the divergence of $z$, and $Y_{m1}$ and $Y_{m2}$ are the output probability vectors of the models under test.

% add the example results

To illustrate the practical impact of our modified fitness function, reusing the examples above, Image 1, a triggering input, receives a fitness score of 1.3, whereas Image 2, a non-triggering input, is assigned a score of 0.3. 
This adjusted fitness function more accurately characterizes inputs that are likely to reveal significant behavioral differences between models, prioritizing inputs that not only produce different probability vectors but also result in distinct classifications by the models.

\paragraph{\textbf{Diversity-based Fitness Function}}

Given a latent vector, this fitness function computes the contribution of the corresponding GAN-generated image to the diversity of the collected triggering inputs during the search process. To achieve this, the diversity-based fitness function calculates the minimum cosine distance score between the pixel values of the generated image and those of the set of collected triggering inputs during the search. 
%We use the pixel values as features. 

%For the similarity metric, we opt for cosine similarity, which is adept at measuring the angular difference between two vectors, rendering it exceptionally suitable for comparing image similarities within a high-dimensional context. Cosine similarity's effectiveness in identifying perceptual similarities between images, as opposed to mere pixel-value differences, is well-documented in the domain of image processing and information retrieval [ref: Singhal, A. (2001). Modern Information Retrieval: A Brief Overview. IEEE Data Eng. Bull.]

To reduce the computational complexity of comparing each generated image against all collected triggering inputs at each iteration of the search process, we approximate such distance by applying a clustering algorithm to group similar inputs. Then, we only return the minimum cosine distance between the generated image and the representative triggering inputs for all clusters. Preliminary experiments confirmed that, despite the cost of clustering, significant time could be saved in computing similarity scores of new images, thereby leading to the generation of more triggering inputs. 
More specifically, at each iteration of the search process, we apply the HDBSCAN~\cite{mcinnes2017hdbscan} clustering algorithm on the set of the collected triggering input images to group similar inputs in one cluster. %We use the image pixels as features. 
We select HDBSCAN for its ability to identify clusters of varying densities without requiring the prior specification of the number of clusters. This adaptability makes HDBSCAN particularly suited in our context. 

% Since computing the cosine distance between the generated images and all triggering inputs at each iteration of the search process is highly computationally expensive,
% we approximate such a distance by applying a clustering algorithm to group similar inputs. Then, we only report the maximum cosine distance between the generated image and the representative triggering inputs of each cluster. 

% More specifically, at each iteration of the search process, we apply the HDBSCAN clustering algorithm on the set of the collected triggering input images to group similar inputs in one cluster. We use the pixel images as features. 

Prior to clustering, we implement UMAP for dimensionality reduction, which assists in forming more distinct clusters~\cite{allaoui2020considerably,Aghababaeyan2024} at a lower cost. Subsequently, within each cluster, we identify the representative input---specifically, the input that has the maximum probability of association with its corresponding cluster\footnote{ HDBSCAN Documentation—retrieved October 5, 2024, from \url{https://hdbscan.readthedocs.io/en/latest/soft_clustering_explanation.html}}. The fitness function then calculates the diversity of the GAN-generated image by measuring the cosine distances between this image and the representatives of all clusters. Finally, the diversity-based fitness function returns the minimum of these cosine distances to quantify the diversity contribution of the image,  as illustrated in Equation~\ref{Eq:Diversity_fitness}:

\begin{equation}\label{Eq:Diversity_fitness}
Diversity(z)= \min_{i=1}^{m}{ CosineDistance(image_z, image_i)}
\end{equation}
where $z$ is the latent vector, $image_z$ is the generated image by GAN given $z$, $image_i$ is the representative image of the $i^{th}$ cluster formed by HDBSCAN, and $m$ is the total number of clusters. 

%To compute the contribution of the generated image by GAN given the latent vector in the population, the diversity-based fitness function measures the cosine distances between the generated input and each cluster representative and returns the maximum distance. 

%Before applying HDBSCAN we use UMAP for dimentionality reduction to help the clustering algorithm generating better clusters. We then select the representative input of each cluster based on the probability of its belonging to its corresponding cluster. more specifically, inputs that maximize such a probability are selected as representative.  

\paragraph{\textbf{Multi-Objective Search}}
 
We need to maximize in our search the two aforementioned fitness functions to drive the generation of diverse triggering inputs. This is therefore a multi-objective search problem that can be formalized as follows: 

\begin{equation}
\label{eq:fitness}
\begin{split}
    \max\limits_{z \in \mathbb{P}} \:\:& Fitness(z) =  (Divergence(z),Diversity(z))\\ 
\end{split}
\end{equation}

where $\mathbb{P}$ is the population of latent vectors, $z$ is a latent vector, function $Fitness: z \xrightarrow{} \mathbf{R}^2$ consists of two real-value objective functions $(Divergence(z),Diversity(z))$, and $\mathbf{R}^2$ is the objective space of our optimization problem.

\subsection{Genetic Operators}
We describe below the two genetic operators in our input generation approach.
\subsubsection{\textbf{Crossover}}
The crossover operator takes as input two parents (i.e., two latent vectors) and generates new offspring by combining the selected parents. Similar to existing works that employ NSGA-II on GAN latent vectors~\cite{9896140, YUAN2021120379}, we use the simulated binary crossover (SBX) in our approach as it is suitable for real-coded individuals (i.e., continuous optimization problem)~\cite{YUAN2021120379,deb1995simulated, 9896140}. Unlike traditional binary crossover operators, which are designed for discrete or binary representations, SBX operates directly on the real values of the latent vectors. Let $V_1$ and $V_2$ be the two selected parents for crossover. We provide the following example of the resulting offspring after applying crossover: 
\begin{equation}
Offspring_1(i)= 0.5 [(1+\beta)V_1(i)+(1-\beta)V_2(i)]
\end{equation}

\begin{equation}
Offspring_2(i)= 0.5 [(1-\beta)V_1(i)+(1+\beta)V_2(i)]
\end{equation}

where $\beta$ is a random variable controlling the extent of the crossover and $i$ denotes the index of the latent vectors' components.

\subsubsection{\textbf{Mutation}} \label{subsec:mutation}

Our genetic search also employs a mutation operator to further explore the search space. 
% Since we have a continuous optimization problem, we use the polynomial mutation (PM)~\cite{deb2001multi} of the selected individuals in our search.
Following existing works that use genetic algorithms in GAN latent space~\cite{9896140, YUAN2021120379}, we also adopt polynomial mutation~\cite{deb2001multi} to address the continuous nature of the latent space.

Polynomial mutation applies polynomial-distributed noise to the genes forming the selected individuals in our search. By applying such noise to the genes, it enables the exploration of nearby solutions while maintaining proximity to the original solution. This is also essential for efficiently navigating the complex and high-dimensional search space of GAN latent vectors. Polynomial mutation helps generate diverse solutions by introducing variability in the population of the search. This diversity is crucial for escaping local optima and finding globally optimal or near-optimal solutions.

%especially in our context where finding high-quality latent vectors that lead to the generation of triggering inputs is essential for our problem.

\subsection{Search Algorithm}

\begin{algorithm}[t!]

\DontPrintSemicolon

\KwInput{Initial population of individuals ( $ind$ $\in$ $P$ ), maximum running time ($T_{max}$), trained GAN, two DNNs under test}
\KwOutput{triggering inputs $\alpha$}

$\alpha \xleftarrow{} \emptyset$\;

$T_{start} \gets$ current time ;

\While{Current Time - $T_{start} \leq T_{max}$}{
    
    % Calculate fitness scores for all individuals in $P$ based on $T$ ;
    $P_{new} \xleftarrow{} \emptyset$\;
    $P_{Cross} \xleftarrow{} \emptyset$\;
    $P_{Cross} \xleftarrow{} Cross(P, CrossRate)$\;
    $P_{new} \xleftarrow{} Mutate(P_{Cross}, MutationRate)$\;
    
    \ForEach{$ind \in P$}{
    $\text{image}_{\text{gen}} \xleftarrow{} \text{GAN}(ind)$;

    $[OP1, OP2] \xleftarrow{} \text{Model1}(\text{image}_{\text{gen}}) , \text{Model2}(\text{image}_{\text{gen}})$\;
    
    \If {generated image is triggering}{$\alpha \xleftarrow{} \alpha \cup \{ind\}$}
    }
    $P \xleftarrow{}  Select (P, P_{new}, Fitness)$ \;
    $gen = gen +1$;}
    \KwRet{$\alpha$}\;
\caption{High-level NSGA-II algorithm}
\label{Algorithm-overview}
\end{algorithm}

The main objective of our search algorithm is to maximize the generation of diverse triggering inputs to effectively identify the discrepancies between the DNN models under test. A high-level overview of our NSGA-II search process is outlined in Algorithm~\ref{Algorithm-overview}. Assuming $P$ represents the initial population, consisting of a set of $|P|$ random latent vectors, the algorithm initiates an iterative procedure, executing the following steps in each generation until the maximum execution time ($T$) is reached (lines 3-14).
The search process includes the following steps. First, we create a new empty population $P_{new}$ (line 4). Second, based on predefined crossover and mutation rates, we create offspring using crossover ($Cross$) and mutation ($Mutate$) operators, incorporating newly created individuals to $P_{new}$ (lines 5-7).
Then, for each individual, we generate the corresponding image with the trained GAN and store the triggering images in an archive (lines 8-12). Next, we calculate the fitness scores of the new population by computing the diversity- and divergence-based fitness scores of each individual in the population.  We employ tournament selection ($Select$)~\cite{Miller96} to choose individuals for survival (line 13). Lastly, we update the population by incorporating the fittest individuals, proceeding to the next generation (line 14). The search terminates when the maximum execution time ($T$) is reached and returns the archive of generated triggering inputs (line 15).

We set the NSGA-II parameters in our search algorithm as follows. The crossover rate $CrossRate$ is set to 90\% and the mutation rate $MutationRate$ is set to 10\%, as recommended in the literature~\cite{10.1109/2.294849, Holland1992AdaptationIN,messaoudi2018search}.
Finally, we used Google Colab and the Pymoo library~\cite{blank2020pymoo} to implement our genetic search.

%\subsection{Enhancing the Validity of DiffGAN-Generated Inputs}

\subsection{Filtering Invalid Inputs}

Despite performing well on standard metrics such as FID and KID, \textit{DiffGAN}, like any input generation process, might still produce invalid inputs. To address this issue, we developed a two-stage filtration process for invalid triggering inputs, applied after the final generation step. This process incorporates two state-of-the-art filtration techniques to ensure the quality and validity of the generated images~\cite{1284395, you2023regression}.
Due to the computational cost associated with the filtration process, it is applied only after the final generation step rather than throughout the generation iterations with NSGA-II, optimizing both the testing budget and output quality. 

%In our research on Generative Adversarial Networks (GANs), we aimed to address the challenge of generating invalid or unrealistic images, a persistent issue even when GANs perform well on standard metrics like FID and KID. To tackle this issue, we developed a two-stage filtration process of invalid inputs incorporating two SOTA  techniques to ensure the quality and validity of the generated images.

\subsubsection{Discriminator-Based Filtering}

The first stage of our filtration process leverages the trained discriminator from our GAN model. We continuously evaluate our GAN's performance during training to ensure it reaches a high level of performance (in terms of GAN's evaluation metrics). After training, we use the discriminator to filter out unrealistic or invalid images. We should recall that when properly trained, the discriminator can be highly effective at distinguishing and removing invalid images that do not align with the desired output distribution. % do not meet quality standards. 
Since the discriminator is trained to differentiate between real and generated images, it serves as a straightforward filter, ensuring that only images deemed valid are retained for further analysis.

This approach aligns with findings from the literature~\cite{isola2017image, NEURIPS2022_6174c67b, you2023regression}, which demonstrate the utility of using discriminators not only during the training process but also for post-generation validation. By leveraging the capabilities of our trained discriminator, we ensure a context-specific and reliable validation process that is tailored to the dataset on which the GAN was originally trained.

\subsubsection{SSIM-Based Invalidity Detection}

As a complementary validation method, we incorporated Structural Similarity Index (SSIM)-based detection. SSIM is a metric that evaluates the similarity between two images by comparing their luminosity, contrast, and structure. In our method, each generated image is compared with its most similar real counterpart from the original training set. If the SSIM score is low, it indicates that the generated image is likely to be invalid due to its significant divergence from real data. A low similarity score indicates significant visual differences from its most similar real image, making it likely to be unrealistic or invalid.

To determine the SSIM threshold for filtering invalid inputs, we generated 1,000 test inputs using the trained GAN. For each generated input, we calculated its SSIM score by comparing it to the most similar image from the original training dataset. After manually labeling each input as either valid or invalid, we analyzed the distribution of SSIM scores for both categories. This procedure was applied to both datasets used in our experiments. We consistently observed that invalid inputs exhibited SSIM scores below 40\%. Based on these findings, we selected 40\% as the threshold for filtering out invalid inputs.

%For example, if a generated image has an SSIM score of 0.4 (i.e., 40\% similar) when compared to its most similar image in the GAN training dataset, this suggests that the generated image may no longer resemble the same object or concept. 

%The SSIM scores indicate that we are generating diverse and valid images. Most images fall within the 0.6–0.8 range, showing meaningful similarity to their closest counterparts in the training set while not being redundant or overly similar. Additionally, since no image scores higher than 0.95, we can confirm there are no exact matches or highly similar images to the training data, preserving diversity. By filtering out images with SSIM scores below 0.4, we ensure that we remove only unrealistic images while maintaining both diversity and validity in the generated set.

By integrating discriminator-based filtering with SSIM-based invalidity detection, we developed a two-stage filtration system for enhancing the quality and validity of GAN-generated images. The discriminator provides the first layer of defense against invalid images, while SSIM-based filtration, further refines this process by evaluating the structural similarity between each generated image and its closest counterpart in the original training data.
%while SSIM ensures that the generated images maintain a sufficient level of similarity to real examples from the training data. 
This dual filtration approach significantly reduces the risk of maintaining invalid triggering inputs, leading to more useful \textit{DiffGAN} outputs.

\section{Empirical Evaluation: Questions and Design}
\label{Sec:Evaluation}

\subsection{Research Questions} \label{Sec:RQs}

Our empirical evaluation is designed to answer the following research questions, covering the effectiveness of \textit{DiffGAN} in terms of finding triggering inputs,  their diversity and validity, and our ability to guide the training of a model selection mechanism that better learns the conditions under which a model fares best. 

% Number of triggering inputs
% Diversity
% Validity
% Ablation Study
% Conditions under which one model is better than the other

%\textbf{RQ2. Do we generate more valid triggering inputs than existing baseline with the same testing budget?} 

\textbf{RQ1. Do we generate more triggering inputs than \textit{DRfuzz} with the same testing budget?}

% We aim to investigate in this research question, the capacity of \textit{DiffGAN} to generate more triggering images than existing baselines within the same testing budget, defined by a predetermined execution time. This evaluation holds significant importance for differential testing, particularly when the objective is to uncover triggering images that elucidate behavioral discrepancies among highly accurate DNN models with closely matched performance levels. 
% % Generating a set of test inputs that reveal high behavioral disagreements between the DNN models under test is highly important in differential testing, as it should increase the effectiveness of the testing process while reducing testing costs. 
% Consequently, we aim in this research question to compare the effectiveness of \textit{DiffGAN} with existing baselines in terms of their ability to generate triggering inputs, while considering the same testing budget. 

We aim to study in this research question the effectiveness of \textit{DiffGAN} in producing more triggering inputs compared to a baseline, namely \textit{DRfuzz}, within the same testing budget.  Our choice of \textit{DRfuzz} is carefully justified below, along with a description of the technique. The testing budget in our context is defined by a predetermined and fixed execution time. Generating large numbers of triggering inputs that reveal behavioral disagreement between DNN models is essential to accurately learn when each of the models can be expected to fare better.
Relying on too few triggering images would not enable \textit{DiffGAN} to identify the subtle differences between models, especially those with similar high accuracy. Therefore, our research question aims to evaluate and contrast the efficiency of \textit{DiffGAN} against \textit{DRfuzz} in terms of their ability to generate more triggering inputs within time constraints.

\textbf{RQ2. How do \textit{DiffGAN} and \textit{DRfuzz} compare in terms of generating valid triggering inputs within the same testing budget?}

Generating large numbers of triggering inputs is not sufficient. To the largest extent possible, they also need to be valid. We, therefore, aim to compare the validity of the triggering inputs produced by \textit{DiffGAN} and those generated by \textit{DRfuzz}. The concept of input validity is crucial for DNN testing in general and differential testing in particular since it ensures that the generated inputs are both relevant and meaningful\Rev{~\cite{Tonella2023,you2023regression}}. 
More specifically, they should be valid and belong to the targeted input domain. 
%be valid, and be processed by the DNN models being tested. 
A valid input for a DNN model is defined as one that human experts within the domain can accurately recognize and categorize, using labels from the domain~\cite{Tonella2023}. For example, in the context of handwritten digit recognition, an image is considered valid if it can be recognized as a digit by a human expert, despite potential imperfections. 
We, therefore, aim to compare in this research question the proportions of valid inputs that \textit{DiffGAN} and \textit{DRfuzz} generate. This comparison is crucial because generating a large number of triggering inputs is only beneficial if these inputs are to a large extent valid, as defined above.

\textbf{RQ3. Does \textit{DiffGAN} find more diverse triggering inputs than \textit{DRfuzz}?}

This research question centers on the diversity of the triggering inputs generated by \textit{DiffGAN}, compared to those produced by \textit{DRfuzz}. Diversity in this context refers to the range of conditions under which the inputs cause the DNN models to exhibit behavioral differences. Further, a diverse set of triggering inputs is indicative of the approach's ability to explore and expose a broad spectrum of potential weaknesses across the models under test. Hence, we aim to assess not only the quantity and validity of triggering inputs but also the breadth of scenarios these inputs represent. 

% \textbf{RQ4. How do the different components of \textit{DiffGAN} contribute to generating more triggering inputs?} 

% We aim in this research question to study the effectiveness of \textit{DiffGAN}'s components in generating more triggering inputs. As described in Section~\ref{NNN}, \textit{DiffGAN} relies on a genetic algorithm to guide the generation of triggering inputs. The search algorithm is driven by dedicated fitness functions and customized genetic operators, each designed to direct the generation process toward inputs that effectively reveal diverse behavioral disagreements between DNN models. We aim to study the impact of each component of the search on \textit{DiffGAN}'s effectiveness in generating more triggering inputs through an ablation study. This will help us assess the importance of each component in \textit{DiffGAN} and further justify our design choices.

%\textbf{RQ4. Can we identify the specific conditions under which one model outperforms another?}
\textbf{RQ4. Can training of an ML-based model selection mechanism be guided more effectively using the triggering inputs generated by \textit{DiffGAN} compared to those produced by \textit{DRfuzz}?}

We aim in this research question to determine whether the triggering inputs generated by \textit{DiffGAN} can more effectively guide the training of a model selection mechanism, allowing it to learn the conditions under which one DNN model outperforms another. In many real-world scenarios, it is common to encounter multiple DNNs that perform similarly in terms of overall accuracy, but exhibit significant variations in their performance under specific input conditions\Rev{~\cite{britto2014dynamic,cruz2018dynamic,you2023regression,wang2022bet}}. This presents a challenge for developing a model selection mechanism that can dynamically choose the most suitable model based on the input characteristics without having to run all models in real-time. An effective solution to this problem is to use an ML-based selection mechanism, which can predict which of the available models will perform better for a given input\Rev{~\cite{britto2014dynamic,cruz2018dynamic}}.
However, training such an ML-based model selection mechanism requires diverse and informative training data that captures and reveals a wide range of behavioural differences between the models. This is where the generation of triggering inputs becomes critical for the selection mechanism to more effectively learn under which condition one model is preferable over the other.
Indeed, the effectiveness of the selection mechanism heavily depends on the quality of the triggering inputs used for training. Therefore, we aim to explore in this RQ whether \textit{DiffGAN} can better guide the training process compared to \textit{DRfuzz}. It is important to clarify that we do not intend to propose a novel ensemble method or voting mechanism in this work. Instead, this research question focuses on investigating an important application scenario, that is, studying whether \textit{DiffGAN} helps learn which model fares better under various input conditions and thus trains a model selection mechanism.

\subsection{Baseline Approach}

\textbf{\textit{Selection Process.}} Since our main focus is to generate triggering inputs that reveal behavioral disagreements between pairs of DNN models, we selected the baseline approaches against which to compare our approach according to the following criteria. 

First, the baselines should involve the generation of test inputs for more than one DNN model to reveal differences in DNNs behaviors. This criterion ensures that the baselines under comparison are aligned with our core objective of identifying divergences in DNNs' behavior.

Second, the baseline approaches should be generally applicable and should not make any assumption on the structural similarity between the DNN models under test (e.g., the approach should not only be applicable to retrained models or quantized ones).

Third, the baselines should be peer reviewed, and open-source implementations of the proposed solutions should be provided. This requirement enables rigorous replicability, thus allowing for a fair comparison of \textit{DiffGAN}'s effectiveness against established benchmarks.%Finally, we exclude baselines where the generation of triggering inputs is restricted by the number of initial inputs available in the testing dataset, 

Finally, we exclude baselines that limit the generation of triggering inputs based on the number of initial test inputs available. Specifically, some approaches cap the maximum number of triggering images to match the size of a seed set—a small subset selected from the initial test dataset.
This constraint can introduce bias into comparative evaluations. When methods are assessed within a fixed testing budget, often measured in terms of execution time, such limitations can lead to unfair performance comparisons. If one method is restricted by the number of initial inputs while others are not, it does not provide a level playing field for evaluation.
Furthermore, in practical scenarios, the availability of initial test inputs might be limited~\cite{li2019boosting,xie2019diffchaser}. This scarcity makes it challenging to gauge a baseline method's ability to generate additional triggering inputs beyond the initial set. From a practical standpoint, it is also difficult to accurately predict the maximum number of triggering inputs required to uncover all possible behavioral disagreements between the DNNs under test. Without knowing this number in advance, setting a cap based on initial seed inputs could hinder the thoroughness of the testing process.

%Such constraints can bias comparative results, especially when evaluating baselines within a fixed testing budget, often measured in execution time, leading to an unfair assessment of their performance.  Furthermore, in many practical contexts, the number of initial available test inputs can be limited, making it challenging to estimate the baseline's capacity to generate more triggering inputs.  Also, from a practical standpoint, it is difficult to accurately estimate in advance the maximum number of triggering inputs to generate that could cover all types of behavioral disagreements between the DNN models under test.  

Based on the first criterion we identified 
\textit{DiffChaser}~\cite{xie2019diffchaser}, 
\textit{BET}~\cite{wang2022bet},
\textit{Diverget}~\cite{yahmed2022diverget}, 
\textit{DRfuzz}~\cite{you2023regression}, and \textit{DFlare}~\cite{tian2023finding}, as candidates. %However, \textit{BET} and \textit{Diverget} are not open-source. 
%However, \textit{BET} and \textit{Diverget} were excluded due to their lack of open-source implementations. 
%Additionally, \textit{DFlare} was not selected because its approach to generating triggering inputs is constrained by the number of initial inputs in the testing dataset. The method applies multiple mutation rules to each initial test input to find a triggering input. Once a triggering input is found for a particular initial input, the method stops mutating that input and proceeds to the next one. This process limits the maximum number of possible triggering inputs to the number of initial inputs in the testing dataset, as it generates at most one triggering input per initial input.
%%It applies multiple mutation rules to each input until a triggering one is found, then proceeds to the next, limiting the maximum number of possible triggering inputs to generate the initial inputs in the testing dataset. 
%We also decided not to include \textit{DiffChaser} in our experiments, as \textit{DRfuzz} demonstrated superior performance in generating a greater number of trigger inputs~\cite{you2023regression}.
%Consequently, we only selected \textit{DRfuzz} as a baseline to evaluate the effectiveness of our proposed approach. 
However, \textit{BET} and \textit{Diverget} were excluded due to their lack of open-source implementations. 
We also decided not to include \textit{DiffChaser} in our experiments, as \textit{DRfuzz} has demonstrated superior performance in generating a greater number of trigger inputs~\cite{you2023regression}.
Consequently, we selected \textit{DRfuzz} as a baseline to evaluate the effectiveness of our proposed approach. 
We should note that \textit{DFlare} was excluded because its test generation process is inherently constrained by the number of initial test inputs. The method applies multiple mutation rules to each seed input but halts once a triggering input is found, preventing further mutation. As a result, \textit{DFlare} can generate at most one triggering input per initial seed, significantly limiting the overall number of triggering inputs.
To mitigate this limitation, we attempted to repeatedly feed the same seed inputs into \textit{DFlare} to increase the number of generated inputs. However, due to \textit{DFlare}’s low randomness in input generation, repeated executions produced highly redundant outputs. This redundancy is further amplified by the limited number of mutation operations and the exploitation of the same mutation operators that successfully increase behavioral disagreements between the models under test over the generations. As a result, when repeatedly fed with seed inputs, \textit{DFlare}’s mutation process introduces limited variations, failing to produce a diverse set of triggering inputs.
To quantify this limitation, we compared the diversity of the inputs generated by \textit{DFlare}, \textit{DRfuzz}, and our approach. Our results showed that \textit{DFlare}’s generated inputs had significantly lower diversity scores than those produced by both \textit{DRfuzz} and \textit{DiffGAN}. Therefore, we excluded this baseline in our study.
Detailed results of this comparison are available in Appendix 1.

\textbf{\textit{Selected Baseline.}} \textit{DRfuzz} is an approach designed for generating test inputs that uncover diverse regression faults in updated versions of DNN models. In the context of DNNs, a regression fault occurs when an input, correctly predicted by a previous model version, leads to incorrect predictions in the updated version. \textit{DRfuzz} initiates this process with seed inputs, upon which it applies various mutation rules to produce new inputs that trigger regression faults. To ensure the validity and relevance of these inputs, \textit{DRfuzz} employs the discriminator component of a trained GAN to filter invalid inputs. A customized reward function in \textit{DRfuzz} assigns priority scores to each mutation rule based on their effectiveness in generating diverse triggering inputs from the seed inputs. This selection strategy emphasizes mutation rules that are more likely to uncover behavioral disagreements between model versions. To align \textit{DRfuzz} with our specific research objectives,  we have customized the source code of \textit{DRfuzz} accordingly. We refined the approach to keep not only the inputs that one model predicts correctly while being mispredicted by the other model, but also any inputs that show behavioral disagreements between the models under test.

\begin{table*}[t]
\centering
\caption{Configurations for MNIST Models}
\label{tab:MNIST_Models}
\resizebox{\linewidth}{!}{%
\begin{tabular}{|c|c|c|c|c|c|c|c|}
\hline
\textbf{Pair} & \textbf{Model} & \textbf{Architecture} & \textbf{Training Data} & \textbf{Optimizer} & \textbf{Parameters} & \textbf{Accuracy} & \textbf{\# Mispredicted} \\ \hline
\multirow{2}{*}{\textbf{Pair 1}} & Model 1 & \textbf{4 Conv, 20 layers} & MNIST Org (60,000 images) & Adam (lr=0.001) & \#Param: 330,730, Batch: 128, Epoch: 10 & \textbf{99.26\%} & 74 \\ \cline{2-8} 
 & Model 2 & \textbf{7 Conv, 19 layers} & MNIST Org (60,000 images) & Adam (lr=0.001) & \#Param:: 327,242, Batch: 128, Epoch: 10 & \textbf{99.50\%} & 50 \\ \hline
\multirow{2}{*}{\textbf{Pair 2}} & Model 3 & 7 Conv, 33 layers & \textbf{\begin{tabular}[c]{@{}c@{}}40000 Org + (10\% shifts, 15\% zooms/rotations)\\ on 20,000 MNIST\end{tabular}} & RMSprop (lr=0.001) & \#Param: 696,402, Batch: 128, Epoch: 10 & \textbf{99.09\%} & 91 \\ \cline{2-8} 
 & Model 4 & 7 Conv, 33 layers & \textbf{\begin{tabular}[c]{@{}c@{}}40000 Org+ (5\% shifts, 10\% zooms/rotations) \\ on 20,000 MNIST\end{tabular}} & RMSprop (lr=0.001) & \#Param: 696,402, Batch: 128, Epoch: 10 & \textbf{99.55\%} & 45 \\ \hline
\multirow{2}{*}{\textbf{Pair 3}} & Model 5 & LeNet-5 & MNIST Org (60,000 images) & \textbf{SGD (lr=0.01)} & \#Param: 44,426, Batch: 32, Epoch: 10 & \textbf{97.04\%} & 296 \\ \cline{2-8} 
 & Model 6 & LeNet-5 & MNIST Org (60,000 images) & \textbf{Adam (lr=0.001)} & \#Param: 44,426, Batch: 32, Epoch: 10 & \textbf{98.93\%} & 107 \\ \hline
\multirow{2}{*}{\textbf{Pair 4}} & Model 7 & AlexNet & MNIST Org (60,000 images) & Adam (lr=0.00001) & \textbf{\#Param: 24,733,898, Batch: 128, Epoch: 9} & \textbf{98.85\%} & 115 \\ \cline{2-8} 
 & Model 8 & AlexNet & MNIST Org (60,000 images) & Adam (lr=0.0001) & \textbf{\#Param: 24,733,898, Batch: 256, Epoch: 14} & \textbf{99.33\%} & 67 \\ \hline
\end{tabular}%
}
\end{table*}

\begin{table*}[]
\centering
\caption{Configurations for Cifar-10 Models}
\label{tab:Cifar10_Models}
\resizebox{\linewidth}{!}{%
\begin{tabular}{|c|c|c|c|c|c|c|c|}
\hline
\textbf{Pair} & \textbf{Model} & \textbf{Architecture} & \textbf{Training Data} & \textbf{Optimizer} & \textbf{Parameters} & \textbf{Accuracy} & \textbf{\# Mispredicted} \\ \hline

\multirow{2}{*}{\textbf{Pair 1}} 
& Model 1 & \textbf{ResNet18} & Cifar-10 Org (60,000 images)  & SGD (lr = 1e-1) & \#Param: 11183562, Batch=128, Epoch=200 & \textbf{83.38\%} & 1662 \\ \cline{2-8} 
& Model 2 & \textbf{ResNet20} & Cifar-10 Org (60,000 images) & SGD (lr = 1e-1) & \#Param: 274442, Batch=128, Epoch=200 & \textbf{85.71\%} & 1429 \\ \hline

\multirow{2}{*}{\textbf{Pair 2}} 
& Model 3 & Inception & \textbf{\begin{tabular}[c]{@{}c@{}}40000 Org + (15\% shifts, 5\%zooms, 10\%roations)\\ on 20,000 Cifar-10\end{tabular}} & SGD (lr = 1e-1) & \#Param: 5984858, Batch=128 , Epoch=120& \textbf{93.02\%} & 698 \\ \cline{2-8} 
& Model 4 & Inception & \textbf{\begin{tabular}[c]{@{}c@{}}40000 Org + (10\% shifts, 10\%zooms, 15\%rotations)\\ on 20,000 Cifar-10\end{tabular}}& SGD (lr = 1e-1) & \#Param: 5984858, Batch=128, Epoch=120& \textbf{93.27\%} & 673 \\ \hline

\multirow{2}{*}{\textbf{Pair 3}} 
& Model 5 & Inception & Cifar-10 Org (60,000 images) & \textbf{SGD (Nesterov=True)} & \#Param: 5984858, Batch=128,  Epoch=150 & \textbf{89.10\%} & 1090 \\ \cline{2-8} 
& Model 6 & Inception & Cifar-10 Org (60,000 images) & \textbf{Adam (lr = 1e-3)} & \#Param: 5984858, Batch=128, Epoch=150  & \textbf{92.78\%} & 722 \\ \hline

\multirow{2}{*}{\textbf{Pair 4}} 
& Model 7 & ResNet18 & Cifar-10 Org (60,000 images) & SGD (lr = 1e-1) & \textbf{\begin{tabular}[c]{@{}c@{}}\#Param: 11183562, Batch=250, Epoch=150 \end{tabular}} & \textbf{80.85\%} & 1915 \\ \cline{2-8} 
& Model 8 & ResNet18 & Cifar-10 Org (60,000 images) & SGD (lr = 1e-1) & \textbf{\begin{tabular}[c]{@{}c@{}}\#Param: 11183562, Batch=128, Epoch=200\end{tabular}} & \textbf{83.38\%} & 1662 \\ \hline

\end{tabular}%
}
\end{table*}

\subsection{Datasets and Models}\label{Sec:Dataset}

\begin{figure}[]
    \centering
    \includegraphics[scale=0.325]{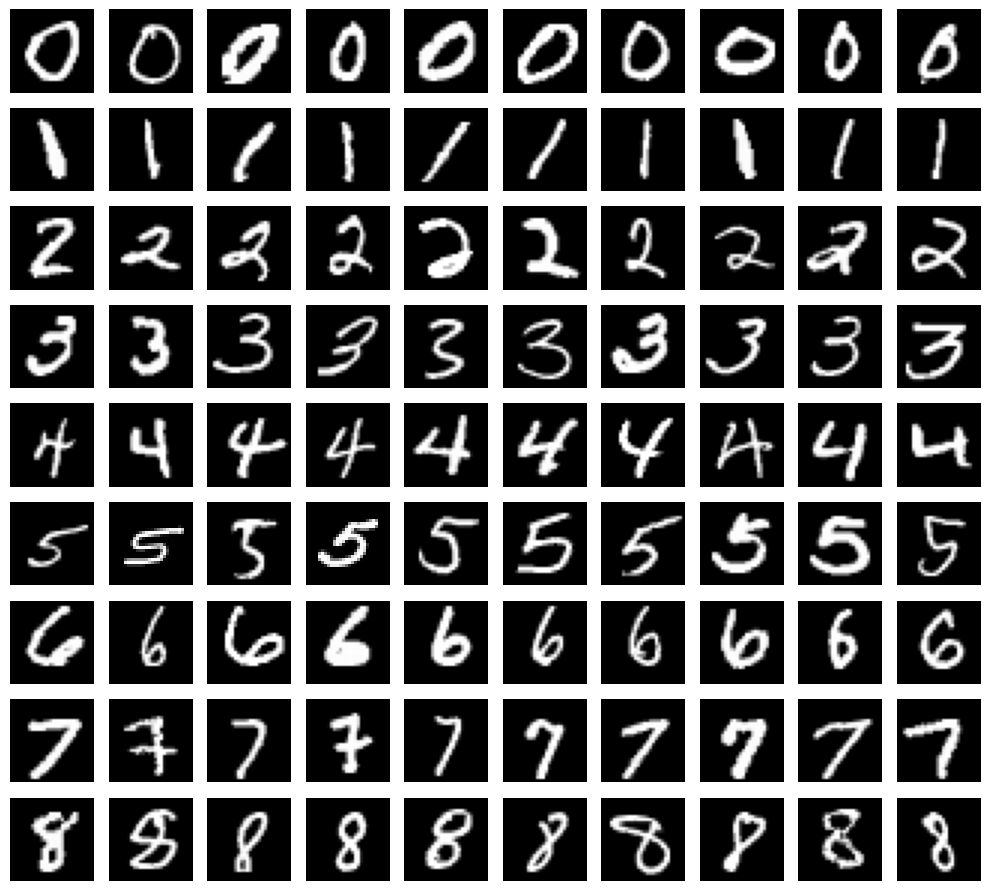}
    \caption{Original MNIST images (considered valid).}
    \label{fig:mnist}
\end{figure}

\begin{figure}[]
    \centering
    \includegraphics[scale=0.367]{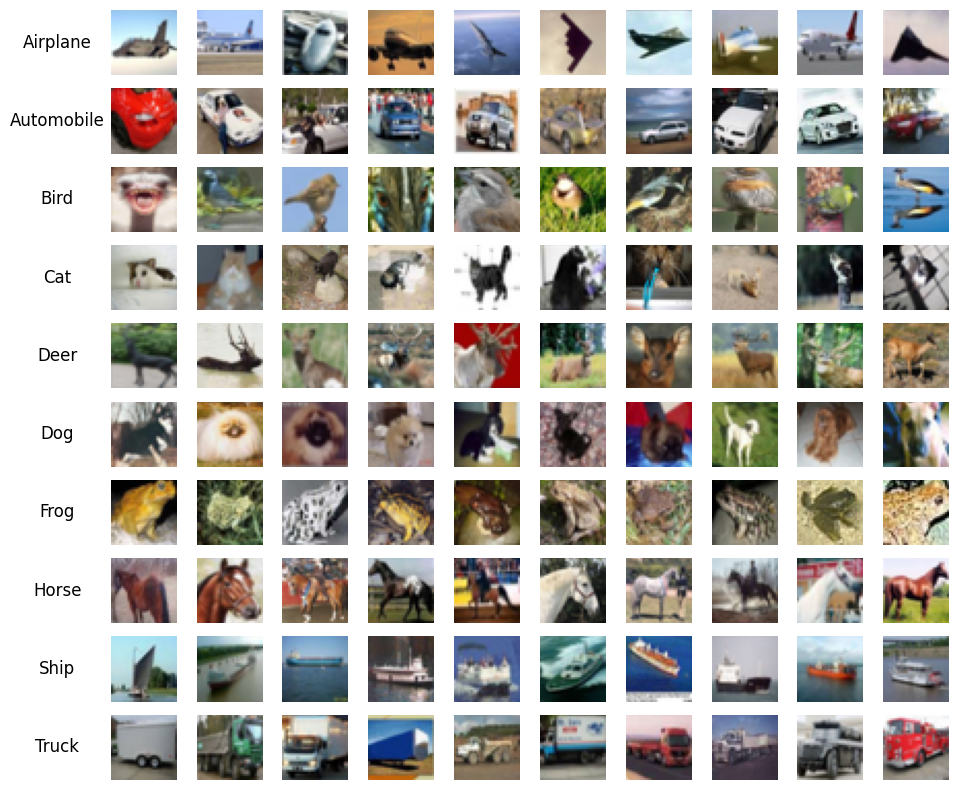}
    \caption{Original Cifar-10 images (considered valid).}
    \label{fig:cifar10}
\end{figure}

We consider different combinations of datasets and models in our empirical evaluation. More specifically, we consider MNIST~\cite{deng2012mnist} and Cifar-10~\cite{Cifar10}, two well-known image recognition datasets. 
As shown in Figure~\ref{fig:mnist}, the MNIST dataset comprises 70,000 black-and-white images of handwritten digits, divided into a training set of 60,000 images and a test set of 10,000 images. The Cifar-10 dataset comprises 60,000 color images distributed across ten categories (Figure~\ref{fig:cifar10}), including 50,000 images designated for training and 10,000 images for testing. %The Fashion-MNIST dataset, similar in structure to MNIST, includes 70,000 grayscale images of fashion items, also divided into 60,000 training images and 10,000 testing images.
These datasets were employed in conjunction with 16 different DNN models, which included both custom-developed and widely used architectures such as ResNet, LeNet-5, AlexNet, and inception models~\cite{krizhevsky2012alexnet, szegedy2015inception, lecun1998lenet}. Comprehensive details about the combinations of datasets and models we used, along with their accuracy, are outlined Tables~\ref{tab:MNIST_Models} and~\ref{tab:Cifar10_Models}.

As shown in the tables, our experimental study spanned four distinct and common sources of variation per dataset across DNN models, including the model structures, training data, optimizers, and hyperparameters. Our experiments thus cover 16 different DNN models, each selected to evaluate the effectiveness of our test generation approach. For each model pair, the differences in DNN models are highlighted in bold in Tables~\ref{tab:MNIST_Models} and~\ref{tab:Cifar10_Models}.  %Our motivation is to study \textit{DiffGAN} under different common, practical scenarios and thus increase the generalizability of our results. 
Typical differences across models include:

\begin{enumerate}
    \item \textbf{Structural Variations:} We considered models with different architectures. For example, in the first configuration in Table~\ref{tab:MNIST_Models}, the two models, although both trained on the MNIST dataset, have a different architecture. More specifically, the first model featuring four convolutional layers among a total of 20 layers, achieved a 99.26\% accuracy on the MNIST test dataset. In contrast, the second model, with seven convolutional layers out of a total of 19 layers, achieved a 99.50\% accuracy.

    \item \textbf{Training Data Variations:} Our analysis also covered models that are architecturally identical but trained on different training datasets. For example, as depicted in the second configuration in Table~\ref{tab:MNIST_Models}, the first DNN model was trained using 40,000 images from the MNIST training dataset and 20,000 MNIST images modified by 10\% shifts and 15\% zooms/rotations, achieving 99.09\% accuracy. Conversely, the second model was trained on 40,000 images from the MNIST training dataset and 20,000 MNIST images adjusted by 5\% shifts and 10\% zooms/rotations, yielding a slightly higher accuracy of 99.55\%.

    \item \textbf{Optimizer Variations:} We also considered the scenario where two models are trained with different optimizers. For instance, in a case study involving the Cifar-10 dataset (see Pair 3 in Table~\ref{tab:Cifar10_Models}), two inception models were trained using distinct optimizers. Specifically, one model was trained with the SGD optimizer and reached an accuracy of 89.10\%, whereas the other model was trained with the Adam optimizer, reaching an accuracy of 92.78\%.

    \item \textbf{Hyperparameter Differences:} Finally, we considered the scenario where two models are different in terms of hyperparameter values. For instance,  we trained two AlexNet models on the MNIST dataset (see Pair 4 in Table~\ref{tab:MNIST_Models}) with varying hyperparameters such as learning rate, epochs, and batch sizes. The first model, set with a learning rate of 0.00001, 9 epochs, and a batch size of 128, reached a 98.85\% accuracy. Conversely, the second model, with a learning rate of 0.0001, 14 epochs, and a batch size of 256, demonstrated a 99.33\% accuracy.

\end{enumerate}

%In our experimental setup, we considered four distinct and common sources of variation across DNN models: model structure, training data, optimizers, and hyperparameters. 
These particular variations were chosen because they represent key aspects of model development and testing that are frequently encountered in practice, for example in frameworks like MLflow\footnote{\url{https://mlflow.org}} to support machine learning development, which is widely used for managing machine learning experiments. MLflow tracks models across these four dimensions, emphasizing their importance in both research and real-world applications.

Further, we selected models ranging from simple architectures, such as LeNet, to more complex ones like AlexNet and ResNet, particularly for the Cifar-10 dataset. This choice allowed us to evaluate our framework across a broad spectrum of models, from smaller, less complex architectures to larger, more sophisticated ones. We could thus assess how well our framework performs under different computational and architectural constraints.

Additionally, we specifically focused on model pairs where the accuracy difference between models was small. More specifically, the accuracy differences across all model pairs range from 0.24\% to 3.68\%. We further report in Tables~\ref{tab:MNIST_Models} and~\ref{tab:Cifar10_Models}  the number of mispredicted inputs for each model and highlight their small differences for each model pair.  This ensures that our evaluation captures subtle differences in model performance. Detecting such small variations is crucial for demonstrating that our framework can reliably identify differences even when models achieve nearly identical, high accuracy.

\label{Sec:Results}

% % Display CIFAR-10 Image
% \begin{figure}[H]
%     \centering
%     \includegraphics[width=\textwidth]{figures/Cifar10.png}
%     \caption{Examples of original CIFAR-10 images. Each row shows 10 samples from one of the 10 classes: Airplane, Automobile, Bird, Cat, Deer, Dog, Frog, Horse, Ship, and Truck.}
%     \label{fig:cifar10}
% \end{figure}

% \vspace{1cm}

% % Display MNIST Image
% \begin{figure}[H]
%     \centering
%     \includegraphics[width=\textwidth]{figures/mnist.png}
%     \caption{Examples of original MNIST images. Each row shows 10 samples from one of the 10 handwritten digits: 0 to 9.}
%     \label{fig:mnist}
% \end{figure}

% Display CIFAR-10 Image in one column

% \vspace{0.cm}

% Display MNIST Image in one column

% \begin{figure}[]
%     \centering
%     \includegraphics[scale=0.25]{figures/mnist.png}
%     \caption{Examples of original MNIST images. }
%     \label{fig:mnist}
% \end{figure}

% \label{Sec:Results}
% \begin{figure}[]
%     \centering
%     \includegraphics[scale=0.3]{figures/Cifar10.png}
%     \caption{Examples of original Cifar-10 images.}
%     \label{fig:cifar10}
% \end{figure}

\subsection{Choice of the Fitness Functions}\label{Sec: Ablation Study}

To assess the contribution of each fitness function to the effectiveness of our approach in terms of generating diverse triggering inputs, we conducted an ablation study by isolating the impact of each fitness function used in our multi-objective search: divergence and diversity. 
We executed three configurations of our search algorithm:
\begin{itemize} 
\item \textbf{Divergence Only:} The search is guided solely by the divergence-based fitness function, which maximizes the disagreement between the output probability vectors of the DNN models under test. 
\item \textbf{Diversity Only:} The search is guided solely by the diversity-based fitness function, which maximizes the diversity among the generated inputs. 
\item \textbf{Combined Objectives (Full \textit{DiffGAN}):} The full \textit{DiffGAN} configuration using \textit{NSGA-II} to optimize both fitness functions simultaneously. 
\end{itemize}

Due to the computational expense of our experiments, we conducted this study on two model pairs with the Cifar-10 dataset.  Seeing significant differences across configurations on such pairs would, however, be enough to justify the combined use of divergence and diversity. 
%Each configuration was executed for one hour.
We studied both the total number of triggering inputs generated by each configuration and their diversity. To quantify diversity, we leveraged two metrics: Shannon Diversity~\cite{6773024,119732} and Geometric Diversity~\cite{aghababaeyan2021black}. A detailed explanation of these metrics, along with our methodology for computing diversity scores, is provided in Section~\ref{sec:RQ3_res}. The results of this ablation study are summarized in Table~\ref{tab:ablation_pairs_1_2}.

\begin{table*}[t]
\centering
\caption{Ablation Study Results for Two Model Pairs with Cifar-10}
\begin{tabular}{|c|l|c|c|c|}
\hline
\textbf{Models \& Dataset} & \textbf{Configuration} & \textbf{GD} & \textbf{Exponential Shannon ($e^{H}$)} & \textbf{\# Triggering Images} \\
\hline
\multirow{3}{*}{M1, M2 (Cifar-10)} 
    & \textit{DiffGAN} & 2396.99 & $e^{14.84} = 2.79 \times 10^6$ & 80,700 \\ \cline{2-5}
    & Divergence Only  &  739.10 & $e^{13.23} = 5.57 \times 10^5$ & \textbf{109,146} \\ \cline{2-5}
    & Diversity Only  & \textbf{2656.00} & \textbf{$e^{14.94} = 3.08 \times 10^6$} & 14,538 \\ \cline{2-5}
\hline
\multirow{3}{*}{M3, M4 (Cifar-10)} 
    & \textit{DiffGAN} & 2395.81 & $e^{14.84} = 2.79 \times 10^6$ & 62,596 \\ \cline{2-5}
    & Divergence Only  &  892.54 & $e^{13.46} = 7.01 \times 10^5$ & \textbf{90,811} \\ \cline{2-5}
    & Diversity Only  &\textbf{ 2652.34} & $e^{14.94} = 3.08 \times 10^6$ & 11,746 \\ \cline{2-5}
\hline
\end{tabular}
\label{tab:ablation_pairs_1_2}
\end{table*}

As shown in the table, we had consistent results for both model pairs. 
More specifically, our results show that using the divergence-based fitness function alone leads to a higher number of triggering inputs compared to the diversity-only configuration. However, these inputs often exhibit redundancy, which limits the exploration of diverse disagreement scenarios. On the other hand, the diversity-only configuration generates more diverse inputs but results in fewer behavioral disagreements between models. The full \textit{DiffGAN} configuration, which optimizes both objectives simultaneously, provides a good trade-off between the number of generated triggering inputs and their diversity. Not only does it produce a substantial number of triggering inputs, but it also maintains high diversity among them. This confirms the complementary nature of the two fitness functions and underscores the importance of combining them to generate a diverse and informative set of triggering inputs within a limited testing budget.

%However, these inputs exhibit a high degree of redundancy, which limits the exploration of diverse disagreement scenarios. On the other hand, the diversity-only configuration generates more diverse inputs but results in fewer behavioral disagreements between models. The full \textit{DiffGAN} configuration, which optimizes both objectives simultaneously, provides a good trade-off between the number of generated triggering inputs and their diversity. Not only does it produces a substantial number of triggering inputs, but it also maintains high diversity among them. This confirms the complementary nature of the two fitness functions and underscores the importance of combining them to generate a diverse and informative set of triggering inputs within a limited testing budget.}

\section{Empirical Results}

In this section, we describe our experimental evaluation and report the results. We first investigate whether we generate more triggering inputs than the baseline with the same testing budget. We then study the validity and diversity of the generated triggering inputs by both approaches. Finally, we present an important application scenario of the generated triggering inputs, namely the development of a machine learning-based mechanism to predict the best model's output based on input characteristics, and report the corresponding results. % describe our experiments and the results of our analysis of the specific conditions under which one model outperforms another. 

\subsection{\textbf{RQ1. Do we generate more triggering inputs than the existing baseline approach with the same testing budget?}} 

To answer this research question, we executed \textit{DiffGAN} and \textit{DRfuzz} considering the same execution time $T \in [1h, 2h, 3h]$,  across four pairs of DNN models and two distinct datasets, resulting in eight unique combinations (Section~\ref{Sec:Dataset}). Because of the randomness in \textit{DiffGAN} and \textit{DRfuzz}, we re-executed each of them three times and reported the corresponding number of triggering inputs in Tables~\ref{tab:Rq1-1} and~\ref{tab:Rq1-2}. This decision to limit the execution of the approaches to three iterations was initially due to the considerable execution time associated with the experiments. More specifically, our experiments take 96h (8 model pairs x 6h x 2 approaches) if we execute each approach once. As shown in the tables, \textit{DiffGAN} consistently outperforms \textit{DRfuzz} in terms of generating more triggering inputs across all models and datasets. 
%This consistent performance across different settings highlights the effectiveness of \textit{DiffGAN} in generating more triggering inputs. 
More specifically, our results show that \textit{DiffGAN} generates on average four times more triggering inputs than \textit{DRfuzz} when considering the same testing budget. 
We also calculated the Coefficient of Variation (CV)~\cite{everitt1998cambridge}, a measure of relative variability that compares the standard deviation to the mean, independent of the data's range. %This makes CV particularly useful for comparing variability across datasets with different units or scales. 
We computed the CV score for each model pair and testing budget across the three runs. 
A lower CV indicates higher consistency across different runs. For example, in our analysis of the MNIST dataset, \textit{DRfuzz} showed a CV between 0.33\% and  21.23\%,% indicating small variability in the number of generated triggering inputs across the three runs. 
while \textit{DiffGAN} exhibited a CV between 1.33\% and 12.20\%, suggesting less variability in the number of generated triggering inputs across the different runs. These observations are also consistent for the Cifar-10 dataset. These CV values demonstrate the relative stability and reliability of \textit{DiffGAN} in generating triggering images, with less variability compared to \textit{DRfuzz}.

% Minimum CV: 0.006 (M7, M8 - 2h)
% Maximum CV: 0.204 (M7, M8 - 1h) DRFUZZ Cifar10,     

% Minimum CV: 0.002 (M1, M2 - 1h)
% Maximum CV: 0.087 (M7, M8 - 2h)
% So, the CV for DiffGAN ranges from 0.002 to 0.087 Cifar10

%We should note that, unlike traditional mutation-based methods such as \textit{DRfuzz}, the generative model in \textit{DiffGAN} continuously learns and adapts based on the discriminator's feedback, improving its ability to generate inputs that are more likely to trigger behavioral disagreements between the DNN models under test. This iterative process leads to a more efficient and targeted generation of inputs that challenge the model, thus producing more triggering inputs than mutation-based methods. 

We should note that \textit{DRfuzz} generates inputs by applying a sequence of mutation rules to seed images, optimizing the order of these transformations to increase the likelihood of finding triggering inputs. However, this approach is by definition limited by the seed images and the specific transformations applied to them, which significantly restricts the potential triggering inputs that can be generated. In contrast, \textit{DiffGAN} explores a much larger input space by optimizing latent vectors within the GAN's input space through genetic search. The GAN's input space inherently offers a broader range of possible image variations than the transformations applied in \textit{DRfuzz}. By optimizing these latent vectors, \textit{DiffGAN} can generate novel combinations of inputs that may not be generated through mutation-based methods alone. This broader exploration capability thus allows \textit{DiffGAN} to produce more triggering inputs, as it can effectively explore the GAN's latent space to uncover a wider range of inputs that may reveal more behavioral disagreements between the DNN models under test.

Moreover, when comparing the results to the initial set of triggering inputs in the original testing datasets, both \textit{DiffGAN} and \textit{DRfuzz} demonstrate significant improvements. The improvement is measured as the ratio of the difference between the number of triggering inputs generated through the differential testing approach and the size of the initial set of triggering inputs, divided by the size of this initial set. Specifically, \textit{DiffGAN} shows an improvement ranging from 2,660\% to 12,727\% in the MNIST dataset and from 3,956\% to 26,606\% in the Cifar-10 dataset. Moreover,  \textit{DRfuzz} achieves an improvement ranging from approximately 954\% to 5,465\% in configurations with MNIST and from 908\% to 3,509\% in configurations with Cifar-10. Though both approaches are effective in enhancing the generation of triggering inputs, \textit{DiffGAN} is significantly more so.

\begin{table}[]
\caption{\ RQ1 results: Number of generated triggering images for models trained on MNIST dataset using DiffGAN and DRfuzz.\Revv{The \#TI (Initial) column indicates the number of triggering inputs identified in the original test dataset for each set of models.}}
\large
\label{tab:Rq1-1}
\resizebox{\columnwidth}{!}{
\begin{tabular}{|
>{\columncolor[HTML]{FFFFFF}}c |
>{\columncolor[HTML]{FFFFFF}}c 
>{\columncolor[HTML]{FFFFFF}}c |
>{\columncolor[HTML]{FFFFFF}}c |
>{\columncolor[HTML]{FFFFFF}}c |
>{\columncolor[HTML]{FFFFFF}}c |}
\hline
{\color[HTML]{000000} \textbf{Models}}                                  & \multicolumn{2}{c|}{\cellcolor[HTML]{FFFFFF}{\color[HTML]{000000} \textbf{Configs}}}                                   & {\color[HTML]{000000} \textbf{DRfuzz}} & {\color[HTML]{000000} \textbf{\#TI (Initial) }} & {\color[HTML]{000000} \textbf{DiffGAN}} \\ \hline
\cellcolor[HTML]{FFFFFF}{\color[HTML]{000000} }                         & \multicolumn{1}{c|}{\cellcolor[HTML]{FFFFFF}{\color[HTML]{000000} }}                     & {\color[HTML]{000000} run1} & {\color[HTML]{000000} 1229}            & {\color[HTML]{000000} 82}                      & {\color[HTML]{000000} \textbf{4932}}    \\ \cline{3-6} 
\cellcolor[HTML]{FFFFFF}{\color[HTML]{000000} }                         & \multicolumn{1}{c|}{\cellcolor[HTML]{FFFFFF}{\color[HTML]{000000} }}                     & {\color[HTML]{000000} run2} & {\color[HTML]{000000} 1237}            & {\color[HTML]{000000} 82}                      & {\color[HTML]{000000} \textbf{5188}}    \\ \cline{3-6} 
\cellcolor[HTML]{FFFFFF}{\color[HTML]{000000} }                         & \multicolumn{1}{c|}{\multirow{-3}{*}{\cellcolor[HTML]{FFFFFF}{\color[HTML]{000000} 1h}}} & {\color[HTML]{000000} run3} & {\color[HTML]{000000} 1228}            & {\color[HTML]{000000} 82}                      & {\color[HTML]{000000} \textbf{4963}}    \\ \cline{2-6} 
\cellcolor[HTML]{FFFFFF}{\color[HTML]{000000} }                         & \multicolumn{1}{c|}{\cellcolor[HTML]{FFFFFF}{\color[HTML]{000000} }}                     & {\color[HTML]{000000} run1} & {\color[HTML]{000000} 2724}            & {\color[HTML]{000000} 82}                      & {\color[HTML]{000000} \textbf{8091}}    \\ \cline{3-6} 
\cellcolor[HTML]{FFFFFF}{\color[HTML]{000000} }                         & \multicolumn{1}{c|}{\cellcolor[HTML]{FFFFFF}{\color[HTML]{000000} }}                     & {\color[HTML]{000000} run2} & {\color[HTML]{000000} 2525}            & {\color[HTML]{000000} 82}                      & {\color[HTML]{000000} \textbf{7539}}    \\ \cline{3-6} 
\cellcolor[HTML]{FFFFFF}{\color[HTML]{000000} }                         & \multicolumn{1}{c|}{\multirow{-3}{*}{\cellcolor[HTML]{FFFFFF}{\color[HTML]{000000} 2h}}} & {\color[HTML]{000000} run3} & {\color[HTML]{000000} 3427}            & {\color[HTML]{000000} 82}                      & {\color[HTML]{000000} \textbf{6853}}    \\ \cline{2-6} 
\cellcolor[HTML]{FFFFFF}{\color[HTML]{000000} }                         & \multicolumn{1}{c|}{\cellcolor[HTML]{FFFFFF}{\color[HTML]{000000} }}                     & {\color[HTML]{000000} run1} & {\color[HTML]{000000} 3745}            & {\color[HTML]{000000} 82}                      & {\color[HTML]{000000} \textbf{10077}}   \\ \cline{3-6} 
\cellcolor[HTML]{FFFFFF}{\color[HTML]{000000} }                         & \multicolumn{1}{c|}{\cellcolor[HTML]{FFFFFF}{\color[HTML]{000000} }}                     & {\color[HTML]{000000} run2} & {\color[HTML]{000000} 5056}            & {\color[HTML]{000000} 82}                      & {\color[HTML]{000000} \textbf{10638}}   \\ \cline{3-6} 
\multirow{-9}{*}{\cellcolor[HTML]{FFFFFF}{\color[HTML]{000000} M1, M2}} & \multicolumn{1}{c|}{\multirow{-3}{*}{\cellcolor[HTML]{FFFFFF}{\color[HTML]{000000} 3h}}} & {\color[HTML]{000000} run3} & {\color[HTML]{000000} 4888}            & {\color[HTML]{000000} 82}                      & {\color[HTML]{000000} \textbf{10839}}   \\ \hline
\cellcolor[HTML]{FFFFFF}{\color[HTML]{000000} }                         & \multicolumn{1}{c|}{\cellcolor[HTML]{FFFFFF}{\color[HTML]{000000} }}                     & {\color[HTML]{000000} run1} & {\color[HTML]{000000} 1368}            & {\color[HTML]{000000} 95}                      & {\color[HTML]{000000} \textbf{4327}}    \\ \cline{3-6} 
\cellcolor[HTML]{FFFFFF}{\color[HTML]{000000} }                         & \multicolumn{1}{c|}{\cellcolor[HTML]{FFFFFF}{\color[HTML]{000000} }}                     & {\color[HTML]{000000} run2} & {\color[HTML]{000000} 1517}            & {\color[HTML]{000000} 95}                      & {\color[HTML]{000000} \textbf{4531}}    \\ \cline{3-6} 
\cellcolor[HTML]{FFFFFF}{\color[HTML]{000000} }                         & \multicolumn{1}{c|}{\multirow{-3}{*}{\cellcolor[HTML]{FFFFFF}{\color[HTML]{000000} 1h}}} & {\color[HTML]{000000} run3} & {\color[HTML]{000000} 1702}            & {\color[HTML]{000000} 95}                      & {\color[HTML]{000000} \textbf{4161}}    \\ \cline{2-6} 
\cellcolor[HTML]{FFFFFF}{\color[HTML]{000000} }                         & \multicolumn{1}{c|}{\cellcolor[HTML]{FFFFFF}{\color[HTML]{000000} }}                     & {\color[HTML]{000000} run1} & {\color[HTML]{000000} 3221}            & {\color[HTML]{000000} 95}                      & {\color[HTML]{000000} \textbf{5607}}    \\ \cline{3-6} 
\cellcolor[HTML]{FFFFFF}{\color[HTML]{000000} }                         & \multicolumn{1}{c|}{\cellcolor[HTML]{FFFFFF}{\color[HTML]{000000} }}                     & {\color[HTML]{000000} run2} & {\color[HTML]{000000} 3227}            & {\color[HTML]{000000} 95}                      & {\color[HTML]{000000} \textbf{6980}}    \\ \cline{3-6} 
\cellcolor[HTML]{FFFFFF}{\color[HTML]{000000} }                         & \multicolumn{1}{c|}{\multirow{-3}{*}{\cellcolor[HTML]{FFFFFF}{\color[HTML]{000000} 2h}}} & {\color[HTML]{000000} run3} & {\color[HTML]{000000} 2953}            & {\color[HTML]{000000} 95}                      & {\color[HTML]{000000} \textbf{5308}}    \\ \cline{2-6} 
\cellcolor[HTML]{FFFFFF}{\color[HTML]{000000} }                         & \multicolumn{1}{c|}{\cellcolor[HTML]{FFFFFF}{\color[HTML]{000000} }}                     & {\color[HTML]{000000} run1} & {\color[HTML]{000000} 4493}            & {\color[HTML]{000000} 95}                      & {\color[HTML]{000000} \textbf{7143}}    \\ \cline{3-6} 
\cellcolor[HTML]{FFFFFF}{\color[HTML]{000000} }                         & \multicolumn{1}{c|}{\cellcolor[HTML]{FFFFFF}{\color[HTML]{000000} }}                     & {\color[HTML]{000000} run2} & {\color[HTML]{000000} 4737}            & {\color[HTML]{000000} 95}                      & {\color[HTML]{000000} \textbf{7243}}    \\ \cline{3-6} 
\multirow{-9}{*}{\cellcolor[HTML]{FFFFFF}{\color[HTML]{000000} M3, M4}} & \multicolumn{1}{c|}{\multirow{-3}{*}{\cellcolor[HTML]{FFFFFF}{\color[HTML]{000000} 3h}}} & {\color[HTML]{000000} run3} & {\color[HTML]{000000} 5089}            & {\color[HTML]{000000} 95}                      & {\color[HTML]{000000} \textbf{6880}}    \\ \hline
\cellcolor[HTML]{FFFFFF}{\color[HTML]{000000} }                         & \multicolumn{1}{c|}{\cellcolor[HTML]{FFFFFF}{\color[HTML]{000000} }}                     & {\color[HTML]{000000} run1} & {\color[HTML]{000000} 3743}            & {\color[HTML]{000000} 268}                     & {\color[HTML]{000000} \textbf{6399}}    \\ \cline{3-6} 
\cellcolor[HTML]{FFFFFF}{\color[HTML]{000000} }                         & \multicolumn{1}{c|}{\cellcolor[HTML]{FFFFFF}{\color[HTML]{000000} }}                     & {\color[HTML]{000000} run2} & {\color[HTML]{000000} 3279}            & {\color[HTML]{000000} 268}                     & {\color[HTML]{000000} \textbf{8191}}    \\ \cline{3-6} 
\cellcolor[HTML]{FFFFFF}{\color[HTML]{000000} }                         & \multicolumn{1}{c|}{\multirow{-3}{*}{\cellcolor[HTML]{FFFFFF}{\color[HTML]{000000} 1h}}} & {\color[HTML]{000000} run3} & {\color[HTML]{000000} 3044}            & {\color[HTML]{000000} 268}                     & {\color[HTML]{000000} \textbf{7600}}    \\ \cline{2-6} 
\cellcolor[HTML]{FFFFFF}{\color[HTML]{000000} }                         & \multicolumn{1}{c|}{\cellcolor[HTML]{FFFFFF}{\color[HTML]{000000} }}                     & {\color[HTML]{000000} run1} & {\color[HTML]{000000} 7837}            & {\color[HTML]{000000} 268}                     & {\color[HTML]{000000} \textbf{12560}}   \\ \cline{3-6} 
\cellcolor[HTML]{FFFFFF}{\color[HTML]{000000} }                         & \multicolumn{1}{c|}{\cellcolor[HTML]{FFFFFF}{\color[HTML]{000000} }}                     & {\color[HTML]{000000} run2} & {\color[HTML]{000000} 7023}            & {\color[HTML]{000000} 268}                     & {\color[HTML]{000000} \textbf{11911}}   \\ \cline{3-6} 
\cellcolor[HTML]{FFFFFF}{\color[HTML]{000000} }                         & \multicolumn{1}{c|}{\multirow{-3}{*}{\cellcolor[HTML]{FFFFFF}{\color[HTML]{000000} 2h}}} & {\color[HTML]{000000} run3} & {\color[HTML]{000000} 7161}            & {\color[HTML]{000000} 268}                     & {\color[HTML]{000000} \textbf{10866}}   \\ \cline{2-6} 
\cellcolor[HTML]{FFFFFF}{\color[HTML]{000000} }                         & \multicolumn{1}{c|}{\cellcolor[HTML]{FFFFFF}{\color[HTML]{000000} }}                     & {\color[HTML]{000000} run1} & {\color[HTML]{000000} 10766}           & {\color[HTML]{000000} 268}                     & {\color[HTML]{000000} \textbf{14927}}   \\ \cline{3-6} 
\cellcolor[HTML]{FFFFFF}{\color[HTML]{000000} }                         & \multicolumn{1}{c|}{\cellcolor[HTML]{FFFFFF}{\color[HTML]{000000} }}                     & {\color[HTML]{000000} run2} & {\color[HTML]{000000} 9601}            & {\color[HTML]{000000} 268}                     & {\color[HTML]{000000} \textbf{16271}}   \\ \cline{3-6} 
\multirow{-9}{*}{\cellcolor[HTML]{FFFFFF}{\color[HTML]{000000} M5, M6}} & \multicolumn{1}{c|}{\multirow{-3}{*}{\cellcolor[HTML]{FFFFFF}{\color[HTML]{000000} 3h}}} & {\color[HTML]{000000} run3} & {\color[HTML]{000000} 10226}           & {\color[HTML]{000000} 268}                     & {\color[HTML]{000000} \textbf{15430}}   \\ \hline
\cellcolor[HTML]{FFFFFF}{\color[HTML]{000000} }                         & \multicolumn{1}{c|}{\cellcolor[HTML]{FFFFFF}{\color[HTML]{000000} }}                     & {\color[HTML]{000000} run1} & {\color[HTML]{000000} 1478}            & {\color[HTML]{000000} 112}                     & {\color[HTML]{000000} \textbf{6349}}    \\ \cline{3-6} 
\cellcolor[HTML]{FFFFFF}{\color[HTML]{000000} }                         & \multicolumn{1}{c|}{\cellcolor[HTML]{FFFFFF}{\color[HTML]{000000} }}                     & {\color[HTML]{000000} run2} & {\color[HTML]{000000} 1198}            & {\color[HTML]{000000} 112}                     & {\color[HTML]{000000} \textbf{5196}}    \\ \cline{3-6} 
\cellcolor[HTML]{FFFFFF}{\color[HTML]{000000} }                         & \multicolumn{1}{c|}{\multirow{-3}{*}{\cellcolor[HTML]{FFFFFF}{\color[HTML]{000000} 1h}}} & {\color[HTML]{000000} run3} & {\color[HTML]{000000} 865}             & {\color[HTML]{000000} 112}                     & {\color[HTML]{000000} \textbf{5385}}    \\ \cline{2-6} 
\cellcolor[HTML]{FFFFFF}{\color[HTML]{000000} }                         & \multicolumn{1}{c|}{\cellcolor[HTML]{FFFFFF}{\color[HTML]{000000} }}                     & {\color[HTML]{000000} run1} & {\color[HTML]{000000} 2360}            & {\color[HTML]{000000} 112}                     & {\color[HTML]{000000} \textbf{8195}}    \\ \cline{3-6} 
\cellcolor[HTML]{FFFFFF}{\color[HTML]{000000} }                         & \multicolumn{1}{c|}{\cellcolor[HTML]{FFFFFF}{\color[HTML]{000000} }}                     & {\color[HTML]{000000} run2} & {\color[HTML]{000000} 2088}            & {\color[HTML]{000000} 112}                     & {\color[HTML]{000000} \textbf{7933}}    \\ \cline{3-6} 
\cellcolor[HTML]{FFFFFF}{\color[HTML]{000000} }                         & \multicolumn{1}{c|}{\multirow{-3}{*}{\cellcolor[HTML]{FFFFFF}{\color[HTML]{000000} 2h}}} & {\color[HTML]{000000} run3} & {\color[HTML]{000000} 2686}            & {\color[HTML]{000000} 112}                     & {\color[HTML]{000000} \textbf{8046}}    \\ \cline{2-6} 
\cellcolor[HTML]{FFFFFF}{\color[HTML]{000000} }                         & \multicolumn{1}{c|}{\cellcolor[HTML]{FFFFFF}{\color[HTML]{000000} }}                     & {\color[HTML]{000000} run1} & {\color[HTML]{000000} 3246}            & {\color[HTML]{000000} 112}                     & {\color[HTML]{000000} \textbf{11538}}   \\ \cline{3-6} 
\cellcolor[HTML]{FFFFFF}{\color[HTML]{000000} }                         & \multicolumn{1}{c|}{\cellcolor[HTML]{FFFFFF}{\color[HTML]{000000} }}                     & {\color[HTML]{000000} run2} & {\color[HTML]{000000} 4400}            & {\color[HTML]{000000} 112}                     & {\color[HTML]{000000} \textbf{13622}}   \\ \cline{3-6} 
\multirow{-9}{*}{\cellcolor[HTML]{FFFFFF}{\color[HTML]{000000} M7, M8}} & \multicolumn{1}{c|}{\multirow{-3}{*}{\cellcolor[HTML]{FFFFFF}{\color[HTML]{000000} 3h}}} & {\color[HTML]{000000} run3} & {\color[HTML]{000000} 3141}            & {\color[HTML]{000000} 112}                     & {\color[HTML]{000000} \textbf{15003}}   \\ \hline
\end{tabular}
}
\end{table}

% Cifar10
\begin{table}[t]
\caption{RQ1 results: Number of generated triggering images for models trained on the Cifar-10 dataset using DiffGAN and DRfuzz. \Revv{The \#TI (Initial) column indicates the number of triggering inputs identified in the original test dataset for each set of models.}}
\normalsize
\label{tab:Rq1-2}
\resizebox{\columnwidth}{!}{
\begin{tabular}{|
>{\columncolor[HTML]{FFFFFF}}c |
>{\columncolor[HTML]{FFFFFF}}c 
>{\columncolor[HTML]{FFFFFF}}c |
>{\columncolor[HTML]{FFFFFF}}c |
>{\columncolor[HTML]{FFFFFF}}c |
>{\columncolor[HTML]{FFFFFF}}c |}
\hline
{\color[HTML]{000000} \textbf{Models}}                                  & \multicolumn{2}{c|}{\cellcolor[HTML]{FFFFFF}{\color[HTML]{000000} \textbf{Configs}}}                                   & {\color[HTML]{000000} \textbf{DRfuzz}} & {\color[HTML]{000000} \textbf{\#TI (Initial)}} & {\color[HTML]{000000} \textbf{DiffGAN}} \\ \hline
\cellcolor[HTML]{FFFFFF}{\color[HTML]{000000} }                         & \multicolumn{1}{c|}{\cellcolor[HTML]{FFFFFF}{\color[HTML]{000000} }}                     & {\color[HTML]{000000} run1} & {\color[HTML]{000000} 36329}           & {\color[HTML]{000000} 1993}                    & {\color[HTML]{000000} \textbf{80700}}   \\ \cline{3-6} 
\cellcolor[HTML]{FFFFFF}{\color[HTML]{000000} }                         & \multicolumn{1}{c|}{\cellcolor[HTML]{FFFFFF}{\color[HTML]{000000} }}                     & {\color[HTML]{000000} run2} & {\color[HTML]{000000} 36916}           & {\color[HTML]{000000} 1993}                    & {\color[HTML]{000000} \textbf{80875}}   \\ \cline{3-6} 
\cellcolor[HTML]{FFFFFF}{\color[HTML]{000000} }                         & \multicolumn{1}{c|}{\multirow{-3}{*}{\cellcolor[HTML]{FFFFFF}{\color[HTML]{000000} 1h}}} & {\color[HTML]{000000} run3} & {\color[HTML]{000000} 30284}           & {\color[HTML]{000000} 1993}                    & {\color[HTML]{000000} \textbf{80952}}   \\ \cline{2-6} 
\cellcolor[HTML]{FFFFFF}{\color[HTML]{000000} }                         & \multicolumn{1}{c|}{\cellcolor[HTML]{FFFFFF}{\color[HTML]{000000} }}                     & {\color[HTML]{000000} run1} & {\color[HTML]{000000} 41345}           & {\color[HTML]{000000} 1993}                    & {\color[HTML]{000000} \textbf{156002}}  \\ \cline{3-6} 
\cellcolor[HTML]{FFFFFF}{\color[HTML]{000000} }                         & \multicolumn{1}{c|}{\cellcolor[HTML]{FFFFFF}{\color[HTML]{000000} }}                     & {\color[HTML]{000000} run2} & {\color[HTML]{000000} 40009}           & {\color[HTML]{000000} 1993}                    & {\color[HTML]{000000} \textbf{154942}}  \\ \cline{3-6} 
\cellcolor[HTML]{FFFFFF}{\color[HTML]{000000} }                         & \multicolumn{1}{c|}{\multirow{-3}{*}{\cellcolor[HTML]{FFFFFF}{\color[HTML]{000000} 2h}}} & {\color[HTML]{000000} run3} & {\color[HTML]{000000} 46268}           & {\color[HTML]{000000} 1993}                    & {\color[HTML]{000000} \textbf{154489}}  \\ \cline{2-6} 
\cellcolor[HTML]{FFFFFF}{\color[HTML]{000000} }                         & \multicolumn{1}{c|}{\cellcolor[HTML]{FFFFFF}{\color[HTML]{000000} }}                     & {\color[HTML]{000000} run1} & {\color[HTML]{000000} 62148}           & {\color[HTML]{000000} 1993}                    & {\color[HTML]{000000} \textbf{200924}}  \\ \cline{3-6} 
\cellcolor[HTML]{FFFFFF}{\color[HTML]{000000} }                         & \multicolumn{1}{c|}{\cellcolor[HTML]{FFFFFF}{\color[HTML]{000000} }}                     & {\color[HTML]{000000} run2} & {\color[HTML]{000000} 62456}           & {\color[HTML]{000000} 1993}                    & {\color[HTML]{000000} \textbf{195712}}  \\ \cline{3-6} 
\multirow{-9}{*}{\cellcolor[HTML]{FFFFFF}{\color[HTML]{000000} M1, M2}} & \multicolumn{1}{c|}{\multirow{-3}{*}{\cellcolor[HTML]{FFFFFF}{\color[HTML]{000000} 3h}}} & {\color[HTML]{000000} run3} & {\color[HTML]{000000} 66384}           & {\color[HTML]{000000} 1993}                    & {\color[HTML]{000000} \textbf{199785}}  \\ \hline
\cellcolor[HTML]{FFFFFF}{\color[HTML]{000000} }                         & \multicolumn{1}{c|}{\cellcolor[HTML]{FFFFFF}{\color[HTML]{000000} }}                     & {\color[HTML]{000000} run1} & {\color[HTML]{000000} 7040}            & {\color[HTML]{000000} 618}                     & {\color[HTML]{000000} \textbf{62596}}   \\ \cline{3-6} 
\cellcolor[HTML]{FFFFFF}{\color[HTML]{000000} }                         & \multicolumn{1}{c|}{\cellcolor[HTML]{FFFFFF}{\color[HTML]{000000} }}                     & {\color[HTML]{000000} run2} & {\color[HTML]{000000} 7203}                & {\color[HTML]{000000} 618}                     & {\color[HTML]{000000} \textbf{61854}}   \\ \cline{3-6} 
\cellcolor[HTML]{FFFFFF}{\color[HTML]{000000} }                         & \multicolumn{1}{c|}{\multirow{-3}{*}{\cellcolor[HTML]{FFFFFF}{\color[HTML]{000000} 1h}}} & {\color[HTML]{000000} run3} & {\color[HTML]{000000} 6913}            & {\color[HTML]{000000} 618}                     & {\color[HTML]{000000} \textbf{62609}}   \\ \cline{2-6} 
\cellcolor[HTML]{FFFFFF}{\color[HTML]{000000} }                         & \multicolumn{1}{c|}{\cellcolor[HTML]{FFFFFF}{\color[HTML]{000000} }}                     & {\color[HTML]{000000} run1} & {\color[HTML]{000000} 14725}           & {\color[HTML]{000000} 618}                     & {\color[HTML]{000000} \textbf{103446}}  \\ \cline{3-6} 
\cellcolor[HTML]{FFFFFF}{\color[HTML]{000000} }                         & \multicolumn{1}{c|}{\cellcolor[HTML]{FFFFFF}{\color[HTML]{000000} }}                     & {\color[HTML]{000000} run2} & {\color[HTML]{000000} 14159}           & {\color[HTML]{000000} 618}                     & {\color[HTML]{000000} \textbf{104297}}  \\ \cline{3-6} 
\cellcolor[HTML]{FFFFFF}{\color[HTML]{000000} }                         & \multicolumn{1}{c|}{\multirow{-3}{*}{\cellcolor[HTML]{FFFFFF}{\color[HTML]{000000} 2h}}} & {\color[HTML]{000000} run3} & {\color[HTML]{000000} 12262}           & {\color[HTML]{000000} 618}                     & {\color[HTML]{000000} \textbf{119261}}  \\ \cline{2-6} 
\cellcolor[HTML]{FFFFFF}{\color[HTML]{000000} }                         & \multicolumn{1}{c|}{\cellcolor[HTML]{FFFFFF}{\color[HTML]{000000} }}                     & {\color[HTML]{000000} run1} & {\color[HTML]{000000} 18717}           & {\color[HTML]{000000} 618}                     & {\color[HTML]{000000} \textbf{156179}}  \\ \cline{3-6} 
\cellcolor[HTML]{FFFFFF}{\color[HTML]{000000} }                         & \multicolumn{1}{c|}{\cellcolor[HTML]{FFFFFF}{\color[HTML]{000000} }}                     & {\color[HTML]{000000} run2} & {\color[HTML]{000000} 20832}           & {\color[HTML]{000000} 618}                     & {\color[HTML]{000000} \textbf{160324}}  \\ \cline{3-6} 
\multirow{-9}{*}{\cellcolor[HTML]{FFFFFF}{\color[HTML]{000000} M3, M4}} & \multicolumn{1}{c|}{\multirow{-3}{*}{\cellcolor[HTML]{FFFFFF}{\color[HTML]{000000} 3h}}} & {\color[HTML]{000000} run3} & {\color[HTML]{000000} 19915}           & {\color[HTML]{000000} 618}                     & {\color[HTML]{000000} \textbf{178621}}  \\ \hline
\cellcolor[HTML]{FFFFFF}{\color[HTML]{000000} }                         & \multicolumn{1}{c|}{\cellcolor[HTML]{FFFFFF}{\color[HTML]{000000} }}                     & {\color[HTML]{000000} run1} & {\color[HTML]{000000} 9553}            & {\color[HTML]{000000} 959}                     & {\color[HTML]{000000} \textbf{65544}}   \\ \cline{3-6} 
\cellcolor[HTML]{FFFFFF}{\color[HTML]{000000} }                         & \multicolumn{1}{c|}{\cellcolor[HTML]{FFFFFF}{\color[HTML]{000000} }}                     & {\color[HTML]{000000} run2} & {\color[HTML]{000000} 9851}            & {\color[HTML]{000000} 959}                     & {\color[HTML]{000000} \textbf{67152}}   \\ \cline{3-6} 
\cellcolor[HTML]{FFFFFF}{\color[HTML]{000000} }                         & \multicolumn{1}{c|}{\multirow{-3}{*}{\cellcolor[HTML]{FFFFFF}{\color[HTML]{000000} 1h}}} & {\color[HTML]{000000} run3} & {\color[HTML]{000000} 9605}            & {\color[HTML]{000000} 959}                     & {\color[HTML]{000000} \textbf{65544}}   \\ \cline{2-6} 
\cellcolor[HTML]{FFFFFF}{\color[HTML]{000000} }                         & \multicolumn{1}{c|}{\cellcolor[HTML]{FFFFFF}{\color[HTML]{000000} }}                     & {\color[HTML]{000000} run1} & {\color[HTML]{000000} 19248}           & {\color[HTML]{000000} 959}                     & {\color[HTML]{000000} \textbf{124322}}  \\ \cline{3-6} 
\cellcolor[HTML]{FFFFFF}{\color[HTML]{000000} }                         & \multicolumn{1}{c|}{\cellcolor[HTML]{FFFFFF}{\color[HTML]{000000} }}                     & {\color[HTML]{000000} run2} & {\color[HTML]{000000} 18352}           & {\color[HTML]{000000} 959}                     & {\color[HTML]{000000} \textbf{126532}}  \\ \cline{3-6} 
\cellcolor[HTML]{FFFFFF}{\color[HTML]{000000} }                         & \multicolumn{1}{c|}{\multirow{-3}{*}{\cellcolor[HTML]{FFFFFF}{\color[HTML]{000000} 2h}}} & {\color[HTML]{000000} run3} & {\color[HTML]{000000} 17120}           & {\color[HTML]{000000} 959}                     & {\color[HTML]{000000} \textbf{122833}}  \\ \cline{2-6} 
\cellcolor[HTML]{FFFFFF}{\color[HTML]{000000} }                         & \multicolumn{1}{c|}{\cellcolor[HTML]{FFFFFF}{\color[HTML]{000000} }}                     & {\color[HTML]{000000} run1} & {\color[HTML]{000000} 33369}           & {\color[HTML]{000000} 959}                     & {\color[HTML]{000000} \textbf{224578}}  \\ \cline{3-6} 
\cellcolor[HTML]{FFFFFF}{\color[HTML]{000000} }                         & \multicolumn{1}{c|}{\cellcolor[HTML]{FFFFFF}{\color[HTML]{000000} }}                     & {\color[HTML]{000000} run2} & {\color[HTML]{000000} 34321}           & {\color[HTML]{000000} 959}                     & {\color[HTML]{000000} \textbf{208223}}  \\ \cline{3-6} 
\multirow{-9}{*}{\cellcolor[HTML]{FFFFFF}{\color[HTML]{000000} M5, M6}} & \multicolumn{1}{c|}{\multirow{-3}{*}{\cellcolor[HTML]{FFFFFF}{\color[HTML]{000000} 3h}}} & {\color[HTML]{000000} run3} & {\color[HTML]{000000} 32499}           & {\color[HTML]{000000} 959}                     & {\color[HTML]{000000} \textbf{208522}}  \\ \hline
\cellcolor[HTML]{FFFFFF}{\color[HTML]{000000} }                         & \multicolumn{1}{c|}{\cellcolor[HTML]{FFFFFF}{\color[HTML]{000000} }}                     & {\color[HTML]{000000} run1} & {\color[HTML]{000000} 26068}           & {\color[HTML]{000000} 1750}                    & {\color[HTML]{000000} \textbf{74039}}   \\ \cline{3-6} 
\cellcolor[HTML]{FFFFFF}{\color[HTML]{000000} }                         & \multicolumn{1}{c|}{\cellcolor[HTML]{FFFFFF}{\color[HTML]{000000} }}                     & {\color[HTML]{000000} run2} & {\color[HTML]{000000} 26737}           & {\color[HTML]{000000} 1750}                    & {\color[HTML]{000000} \textbf{73702}}   \\ \cline{3-6} 
\cellcolor[HTML]{FFFFFF}{\color[HTML]{000000} }                         & \multicolumn{1}{c|}{\multirow{-3}{*}{\cellcolor[HTML]{FFFFFF}{\color[HTML]{000000} 1h}}} & {\color[HTML]{000000} run3} & {\color[HTML]{000000} 17765}           & {\color[HTML]{000000} 1750}                    & {\color[HTML]{000000} \textbf{75850}}   \\ \cline{2-6} 
\cellcolor[HTML]{FFFFFF}{\color[HTML]{000000} }                         & \multicolumn{1}{c|}{\cellcolor[HTML]{FFFFFF}{\color[HTML]{000000} }}                     & {\color[HTML]{000000} run1} & {\color[HTML]{000000} 46005}           & {\color[HTML]{000000} 1750}                    & {\color[HTML]{000000} \textbf{156716}}  \\ \cline{3-6} 
\cellcolor[HTML]{FFFFFF}{\color[HTML]{000000} }                         & \multicolumn{1}{c|}{\cellcolor[HTML]{FFFFFF}{\color[HTML]{000000} }}                     & {\color[HTML]{000000} run2} & {\color[HTML]{000000} 45730}           & {\color[HTML]{000000} 1750}                    & {\color[HTML]{000000} \textbf{133703}}  \\ \cline{3-6} 
\cellcolor[HTML]{FFFFFF}{\color[HTML]{000000} }                         & \multicolumn{1}{c|}{\multirow{-3}{*}{\cellcolor[HTML]{FFFFFF}{\color[HTML]{000000} 2h}}} & {\color[HTML]{000000} run3} & {\color[HTML]{000000} 45363}           & {\color[HTML]{000000} 1750}                    & {\color[HTML]{000000} \textbf{133569}}  \\ \cline{2-6} 
\cellcolor[HTML]{FFFFFF}{\color[HTML]{000000} }                         & \multicolumn{1}{c|}{\cellcolor[HTML]{FFFFFF}{\color[HTML]{000000} }}                     & {\color[HTML]{000000} run1} & {\color[HTML]{000000} 63434}           & {\color[HTML]{000000} 1750}                    & {\color[HTML]{000000} \textbf{243193}}  \\ \cline{3-6} 
\cellcolor[HTML]{FFFFFF}{\color[HTML]{000000} }                         & \multicolumn{1}{c|}{\cellcolor[HTML]{FFFFFF}{\color[HTML]{000000} }}                     & {\color[HTML]{000000} run2} & {\color[HTML]{000000} 63879}           & {\color[HTML]{000000} 1750}                    & {\color[HTML]{000000} \textbf{221499}}  \\ \cline{3-6} 
\multirow{-9}{*}{\cellcolor[HTML]{FFFFFF}{\color[HTML]{000000} M7, M8}} & \multicolumn{1}{c|}{\multirow{-3}{*}{\cellcolor[HTML]{FFFFFF}{\color[HTML]{000000} 3h}}} & {\color[HTML]{000000} run3} & {\color[HTML]{000000} 62148}           & {\color[HTML]{000000} 1750}                    & {\color[HTML]{000000} \textbf{218304}}  \\ \hline
\end{tabular}
}
\end{table}

\begin{tcolorbox}
\textbf{Answer to RQ1:} \textit{DiffGAN} significantly and consistently outperforms \textit{DRfuzz} in generating many more triggering inputs within a given time budget, effectively revealing about four times more behavioral disagreements between DNN models than \textit{DRfuzz}. 
\end{tcolorbox}

%\subsubsection{\textbf{RQ2. How do DiffGAN and the baseline approach compare in terms of generating valid triggering inputs?}} 

% \begin{table*}[t]  % Use table* to span across both columns
% \centering
% \caption{Validity Comparison}
% \label{tab:Validity}
% \resizebox{\textwidth}{!}{%
% \begin{tabular}{|c|c|c|c|c|c|c|c|}
% \hline
% \textbf{S} & \textbf{Method} & \textbf{Set Size} & \textbf{Av. V ratio} & \textbf{V range (95\% conf.)} & \textbf{\#V in set} & \textbf{\#Est. V in all} & \textbf{Standard Error} \\ \hline
% \multirow{2}{*}{\textbf{S1}} & DiffGAN & 379 & 0.9125 & {[}0.8819, 0.9396{]} & 346 & \textbf{8292} & 0.2895 \\ \cline{2-8} 
%  & DRfuzz & 365 & 0.8246 & {[}0.7848, 0.8635{]} & 301 & 1999 & 0.3853 \\ \hline
% \multirow{2}{*}{\textbf{S2}} & DiffGAN & 377 & 0.80 & {[}0.76, 0.8329{]} & 300 & \textbf{6555} & 0.20 \\ \cline{2-8} 
%  & DRfuzz & 364 & 0.79 & {[}0.7582, 0.82{]} & 288 & 1945 & 0.21 \\ \hline
% \multirow{2}{*}{\textbf{S3}} & DiffGAN & 381 & 0.8973 & {[}0.8687, 0.9265{]} & 342 & \textbf{10286} & 0.2946 \\ \cline{2-8} 
%  & DRfuzz & 377 & 0.8094 & {[}0.7692, 0.8488{]} & 305 & 4882 & 0.4054 \\ \hline
% \multirow{2}{*}{\textbf{S4}} & DiffGAN & 381 & 0.811 & {[}0.7717, 0.8478{]} & 309 & \textbf{6535} & 0.3616 \\ \cline{2-8} 
%  & DRfuzz & 378 & 0.7469 & {[}0.7037, 0.7910{]} & 282 & 1430 & 0.4502 \\ \hline
% \end{tabular}%
% }
% \end{table*}

\subsection{\textbf{RQ2. How do \textit{DiffGAN} and \textit{DRfuzz} compare in terms of generating valid triggering inputs within the same testing budget?}} 

% Validity Criteria for MNIST GAN-Generated Images:
% Clear Representation of a Single Digit:
% The image should represent one recognizable digit from the set {0, 1, 2, ..., 9}.
% If the image shows two or more numbers or no number, it is considered invalid.
% Blended or Ambiguous Digits:

% If the image is a 50/50 blend of two digits and cannot be clearly identified as one specific number, it is considered invalid.
% However, if 2 out of 3 reviewers agree that the image is 70-80\% likely to belong to one specific digit, the image is considered valid.
% Completeness of the Digit:
% The digit should be fully formed and clearly distinguishable.
% If the digit is incomplete or indistinguishable, the image is invalid.
% No Non-digit Images:
% If the image contains random shapes, lines, or drawings that do not represent a digit, it is considered invalid.
% Completely White or Dark Images:
% Images that are completely white or black with no visible digit are considered invalid.

%Summary of Invalid Cases:
% The image contains two numbers or no number.
% The image is a 50/50 blend and cannot be clearly identified as a specific digit.
% The digit is incomplete or a blend that cannot be attributed to one cluster by reviewer consensus.
% The image contains random shapes or non-digit drawings.
% The image is completely white or dark, with no visible content.

% Input Generation (3 Runs for Each Approach):

To systematically compare \textit{DiffGAN} and \textit{DRfuzz} in terms of generating valid triggering inputs within a given time budget, we considered the inputs generated in RQ1 by both methods. \Revv{More specifically, we considered all four model pairs using the MNIST dataset and two model pairs for the CIFAR-10 dataset.  We ran both approaches for two hours on these models. }
%We chose to focus on MNIST in this RQ due to how resource-intensive is the evaluation process. Indeed,
We did not consider all model pairs in this RQ, as the validity assessments require significant manual effort, particularly for labeling generated inputs to ensure they are contextually valid. The lower resolution and greater complexity of CIFAR-10 images further complicate this process, as accurately labeling and validating inputs in CIFAR-10 is considerably more challenging compared to MNIST. Given these challenges, we opted to limit our validity study to the MNIST dataset to ensure its feasibility.

As mentioned in RQ1, we ran each method three times to minimize randomness and recorded all images generated by each approach during each run. After completing the runs, for each approach and model pair, we aggregated the images generated from all three runs into a comprehensive dataset. By combining the results from three separate runs for each configuration, we were able to mitigate the effects of randomness in the generation process, thereby increasing the reliability of our comparisons. 
%This approach ensures that any variability in the outputs is accounted for, providing a more accurate assessment of each method's performance.

\begin{table}[]
\centering
\caption{MNIST Improvement}
\label{tab:imp}
\begin{tabular}{|c|c|c|c|}
\hline
\textbf{Models} & \textbf{Time} & \textbf{DRfuzz   Imp (\%)} & \textbf{DiffGAN   Imp (\%)} \\ \hline
\multirow{3}{*}{M1, M2} & 1h & 1401.62 & 6030.08 \\ \cline{2-4} 
 & 2h & 3425.61 & 9029.67 \\ \cline{2-4} 
 & 3h & 5465.85 & 12722 \\ \hline
\multirow{3}{*}{M3, M4} & 1h & 1610.87 & 4462.81 \\ \cline{2-4} 
 & 2h & 3103.16 & 6466.66 \\ \cline{2-4} 
 & 3h & 4655.79 & 7095.44 \\ \hline
\multirow{3}{*}{M5, M6} & 1h & 1679.29 & 3037.69 \\ \cline{2-4} 
 & 2h & 3494.03 & 5388.33 \\ \cline{2-4} 
 & 3h & 2770 & 4508.1 \\ \hline
\multirow{3}{*}{M7, M8} & 1h & 898.21 & 4871.43 \\ \cline{2-4} 
 & 2h & 2291.07 & 8100.3 \\ \cline{2-4} 
 & 3h & 3080.65 & 12990.18 \\ \hline
\end{tabular}
\end{table}

\begin{table}[]
\centering
\caption{Cifar-10 Improvement}
\label{tab:imp2}
\begin{tabular}{|c|c|c|c|}
\hline
\textbf{Models} & \textbf{Time} & \textbf{DRfuzz   Imp (\%)} & \textbf{DiffGAN   Imp (\%)} \\ \hline
\multirow{3}{*}{M1, M2} & 1h & 1631.62 & 3959.07 \\ \cline{2-4} 
 & 2h & 2031.51 & 7559.31 \\ \cline{2-4} 
 & 3h & 3053.63 & 10094.88 \\ \hline
\multirow{3}{*}{M3, M4} & 1h & 1054.86 & 10055.92 \\ \cline{2-4} 
 & 2h & 2120.68 & 17201.02 \\ \cline{2-4} 
 & 3h & 3106.35 & 23296.37 \\ \hline
\multirow{3}{*}{M5, M6} & 1h & 914.39 & 6959.54 \\ \cline{2-4} 
 & 2h & 2409.05 & 16357.15 \\ \cline{2-4} 
 & 3h & 3390.65 & 23611.05 \\ \hline
\multirow{3}{*}{M7, M8} & 1h & 1254.76 & 4195.18 \\ \cline{2-4} 
 & 2h & 35699.33 & 147930 \\ \cline{2-4} 
 & 3h & 63013.5 & 219901.5 \\ \hline
\end{tabular}
\end{table}
\noindent \textbf{\textit{Removing redundant inputs.}} To eliminate redundancy from the aggregated dataset, we employed a simple hashing-based technique that focused on the exact byte-level representation of each image. Specifically, we converted each image into its byte form and computed a hash value using Python's built-in hash() function\footnote{Python documentation, \texttt{hash()} function, available at: \url{https://docs.python.org/3/library/functions.html##hash}}. By comparing these hash values, we identified and removed duplicate images that were identical at the byte level. This method ensured that only unique images---in terms of their pixel data---were retained for further analysis.

\noindent \textbf{\textit{Selecting representative samples.}} Due to the large number of triggering inputs generated by \textit{DiffGAN} and \textit{DRfuzz}, we selected a sample of triggering inputs from each aggregated dataset. For each setting, we determined the sample size that can confidently represent the whole corresponding dataset with a 95\% confidence level and a margin of error of 5\%\footnote{As computed from https://www.calculator.net/sample-size-calculator.html}. \Revv{As a result, in our experiments, the sample size varied from 364 to 384 across the six settings that we considered, as shown in Tables~\ref{tab:Validity} and~\ref{tab:Validity2}. } This approach helped choose a minimal subset that was large enough to provide results that accurately reflected the overall distribution of inputs generated by \textit{DiffGAN} and \textit{DRfuzz}~\cite{singh2014sampling}.

\begin{figure}[ht]
    \centering
    % First row - Valid images
    \begin{subfigure}[b]{0.48\textwidth}
        \centering
        \includegraphics[width=\textwidth]{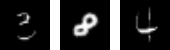}
        \caption{Valid Images - DRfuzz}
    \end{subfigure}
    \hfill
    \vspace{0.05cm}  
    \begin{subfigure}[b]{0.48\textwidth}
        \centering
        \includegraphics[width=\textwidth]{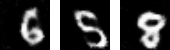}
        \caption{Valid Images - DiffGAN}
    \end{subfigure}
    
    \vspace{0.5cm} % Add space between rows
    
    % Second row - Invalid images
    \begin{subfigure}[b]{0.48\textwidth}
        \centering
        \includegraphics[width=\textwidth]{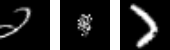}
        \caption{Invalid Images - DRfuzz}
    \end{subfigure}
    \hfill
    \vspace{0.05cm}  
    \begin{subfigure}[b]{0.48\textwidth}
        \centering
        \includegraphics[width=\textwidth]{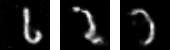}
        \caption{Invalid Images - DiffGAN}
    \end{subfigure}
    
    \caption{Examples of Valid and Invalid Images for DRfuzz and DiffGAN}
    \label{fig:valid_invalid_comparison}
\end{figure}

\noindent \textbf{\textit{Manual validity checking.}} From the non-redundant set, we randomly sampled images for validation, according to the sample size determined in the previous step. To ensure unbiased evaluation, we shuffled images generated by both \textit{DiffGAN} and \textit{DRfuzz} and presented them anonymously to three independent expert evaluators. These experts were tasked with assessing the validity of each image based on predefined criteria. Specifically, they labeled images as either valid or invalid. For example for MNIST dataset, invalid images were defined as those that displayed two numbers, no number, or other issues such as (1) Images that were a 50/50 blend of two digits and could not be clearly identified as a specific digit, (2) Incomplete or blended digits that could not be classified into one category by consensus, (3) Images containing random shapes or non-digit drawings, or (4) Images that were completely white or dark, with no discernible content. Examples of valid and invalid inputs generated by both approaches are depicted in Figure~\ref{fig:valid_invalid_comparison}. To determine the validity of each image, we considered the agreement among the evaluators. If at least two out of three evaluators labeled an image as valid, we considered it valid. To quantify the level of agreement between the evaluators, we employ Cohen's kappa~\cite{banerjee1999beyond}---a statistical measure that accounts for agreement occurring by chance. In our study, the value of Cohen's kappa is 70\%,  indicating a high level of agreement among the evaluators and underscoring the reliability of the manual labeling process.

% Bootstrapping:

\noindent \textbf{\textit{Bootstrapping.}} After completing the labeling process, we applied bootstrapping to estimate the probability distribution of the validity ratio of the entire generated inputs. Bootstrapping is a nonparametric, simulation-based technique that allows statistical inference from small samples~\cite{Mooney1993-MOOBAN}. By resampling the labeled subset (with replacement) multiple times, this method estimates the probability distribution of the validity ratio of the entire dataset of generated inputs. More specifically, we resampled the labeled subset 1000 times with replacement, calculating the validity ratio for each resample. The resulting distribution of validity ratios provided a more reliable estimate of the dataset's overall validity, accounting for variability and reducing potential bias from relying on a single sample. This resampling approach allowed us to estimate the confidence interval for the percentage of valid inputs across the entire dataset.

\noindent \textbf{\textit{Reporting the total number of valid inputs.}} We calculated the proportion of valid images for each method by dividing the number of valid images by the total number of sampled images. 
We then multiplied the validity ratio, calculated from the subset, by the average number of images generated in the runs to estimate the expected number of valid images in each run. Additionally, we determined the 95\% confidence interval as reported in Table~\ref{tab:Validity}.

% We then multiplied this validity ratio calculate from  the subset by the average number of images generated in the runs to estimate the expected number of valid images in the , the 95\% confidence interval, and the number of valid images, as reported in Table~\ref{tab:Validity}.

Finally, we compared the expected number of valid inputs generated by \textit{DiffGAN} and \textit{DRfuzz} to assess which method was more effective at producing valid triggering inputs within the same time budget of two hours. \Revv{The results, presented in Tables~\ref{tab:Validity} and~\ref{tab:Validity2} , clearly suggest that \textit{DiffGAN} significantly and consistently outperforms \textit{DRfuzz} in generating more valid triggering inputs for all the model pairs and datasets we considered in our experiments.} For instance, in the first setting, it is estimated that \textit{DiffGAN} generated 8,292 valid triggering inputs, whereas \textit{DRfuzz} produced only 1,999 such inputs. This corresponds to a validity ratio of 91.25\% for \textit{DiffGAN} compared to 82.46\% for \textit{DRfuzz}. %Importantly, those results were consistent across all the settings we considered. 
To further analyze the statistical significance of our results we performed a statistical analysis using the Wilcoxon signed-rank test~\cite{macfarland2016wilcoxon}, with a significance level of $\alpha = 0.05$, to assess whether \textit{DiffGAN} significantly outperforms \textit{DRfuzz} in generating valid triggering inputs.  
The Wilcoxon signed-rank test, a non-parametric method, is used to compare the medians of continuous variables for paired samples and is ideal for our case since it does not require any assumptions about the data distribution.
%We applied the paired test to compare the effectiveness of \textit{DiffGAN} and \textit{DRfuzz} under the same datasets and experimental conditions. 
Instead of considering the number of valid inputs, we relied on the validity ratio due to differences in sample sizes. We evaluated the validity ratios from 4000 samples for \textit{DiffGAN} and 4000 samples for \textit{DRfuzz}, selected via bootstrapping across four different scenarios (1000 samples per scenario). The Wilcoxon signed-rank test, with \textit{w-statistic} = 343465.5 and a \textit{p-value} of 0.000001, confirmed that \textit{DiffGAN} significantly outperformed \textit{DRfuzz} at generating valid triggering inputs across all scenarios.

% We applied a paired test because we compared the effectiveness of DiffGAN and DRfuzz across the same datasets and experimental settings. For the statistical tests, we collected paired data on the number of valid inputs generated by \textit{DiffGAN} and \textit{DRfuzz} from the 4000 samples selected through bootstrapping across four different settings (1000 samples per setting). The analysis revealed that all p-values were below 0.05, indicating that \textit{DiffGAN} significantly outperformed \textit{DRfuzz} in generating more valid triggering inputs across all settings.

%We should note that \textit{DiffGAN} outperforms \textit{DRfuzz} in generating more valid triggering inputs primarily due to validation mechanisms it employs throughout the input generation process. Unlike \textit{DRfuzz}, which relies solely on the GAN discriminator to determine the validity of generated inputs, \textit{DiffGAN} incorporates both the discriminator and a structural similarity index (SSIM) score for more robust filtering of invalid inputs.  Additionally, \textit{DiffGAN} leverages the generative capabilities of GANs, which are specifically designed to create realistic and valid inputs.  On the other hand, \textit{DRfuzz} applies multiple mutation rules to the same seed image, which can result in invalid inputs.

\begin{table*}[t]  % Use table* to span across both columns
\centering
\caption{Validity Comparison MNIST}
\label{tab:Validity}
\resizebox{\textwidth}{!}{%
\begin{tabular}{|c|c|c|c|c|c|}
\hline
\textbf{Models} & \textbf{Method} & \textbf{Set Size} & \textbf{Av. Validity ratio} & \textbf{Validity range (95\% conf.)} & \textbf{\#Est. Valid inputs} \\ \hline
\multirow{2}{*}{\textbf{M1, M2}} & DiffGAN & 379 & 0.9125 & {[}0.8819, 0.9396{]} & \textbf{8292} \\ \cline{2-6} 
 & DRfuzz & 365 & 0.8246 & {[}0.7848, 0.8635{]} & 1999 \\ \hline
\multirow{2}{*}{\textbf{M3, M4}} & DiffGAN & 377 & 0.80 & {[}0.76, 0.8329{]} & \textbf{6555} \\ \cline{2-6} 
 & DRfuzz & 364 & 0.79 & {[}0.7582, 0.82{]} & 1945 \\ \hline
\multirow{2}{*}{\textbf{M5, M6}} & DiffGAN & 381 & 0.8973 & {[}0.8687, 0.9265{]} & \textbf{10286} \\ \cline{2-6} 
 & DRfuzz & 377 & 0.8094 & {[}0.7692, 0.8488{]} & 4882 \\ \hline
\multirow{2}{*}{\textbf{M7, M8}} & DiffGAN & 381 & 0.811 & {[}0.7717, 0.8478{]} & \textbf{6535} \\ \cline{2-6} 
 & DRfuzz & 378 & 0.7469 & {[}0.7037, 0.7910{]} & 1430 \\ \hline
\end{tabular}%
}
\end{table*}

\begin{table*}[t]  % Use table* to span across both columns
\centering
\caption{ Validity Comparison Cifar-10}
\label{tab:Validity2}
\resizebox{\textwidth}{!}{%
\begin{tabular}{|c|c|c|c|c|c|}
\hline
\textbf{Models} & \textbf{Method} & \textbf{Set Size} & \textbf{Av. Validity ratio} & \textbf{Validity range (95\% conf.)} & \textbf{\#Est. Valid inputs} \\ \hline
\multirow{2}{*}{\textbf{M1, M2}} & DiffGAN & 384 & 0.78 & {[}0.7353, 0.8196{]} & \textbf{113,393} \\ \cline{2-6} 
 & DRfuzz & 382 & 0.67 & {[}0.6200, 0.7111{]} & 31,422 \\ \hline
\multirow{2}{*}{\textbf{M3, M4}} & DiffGAN & 384 & 0.71 & {[}0.6619, 0.7531{]} & \textbf{79,613} \\ \cline{2-6} 
 & DRfuzz & 378 & 0.58 & {[}0.5293, 0.6286{]} & 7,782 \\ \hline
\end{tabular}%
}
\end{table*}

It is important to note that \textit{DiffGAN} outperforms \textit{DRfuzz} in generating valid triggering inputs primarily due to the validation mechanisms it employs throughout the input generation process. While \textit{DRfuzz} relies solely on the GAN discriminator to assess the validity of generated inputs, \textit{DiffGAN} incorporates both the generator and the SSIM score, enabling a more effective and comprehensive filtering of invalid inputs. Furthermore, \textit{DiffGAN} leverages the generative capabilities of GANs, which are specifically designed to produce realistic and diverse inputs, enhancing its ability to generate valid samples. In contrast, \textit{DRfuzz} applies multiple mutation rules to the same seed image, which can often lead to the generation of more invalid inputs.

\begin{tcolorbox}
\textbf{Answer to RQ2:} \textit{DiffGAN} consistently and significantly outperforms \textit{DRfuzz} in generating more valid triggering inputs within the same time budget across all experimental settings.
\end{tcolorbox}

\subsection{\textbf{RQ3. Does \textit{DiffGAN} find more diverse triggering inputs than \textit{DRfuzz}?}} \label{sec:RQ3_res}

To answer this research question, we measured the diversity scores of the generated test input sets for \textit{DiffGAN} and \textit{DRfuzz}. We selected well-established diversity metrics that are directly applicable to input sets to measure their diversity scores. Specifically, we used two SOTA metrics: \textit{Geometric Diversity}~\cite{aghababaeyan2021black} and \textit{Shannon} diversity~\cite{6773024,119732}, each leveraging different types of input features. 

%To measure the \textit{Geometric Diversity} score of an input set, we leverage VGG-16 to extract input features while relying on pixel values as features for the \textit{Shannon} diversity metric, in line with several previous studies~\cite{9065274, 119732, garcia2015differential, khorramshahi2020gansvariationalentropyregularizers}. Furthermore, we should note that our prior work~\cite{aghababaeyan2021black, Aghababaeyan2024} has demonstrated the effectiveness of combining the \textit{Geometric Diversity} metric with VGG-16 features in accurately capturing and measuring the diversity of input sets.

\textbf{\textit{Shannon Diversity}}, also known as the \textit{Shannon-Wiener Index} or \textit{Shannon Entropy}, is a SOTA metric used to quantify diversity in a dataset~\cite{9065274, 119732, garcia2015differential, khorramshahi2020gansvariationalentropyregularizers}. %Originally rooted in information theory, which measures the unpredictability or information content of a system, this index has been widely adopted in various fields, including ecology and image analysis, to assess diversity and complexity~\cite{6773024,119732}.
It is calculated as follows:

\begin{equation}
H = -\sum_{i=1}^{S} p_i \log_b(p_i)
\end{equation}

where:
\begin{itemize}
    \item $S$ is the total number of distinct categories (taxa, species, or classes).
    \item $p_i$ is the proportion of the dataset represented by the $i$-th category.
    \item $b$ is the base of the logarithm, which is $e$ (natural logarithm).
\end{itemize}

The metric computes the sum of the proportions of each category multiplied by the logarithm of those proportions. The negative sign ensures that the index is positive, and higher diversity translates into higher values of $H$. To calculate the Shannon diversity for a set of images, each image is first flattened into a one-dimensional array of pixel values. These flattened arrays can then be combined into a single array that represents the distribution of pixel values across the entire image set. The Shannon score, also known as entropy, is then calculated using this distribution, which quantifies the diversity of pixel intensities within the dataset. A higher Shannon score indicates greater diversity within the set of images.
%Shannon’s Diversity Index provides a way to quantify the uncertainty or unpredictability of the data's distribution. In a dataset with a large number of categories that are evenly represented, the index will be high, reflecting a high level of diversity. In contrast, if the dataset is dominated by a few categories, the index will be lower, indicating less diversity.
%The core idea is to measure how evenly the elements in the dataset are spread across the different categories. In the context of image analysis, this could be related to the variety of colors, textures, or patterns within a set of images. A higher Shannon index would imply that the images are diverse in content, while a lower index would suggest uniformity or redundancy.

%Benefits of Using Shannon’s Diversity Index

%\begin{itemize}
%    \item \textbf{Comprehensive Diversity Measure}: Shannon’s Diversity Index accounts for both the richness (number of categories) and the evenness (how evenly distributed these categories are) of the dataset. 
    
%    \item \textbf{Applicability Across Domains}: While originally used in ecology, this index is versatile and can be applied to any field where diversity needs to be measured, including image analysis, text mining, and bioinformatics.
    
%    \item \textbf{Logarithmic Sensitivity}: The logarithmic component of the index means that it is sensitive to changes in the distribution of categories, capturing subtle shifts in diversity.
%\end{itemize}

In some contexts, computing the exponential of the Shannon diversity score is used to provide a more intuitive interpretation of the index. Specifically, by calculating $e^H$, where $H$ is the Shannon diversity of a set of images, we can derive the \textit{effective number of image categories} within the set. This value represents the number of equally likely categories that would produce the same level of entropy as observed in the dataset. 
%For example, if $H$ is 11, computing $e^{11}$ (approximately 59,874.14) means a high degree of diversity. Similarly, if $H$ increases a little to 12, the exponential value $e^{12}$ is substantially larger (approximately 162,754.79), indicating that the dataset is considerably more diverse. 
For example, if $H$ increases from 11 to 12, the exponential value $e^{12}$ is substantially larger (approximately 162,754.79) when compared to  $e^{11}$ = 59,874.14, indicating that the dataset is considerably more diverse. 
This approach allows for the translation of entropy into a more intuitive measure of diversity, making it easier to understand and compare the diversity of different datasets.

\textbf{Geometric Diversity} is a widely used metric for assessing the diversity of test input sets~\cite{aghababaeyan2021black, kulesza2012determinantal, gong2014diverse}. To calculate the \textit{Geometric Diversity} score of an input set, we use features extracted by the VGG16 model. Our previous work~\cite{aghababaeyan2021black, Aghababaeyan2024} has shown that combining the \textit{Geometric Diversity} metric with VGG16 features is highly effective in accurately capturing and quantifying the diversity of input sets.

The \textit{Geometric Diversity} metric, denoted as \( G(S) \), is computed as follows: Given a dataset \( X \) and its feature vectors \( V \), the geometric diversity of a subset \( S \subseteq X \) is defined by:

\begin{equation}
G(S) = \det(V_s \times V_s^{T})
\end{equation}

% \textit{\begin{equation}\label{Eq:GeometricDiversity}
%  GD(S) = det(F_S * {F^\prime_S}^{T} )  
% \end{equation}}

This formula represents the squared volume of the parallelepiped formed by the rows of \( V_s \), where \( V_s \) consists of the feature vectors corresponding to the subset \( S \). A larger volume indicates greater diversity within the subset \( S \) in the feature space.

% \textbf{Geometric Diversity} metric is a widely used metric to measure the diversity of test input sets ~\cite{aghababaeyan2021black, kulesza2012determinantal, gong2014diverse}. To measure the \textit{Geometric Diversity} score of an input set, we leverage VGG-16 to extract input features. We should note that our prior work~\cite{aghababaeyan2021black, Aghababaeyan2024} has demonstrated the effectiveness of combining the \textit{Geometric Diversity} metric with VGG-16 features in accurately capturing and measuring the diversity of input sets.

% The geometric diversity \textit{G(.)} is computed as follows. Given a dataset $X$ and feature vectors $V$, the geometric diversity of a subset $S {\subseteq} X $ is defined as: 

% \begin{equation}\label{Eq:GeometricDiversity}
%  G(S) = det(Vs * Vs^{T} )  
% \end{equation}

% which corresponds to the squared volume of the parallelepiped spanned by the rows of \textit{Vs}, since they correspond to vectors in the feature space. The larger the volume, the more diverse $S$ is in the feature space.

% Shannon’s Diversity Index is a powerful tool for quantifying the diversity within a dataset. It captures both the richness and evenness of the data, providing insights into the complexity and variety present. Whether analyzing ecological data, image datasets, or other complex systems, this metric offers a robust and intuitive measure of diversity, allowing for a deeper understanding of the underlying data patterns.

In this research question, we employed both the Shannon and Geometric diversity metrics to analyze the dataset, aiming to evaluate image diversity from two complementary perspectives. The Shannon diversity metric offers a probabilistic view, emphasizing the distribution of pixel values. In contrast, the Geometric diversity metric assesses diversity based on spatial relationships and geometric patterns within the image features extracted by VGG16. By integrating these two metrics, we achieve a more comprehensive understanding of the datasets' diversity.

%We should note that in this \textit{RQ}, we employed both the Shannon and Geometric diversity metrics to analyze the dataset, aiming to examine the diversity of images from two complementary perspectives. While the Shannon diversity metric  provides a probabilistic view of diversity, focusing on the distribution of pixel values, the Geometric diversity assesses diversity based on spatial relationships and geometric patterns within image features extracted by VGG16. By considering these two metrics, we obtain a more comprehensive understanding of the dataset's diversity.

To measure the diversity scores of the generated test input sets for \textit{DiffGAN} and \textit{DRfuzz}, we used the same subjects and generated triggering inputs as in RQ1. Given that \textit{Geometric Diversity} scores cannot be normalized based on subset size~\cite{aghababaeyan2021black}, we ensured a fair comparison by randomly selecting subsets of equal size for both metrics. Specifically, for each approach, subject, and testing budget, we randomly selected 1,000 unique, generated triggering inputs and calculated their \textit{Shannon} and \textit{Geometric Diversity} scores. To account for the inherent randomness in \textit{DiffGAN} and \textit{DRfuzz}, we repeated this process 30 times and reported the average diversity scores.

The results, as shown in Tables~\ref{tab:MNIST_Diversity} and~\ref{tab:CIFAR10_Diversity}, consistently demonstrate that \textit{DiffGAN} outperforms \textit{DRfuzz} in generating more diverse triggering inputs, when considering both metrics, highlighting its effectiveness in exploring a broader range of input variations. 

Similarly to RQ1, we analysed the variability of our diversity results using the CV score. %which measures the relative variability of the outputs compared to the mean. 
Our CV analysis shows that both \textit{DiffGAN} and \textit{DRfuzz} exhibit low variability across different runs, with \textit{DiffGAN} demonstrating greater consistency. \textit{DRfuzz}'s GD diversity has a CV range of 6.76\% to 18.44\%, while \textit{DiffGAN} is more stable with a CV range of 2.24\% to 14.23\%. In terms of Shannon metrics, \textit{DRfuzz} shows a CV range between 0.09\% and 0.70\%, whereas \textit{DiffGAN} is even more consistent, with a CV range of 0.02\% to 0.33\%. These results underscore the superior stability of \textit{DiffGAN}'s outputs compared to \textit{DRfuzz}. %It is worth noting that the relatively low variability in Shannon diversity scores is due to the logarithmic nature of the metric.  In this work, we report the Shannon diversity score itself, not its exponential form as reported in some prior studies~\cite{REFS} (which is equivalent to the Shannon diversity but exhibits higher scores).

Furthermore,  the superior performance of \textit{DiffGAN} in generating more diverse triggering inputs can be attributed to its large latent space, which allows for a greater variety of generated inputs, and its diversity-based fitness function specifically designed to maximize the diversity of generated inputs, a feature that is limited in \textit{DRfuzz}. Indeed, \textit{DRfuzz} focuses on triggering behavioral disagreements through limited mutation rules, which, while effective in some contexts, do not inherently prioritize or ensure diversity. As a result, the input sets generated by \textit{DRfuzz} tend to be less diverse compared to \textit{DiffGAN}.

\begin{table*}[t]
% \begin{table*}[!h]
% \tiny
\caption{Diversity scores of MNIST}
\label{tab:MNIST_Diversity}
\resizebox{\textwidth}{!}{%
\begin{tabular}{|c|c|cc|cc|}
\hline
\textbf{}                  & \textbf{} & \multicolumn{2}{c|}{\textbf{DRfuzz}}              & \multicolumn{2}{c|}{\textbf{DiffGAN}}                      \\ \hline
                   Models        &     Configs      & \multicolumn{1}{c|}{GD}                & Exponential Shannon ($e^{H}$)    & \multicolumn{1}{c|}{GD}                & Exponential Shannon   ($e^{H}$)       \\ \hline
\multirow{3}{*}{M1, M2} & 1h        & \multicolumn{1}{c|}{422.03}          & $e^{11.36} = 86,361.86$ & \multicolumn{1}{c|}{\textbf{530.62}} & \textbf{$e^{12.10} = 178,632.90$} \\ \cline{2-6} 
                           & 2h        & \multicolumn{1}{c|}{481.24}           & $e^{11.34} = 84,643.49$ & \multicolumn{1}{c|}{\textbf{536.05}} & \textbf{$e^{12.16} = 191,548.60$} \\ \cline{2-6} 
                           & 3h        & \multicolumn{1}{c|}{318.04} & $e^{11.32} = 82,829.03$ & \multicolumn{1}{c|}{\textbf{366.21}}          & \textbf{$e^{12.10} = 178,497.79$}  \\ \hline
\multirow{3}{*}{M3, M4} & 1h        & \multicolumn{1}{c|}{438.42}           & $e^{11.39} = 89,882.98$ & \multicolumn{1}{c|}{\textbf{727.20}}  & \textbf{$e^{12.21} = 200,195.46$} \\ \cline{2-6} 
                           & 2h        & \multicolumn{1}{c|}{592.18}          & $e^{11.39} = 89,238.14$  & \multicolumn{1}{c|}{\textbf{665.31}} & \textbf{$e^{12.27} = 213,472.83$} \\ \cline{2-6} 
                           & 3h        & \multicolumn{1}{c|}{645.69}          & $e^{11.38} = 88,908.85$ & \multicolumn{1}{c|}{\textbf{661.64}} & \textbf{$e^{12.23} = 204,636.72$} \\ \hline
\multirow{3}{*}{M5, M6} & 1h        & \multicolumn{1}{c|}{410.67}          & $e^{11.50} = 98,610.81$ & \multicolumn{1}{c|}{\textbf{712.76}} & \textbf{$e^{12.27} = 213,402.74$} \\ \cline{2-6} 
                           & 2h        & \multicolumn{1}{c|}{536.47}          & $e^{11.43} = 92,631.60$ & \multicolumn{1}{c|}{\textbf{649.41}} & \textbf{$e^{12.23} = 204,753.74$} \\ \cline{2-6} 
                           & 3h        & \multicolumn{1}{c|}{551.06}          & $e^{11.42} = 91,634.01$ & \multicolumn{1}{c|}{\textbf{578.24}} & \textbf{$e^{12.26} = 211,800.43$} \\ \hline
\multirow{3}{*}{M7, M8} & 1h        & \multicolumn{1}{c|}{116.72}          & $e^{11.40} = 90,159.89$ & \multicolumn{1}{c|}{\textbf{481.94}} & \textbf{$e^{12.10} = 178,751.64$}  \\ \cline{2-6} 
                           & 2h        & \multicolumn{1}{c|}{244.65}            & $e^{11.40} = 90,792.41$ & \multicolumn{1}{c|}{\textbf{551.06}} & \textbf{$e^{12.12} = 183,319.27$} \\ \cline{2-6} 
                           & 3h        & \multicolumn{1}{c|}{231.87}          & $e^{11.38} = 87,505.95$ & \multicolumn{1}{c|}{\textbf{666.11}}  & \textbf{$e^{12.14} = 187,734.83$}  \\ \hline
\end{tabular}%
}
\end{table*}

\begin{table*}[t]
\caption{Diversity scores of Cifar-10 }
\label{tab:CIFAR10_Diversity}
\resizebox{\textwidth}{!}{%
\begin{tabular}{|c|c|cc|cc|}
\hline
\textbf{}               & \textbf{} & \multicolumn{2}{c|}{\textbf{DRfuzz}}     & \multicolumn{2}{c|}{\textbf{DiffGAN}}                      \\ \hline
Models                  & Configs   & \multicolumn{1}{c|}{GD}       & Exponential Shannon ($e^{H}$)     & \multicolumn{1}{c|}{GD}                & Exponential Shannon ($e^{H}$)            \\ \hline
\multirow{3}{*}{M1, M2} & 1h        & \multicolumn{1}{c|}{1790.80} & $e^{14.71} = 2,435,217.87$    & \multicolumn{1}{c|}{\textbf{2396.99}} & \textbf{$e^{14.84} = 2,774,122.83$} \\ \cline{2-6} 
                        & 2h        & \multicolumn{1}{c|}{2108.90} & $e^{14.71} = 2,432,731.99$ & \multicolumn{1}{c|}{\textbf{2333.51}} & \textbf{$e^{14.83} = 2,772,404.84$} \\ \cline{2-6} 
                        & 3h        & \multicolumn{1}{c|}{2318.33} & $e^{14.70} = 2,430,965.85$ & \multicolumn{1}{c|}{\textbf{2457.05}} & \textbf{$e^{14.94} = 3,073,279.80$} \\ \hline
\multirow{3}{*}{M3, M4} & 1h        & \multicolumn{1}{c|}{976.64} & $e^{14.71} = 2,434,515.58$ & \multicolumn{1}{c|}{\textbf{2395.81}} & \textbf{$e^{14.84} = 2,774,819.14$} \\ \cline{2-6} 
                        & 2h        & \multicolumn{1}{c|}{1117.67} & $e^{14.71} = 2,432,366.04$ & \multicolumn{1}{c|}{\textbf{2339.19}} & \textbf{$e^{14.83} = 2,772,679.02$} \\ \cline{2-6} 
                        & 3h        & \multicolumn{1}{c|}{1466.08} & $e^{14.70} = 2,429,166.95$ & \multicolumn{1}{c|}{\textbf{2284.04}} & \textbf{$e^{14.83} = 2,772,395.01$}  \\ \hline
\multirow{3}{*}{M5, M6} & 1h        & \multicolumn{1}{c|}{711.47} & $e^{14.73} = 2,453,279.49$ & \multicolumn{1}{c|}{\textbf{2387.32}} & \textbf{$e^{14.84} = 2,774,233.41$} \\ \cline{2-6} 
                        & 2h        & \multicolumn{1}{c|}{1463.54} & $e^{14.71} = 2,432,146.96$ & \multicolumn{1}{c|}{\textbf{2334.41}}  & \textbf{$e^{14.83} = 2,772,466.97$} \\ \cline{2-6} 
                        & 3h        & \multicolumn{1}{c|}{2045.84} & $e^{14.70} = 2,430,669.35$ & \multicolumn{1}{c|}{\textbf{2283.65}} & \textbf{$e^{14.83} = 2,772,148.43$} \\ \hline
\multirow{3}{*}{M7, M8} & 1h        & \multicolumn{1}{c|}{1893.78} & $e^{14.71} = 2,432,859.74$ & \multicolumn{1}{c|}{\textbf{2378.10}} & \textbf{$e^{14.84} = 2,774,307.47$} \\ \cline{2-6} 
                        & 2h        & \multicolumn{1}{c|}{2199.47} & $e^{14.69} = 2,426,034.24$ & \multicolumn{1}{c|}{\textbf{2331.51}} & \textbf{$e^{14.83} = 2,772,278.57$} \\ \cline{2-6} 
                        & 3h        & \multicolumn{1}{c|}{2183.13} & $e^{14.69} = 2,425,208.35$ & \multicolumn{1}{c|}{\textbf{2295.15}} & \textbf{$e^{14.83} = 2,771,688.20$} \\ \hline
\end{tabular}%
}
\end{table*}

\begin{tcolorbox}
\textbf{Answer to RQ3:} \textit{DiffGAN} consistently outperforms \textit{DRfuzz} in generating more diverse triggering inputs across all DNN models and datasets.
\end{tcolorbox}

\begin{table*}[t]

\centering
\caption{Categorization of Model Behavior on 1,000 Triggering Inputs per Model Pair (Counts with Percentages)}
\label{tab:model_behavior}
\begin{tabular}{|c|c|c|c|c|}
\hline
\textbf{Models} & \textbf{Method} & \makecell{\textbf{Only Model A is correct}} & \makecell{\textbf{Only Model B is correct}} & \makecell{\textbf{Both models are incorrect}} \\ \hline
\multirow{2}{*}{M1, M2} 
& DRfuzz  & 282 (28.2\%) & 480 (48.0\%) & 238 (23.8\%)  \\
& DiffGAN & 121 (12.1\%) & 742 (74.2\%) & 137 (13.7\%)  \\ \hline

\multirow{2}{*}{M3, M4} 
& DRfuzz  & 294 (29.4\%) & 586 (58.6\%) & 120 (12.0\%)  \\
& DiffGAN & 285 (28.5\%) & 620 (62.0\%) & 95 (9.5\%)  \\ \hline

\multirow{2}{*}{M5, M6} 
& DRfuzz  & 267 (26.7\%) & 560 (56.0\%) & 173 (17.3\%)  \\
& DiffGAN & 198 (19.8\%) & 687 (68.7\%) & 115 (11.5\%)  \\ \hline

\multirow{2}{*}{M7, M8} 
& DRfuzz  & 303 (30.3\%) & 569 (56.9\%) & 128 (12.8\%)  \\
& DiffGAN & 298 (29.8\%) & 592 (59.2\%) & 110 (11.0\%)  \\ \hline
\end{tabular}
\caption* {Model A refers to the first model in each pair (e.g., M1, M3, etc.).}
\end{table*}

\subsection{\textbf{RQ4. Can training of an ML-based model selection mechanism be guided more effectively using the triggering inputs generated by \textit{DiffGAN} compared to those produced by \textit{DRfuzz}?}} % identify the specific conditions under which one model outperforms another?}} 

We aim in this question to study whether we can better guide the training of an ML-based model selection mechanism that consists of (1) two classifiers (model 1 and model 2) responsible for image classification, and (2) a third model that dynamically selects which of the two classifiers to use based on the input characteristics, which is based on Random Forest (RF). It predicts, based on the characteristics of the incoming input, which of the two classifiers (model 1 or model 2) will produce a more accurate output, without the need to run both models in an online setting.

When trained adequately with triggering inputs, the RF model should learn when each classifier is more accurate, effectively serving as a decision mechanism for selecting the optimal classifier for a given input. We choose RF because:
(1) it scales effectively to handle a large number of features, and (2) its robustness against overfitting~\cite{fernandez2014we, breiman2001random,friedman2001greedy}. We do not intend to provide a novel voting mechanism here, but instead, provide an example of an important application scenario for the triggering inputs we generate. It also serves as a basis to compare the value of the inputs generated by \textit{DiffGAN} to enable such model selection and compare it with that of \textit{DRfuzz}.

To achieve this, we trained the model selection system using two distinct training datasets. The first dataset consists of triggering inputs generated by \textit{DiffGAN}, while the second contains triggering inputs produced by \textit{DRfuzz}. To create these datasets, we leveraged the inputs generated by \textit{DiffGAN} and \textit{DRfuzz} in RQ1 (with an execution time budget of one hour), considering the four model pairs from the MNIST dataset. For each method and model pair, we randomly sampled 1,200 triggering images, resulting in a total of 9,600 images (1,200 images × 4 model pairs × 2 methods). Given the large volume of unlabeled triggering inputs, we outsourced the classification to a professional labeling company. Each input was labeled with a value from 0 to 9, with an additional label for ``\textit{invalid}'' inputs. Invalid images were subsequently filtered out from each of the 1,200-image sets. The number of inputs removed ranged from 21 to 121 for \textit{DRfuzz}, and from 3 to 34 for \textit{DiffGAN}.  It is important to note that we relied on a single labeling company due to cost and time constraints. This company was chosen for its strong reputation and track record of high-quality labeling. To further ensure the accuracy of the labels, we manually validated the labels of a random sample of 1,000 images. We found that all images were correctly labeled, confirming the reliability of the labeling process.

\Revv{ We also manually analyzed the types of triggering inputs generated by \textit{DiffGAN} and \textit{DRfuzz} by randomly sampling 1,000 labeled valid triggering images for each method and model pair. Each input was categorized into one of the following types: (1) only Model A produces the correct label, (2) only Model B produces the correct label, and (3) both models produce incorrect but different labels. Here, Model A refers to the first model in each pair, and Model B to the second.
This categorization enabled us to quantify the nature of disagreements revealed by the generated inputs. As shown in Table \ref{tab:model_behavior}, both \textit{DiffGAN} and \textit{DRfuzz} exhibit all disagreement types, however, in different proportions. Indeed, \textit{DiffGAN} generates significantly fewer inputs where both models are incorrect.   This is practically important as we want to generate as many inputs that distinguish the two models' behavior as possible, regarding conditions when they can be trusted, to learn which model to trust for a given input. In turn, this is useful to devise accurate model selectors, for example, at run-time. In conclusion, these results further explain why \textit{DiffGAN} yields better model selectors than \textit{DRfuzz}.}

%We found that the distributions are similar across both methods, with a noticeable tendency for Model B to produce the correct output more frequently. Note that, across all model pairs, \textit{DiffGAN} generates more cases where only one model is correct (Model B). However, there are still substantial instances where only Model A is correct or where both models are incorrect. This highlights the need for the ML-based model selection mechanism to select which model will produce more accurate results. 

To compare the accuracy of the RF-based model selection systems, we categorized the images into three groups based on whether one model predicted correctly or if both models failed. We excluded images where both models failed from further analysis due to their low occurrence and lack of relevance to the comparison (i.e., if both models fail, it does not matter which classification model is selected). We trained the  RF classifiers using features extracted from a pre-trained VGG16 model, rather than raw pixel data, in an attempt to improve accuracy. The VGG16 model provides feature vectors that capture more abstract and informative patterns than pixel-based features, which is expected to improve the RF model’s ability to distinguish between correct and incorrect predictions. For each method and model pair, we selected 700 valid labeled images and used their VGG16 features to train eight RF-based selection models. 
To ensure a fair comparison between the RF models trained on different datasets in each scenario, we need a common test dataset. For each scenario, we selected 100 images from the \textit{DiffGAN}-generated test set and 100 images from the \textit{DRfuzz}-generated test set, resulting in a combined test set of 200 images. This test set allowed us to evaluate the performance of the RF models using the same test set in an unbiased manner. Additionally, we applied the same feature extraction process as for the training sets.
% To maintain consistent evaluation, we applied the same feature extraction process to the test dataset. We created a test set of 200 images, evenly split between the two test generation methods, and used this set to evaluate the RF models' performance. 

\begin{figure}[ht]
\centering
\includegraphics[scale=0.4]{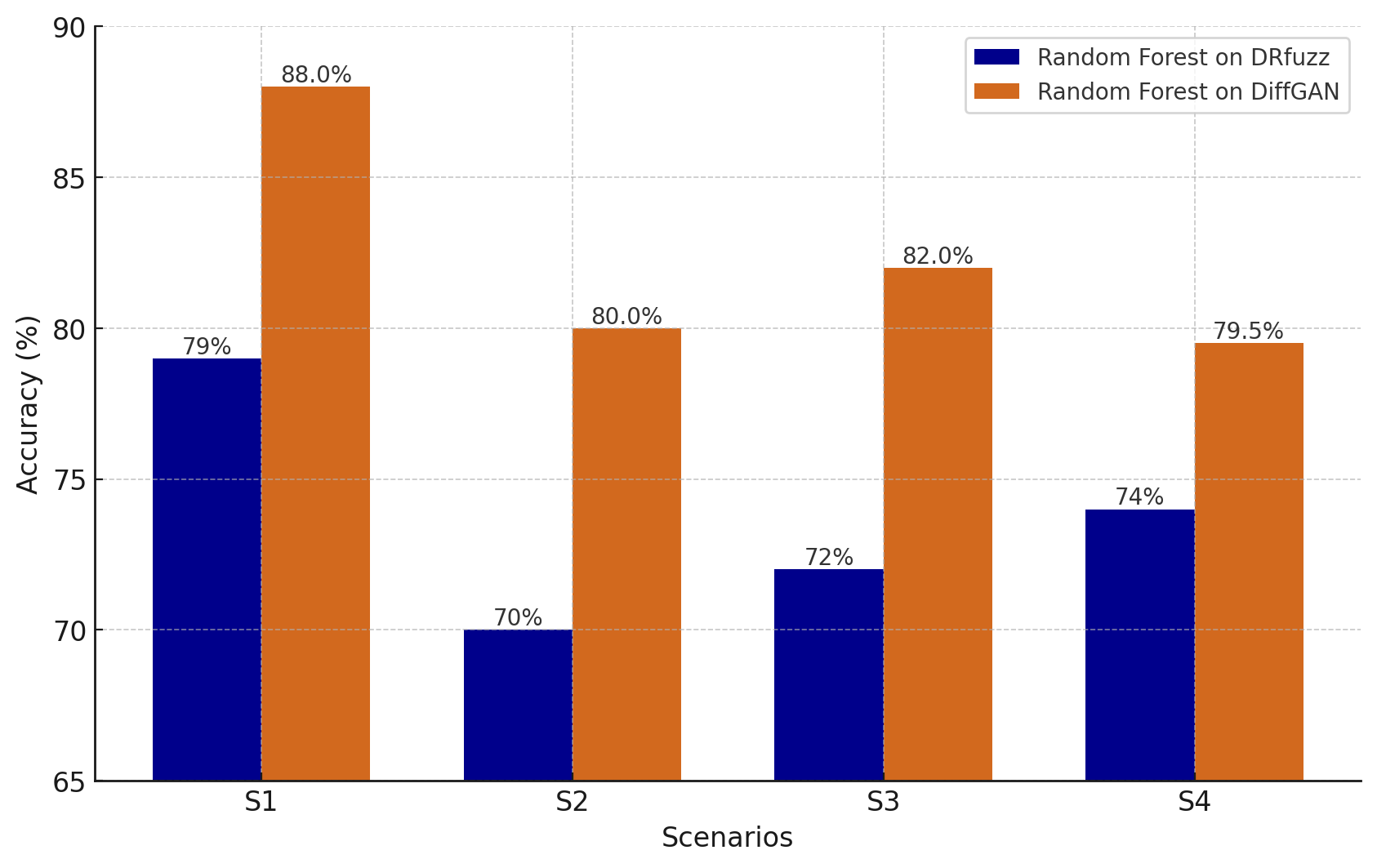}
\caption{Accuracy of Random Forest in Selecting the Best Model}
\label{fig:RQ4}
\end{figure}

We report the accuracy levels of each of the four selection model we trained in Figure~\ref{fig:RQ4}. As clearly visible, the RF-based selection models trained using triggering inputs generated with \textit{DiffGAN} are far more accurate than those trained with inputs from \textit{DRfuzz}. For instance, in the first scenario involving two RF-based selection models built using classifiers Model 1 and Model 2 from Table~\ref{tab:MNIST_Models}, the RF model trained with \textit{DRfuzz} achieved an accuracy of 78\%, whereas the one trained with \textit{DiffGAN} reached a much higher accuracy of 87\%.
Overall, the accuracy levels for the RF models trained with \textit{DiffGAN} range between 79\%  and 87\% compared to a range of 70\% to 78\% for those with \textit{DRfuzz} inputs. This highlights that \textit{DiffGAN} produces more effective triggering inputs for training, allowing the RF model to capture better the conditions under which one classifier outperforms the other. As a result, RF models trained using \textit{DiffGAN} inputs are more accurate in selecting the best-performing classifier, resulting in a more effective model selection mechanism.

% #RF trianed with diffgan, and numbers

This improved accuracy can be attributed to several factors. \textit{DiffGAN} leverages the combination of GAN and NSGA-II to generate highly diverse triggering inputs. Unlike \textit{DRfuzz}, which applies mutation rules to existing seed inputs, \textit{DiffGAN} explores a significantly broader latent input space, enabling the generation of more diverse inputs that expose more pronounced behavioral differences between the models. These behavioral differences are crucial for the RF model to learn the conditions when each classifier performs better. Additionally, \textit{DiffGAN} optimizes its inputs to both maximize diversity and reveal discrepancies, which also enables the RF model to develop a more accurate understanding of the conditions under which one classifier outperforms the other. Though the accuracy of RF models can still be improved, our main goal here was to show the usefulness of \textit{DiffGAN} for one important application, not to optimize model selection. The latter would require labeling a much larger dataset and possibly fine-tuning the RF models or experimenting with other machine learning techniques, which is outside the scope of this paper. 
 
%This highlight that when trained with triggering inputs generated with \textit{DiffGAN} the RF model is more capable to learn the conditions under which one classifier fares better leading to a more accurate model selection mechanism. 

\begin{tcolorbox}
\textbf{Answer to RQ4:} The ML-based model selection mechanisms trained with \textit{DiffGAN} triggering inputs are more accurate in selecting the best-performing classifier among alternative models, leading to better model selection mechanisms. This highlights that \textit{DiffGAN} produces more triggering inputs that are more effective for training, allowing the RF model to capture better the conditions under which one classifier outperforms the other. In turn, this can lead to more accurate voting mechanisms at inference time and thus higher classification accuracy.
\end{tcolorbox}

\section{\textbf{Discussions}}\label{Sec:Discussions}

\textbf{\textit{Relevance of our Approach:}} We present in this paper \textit{DiffGAN}, a test generation approach for the differential testing of DNN models for image classification. \textit{DiffGAN} leverages NSGA-II, a multi-objective optimization algorithm, to effectively explore the latent space of a GAN and generate diverse and valid triggering images. Unlike traditional mutation-based approaches that rely on modifying seed inputs~\cite{xie2019diffchaser,tian2023finding,you2023regression}, \textit{DiffGAN} directly explores the high-dimensional GAN latent space, allowing it to generate entirely new images with novel features that can trigger behavioral disagreements between the models under test. NSGA-II plays a crucial role by balancing two fitness functions: (1) maximizing divergence—ensuring, to generate inputs leading to significant differences in models outputs, and (2) maximizing diversity, thus reducing the generation of redundant images. Our ablation study shows that these two fitness functions are complementary and both essential for generating triggering images.  By integrating GAN-based test inputs generation with NSGA-II-driven optimization, \textit{DiffGAN} not only generates more triggering inputs than the SOTA approach but also ensures that these inputs are valid and more diverse.

\textbf{\textit{Practical Applicability:}} The black-box nature of \textit{DiffGAN} makes it agnostic to the models under test, as it does not require any assumptions on their architectural similarity. This flexibility allows \textit{DiffGAN} to be applied across diverse practical scenarios where models with similar accuracy levels are developed independently or trained using different data and optimization techniques.
The ability of \textit{DiffGAN} to generate diverse and valid triggering inputs makes it a practical approach for evaluating and comparing DNN models with similar accuracy levels. \textit{DiffGAN} is therefore particularly valuable for model selection, regression testing, and model compression, which are common application scenarios in ML development pipelines. In model selection, the triggering inputs generated by \textit{DiffGAN} can help identify behavioral discrepancies, enabling fine-grained analysis beyond accuracy metrics, ensuring the selection of the most adequate model. In regression testing, \textit{DiffGAN} can be used to generate cases where an updated model underperforms compared to its predecessor. In model compression validation, \textit{DiffGAN} can be used to understand when compressed models fare less well than their original versions and whether this is acceptable.
Another important application of \textit{DiffGAN} is ML-based model selection, a scenario informed by our industrial partner. In some autonomous vehicles, classifiers with similar accuracy levels are dynamically selected on the fly to classify sensor-captured images. Since different classifiers specialize in distinct regions of the feature space, selecting the most suitable one for each input improves reliability. Additionally, certain classifiers require more computational resources than others, making dynamic selection crucial for optimizing inference time and energy consumption, particularly in embedded systems with limited resources. By generating diverse inputs that expose model discrepancies, \textit{DiffGAN} enhances the accuracy of model selection mechanisms since their training dataset is enriched by such inputs and is more informative. The relevance of this application is further supported by existing literature on ML-based model selection~\cite{britto2014dynamic,cruz2018dynamic}.

%Generalisability
\textit{\textbf{Generalisability:}} Although \textit{DiffGAN} is primarily designed for image classification tasks, our approach can be extended to other domains such as image regression and segmentation. This extension can be achieved by adapting the divergence fitness function. For instance, in regression tasks, triggering inputs can be defined as those where the difference between the models’ outputs exceeds a predefined threshold. The divergence fitness function can then be tailored to capture a meaningful divergence metric, such as absolute or relative differences between regression model outputs, depending on the specific context. The primary objective would be to maximize the discrepancy between numerical outputs. For segmentation tasks, triggering inputs can be determined based on differences in predicted labels, variations in segmented shapes, or discrepancies in segmented regions. Once a triggering input is defined, the divergence fitness function can be customized to highlight these differences in the models’ outputs effectively. We should note that, for both regression and segmentation, the diversity-based fitness function will remain unchanged as long as the inputs are images. However, if the input type varies, the fitness function should be adapted accordingly to ensure an appropriate measure of diversity.
We should also note that \textit{DiffGAN} is not applicable to tasks involving multi-modal inputs or time-series data in its current form. This arises from the fact that our framework relies on GANs, which are specifically designed for image generation and manipulation. Supporting multi-modal or time-series tasks would require several modifications, such as integrating alternative generative architectures (e.g., Variational Autoencoders or Recurrent Neural Networks) and designing new strategies for defining and optimizing triggering inputs. While \textit{DiffGAN} is designed for image classification, its black-box differential testing approach can be extended to LLMs by adapting input generation using trained text-based models, employing embedding similarity or log-likelihood divergence as a fitness function, and integrating language model discriminators for validity filtering.
As future work, we aim to extend \textit{DiffGAN} to support a broader range of ML tasks beyond image classification. This includes adapting the fitness function to capture better model discrepancies and the diversity of the generated inputs. We also aim to explore how \textit{DiffGAN} can be adapted for other types of inputs, such as textual and audio inputs.
 
\Revv{Furthermore, our evaluation includes both highly accurate model pairs (with accuracy above 98\%) and relatively less accurate ones (ranging from 80\% to 97\%) to enhance the generalizability of our results. For highly accurate models, the generation of additional triggering inputs is considerably more challenging due to the scarcity of disagreement regions in the input space. For less accurate models, while behavioral disagreements may be more frequent, the key challenge remains in generating inputs that are not only valid but also diverse, to avoid trivial or uninformative disagreements. As shown in our ablation study (Table~\ref{tab:ablation_pairs_1_2}), maximizing only divergence without accounting for diversity leads to large numbers of triggering inputs, but many are similar, offering limited insight into the boundary behaviors of the models.
While the original test dataset contains triggering inputs for each model pair, these inputs could be redundant and typically do not sufficiently capture the diverse conditions under which behavioral differences between models occur. Relying solely on these inputs would not provide a comprehensive view of how and when models diverge in their predictions. \textit{DiffGAN} addresses these challenges by generating, on average, four times more triggering inputs than the SOTA baseline, while also significantly improving both the validity and the diversity of these inputs. This richer and more diverse set of valid triggering inputs enables a more comprehensive behavioral analysis of model pairs.}

%One of the key challenges of GANs is their propensity for mode collapse, which limits the diversity of the generated inputs. To mitigate this issue, our approach integrates multiple strategies during both the training phase and test input generation to improve the generation of diverse inputs.
%Specifically, during training, we ensured that GANs were properly optimized, achieving low FID and KID scores (Section~\ref{Sec:GAN_Traning}). These metrics, widely recognized for evaluating the quality and diversity of generated samples, were used to confirm that the trained GANs effectively captured the diversity of the underlying training data distribution~\cite{xu2018empirical}. Low FID and KID scores indicate that the GAN not only avoids mode collapse but also generates realistic and diverse samples. Another factor we used to improve the diversity of the data was generating an augmented training dataset to train the GAN by applying different image transformations (Section\ref{sec:GAN_Dataset}).  This data augmentation exposed the GAN to a more diversified training data, mitigating the risk of mode collapse.

%Mode Collapse

\Revv{\textit{\textbf{GAN Mode Collapse Mitigation:}} One of the key challenges of GANs is their propensity for mode collapse, which limits the diversity of the generated inputs. To mitigate this issue, our approach integrates multiple strategies during both the training phase and test input generation to promote diverse inputs. 
Specifically, to detect and monitor instability during training, we systematically tracked both the generator and discriminator loss curves throughout the training process. This allowed us to observe convergence behavior and identify any signs of instability, such as divergence or mode collapse. In addition, as detailed in Section~\ref{Sec:GAN_Traning}, we employed the Optuna framework to automate hyperparameter optimization. Optuna was used to tune learning rates, batch sizes, number of epochs, and other key model settings, to minimize FID and KID scores, metrics widely accepted for evaluating both the quality and diversity of generated outputs. We further enhanced the robustness of this process by averaging FID and KID scores across ten independent runs per configuration. Optuna’s early-stopping (pruner) mechanism helped discard unstable or underperforming configurations, contributing to a more stable and efficient optimization process.
These measures allowed us to identify configurations that led to more stable and better-quality training outcomes. The final trained models consistently achieved low FID and KID scores, which indicate strong convergence and the ability to capture the underlying data distribution without collapsing to limited modes (Section~\ref{Sec:GAN_Traning}).}

\Revv{We incorporated further strategies to explicitly enhance output diversity and reduce the likelihood of mode collapse. These included both modifications to the GAN training process and techniques applied during test input generation and evaluation:}

\noindent\Revv{\textbf{(1) Data augmentation during GAN training:} We applied a variety of image transformations to augment the original training dataset. This strategy exposed the GAN to a more heterogeneous input distribution, which is essential for preventing overfitting to narrow patterns and mitigating the risk of mode collapse. By training on this enriched dataset, the GAN learned to generalize better and produce outputs that reflect a wider range of valid data characteristics.}

\noindent\Revv{\textbf{(2) Diversity-guided test input generation using feature-based clustering:} During the generation of test inputs, we explicitly incorporated diversity as a key part of the search objective. By clustering generated images based on their feature representations, we ensured that newly generated samples were not only valid but also meaningfully different from previously generated ones. This clustering-based strategy prevented the GAN from generating repetitive or redundant outputs. It encouraged broader exploration of the latent space, thereby minimizing the risk of mode collapse and mitigating its potential consequences during input synthesis.}

\noindent\Revv{\textbf{(3) Quantitative diversity evaluation using two distinct metrics:} To assess the diversity of the outputs objectively, we used two separate diversity metrics and compared our method against multiple baselines. These metrics evaluated both the distributional spread and uniqueness of the generated samples. Our method consistently outperformed the baselines, confirming that our strategies led to more diverse and comprehensive coverage of the input space. This empirical evidence further supports the claim that our model avoided collapse to limited modes.}

\Revv{In combination, these strategies formed a multi-stage framework to avoid mode collapse and ensure high-quality, diverse outputs. Each component reinforced the others: enriched data improved variation, clustering promoted diverse generation, Optuna-based tuning with loss monitoring and early stopping ensured stability, and diversity metrics validated effectiveness.
This integrated process not only reduced the likelihood of mode collapse but also prevented its potential consequences by identifying and correcting instability early and ensuring that the diversity and realism of the generated outputs were preserved.}

\textbf{\textit{Efficiency.}} To evaluate the performance of \textit{DiffGAN} on large image datasets, we analyzed how GAN training time evolves with increasing dataset size. Specifically, we created larger-scale datasets by applying diverse image transformation techniques to the MNIST dataset, expanding the dataset size from 60,000 to 100,000 images. We then trained multiple GANs on these datasets, systematically increasing the number of training samples to assess \textit{DiffGAN}’s scalability. For each training run, we recorded training time and evaluated the quality of generated images using FID and KID scores.

\begin{figure}[]
    \centering
    \includegraphics[width=0.5 \textwidth]{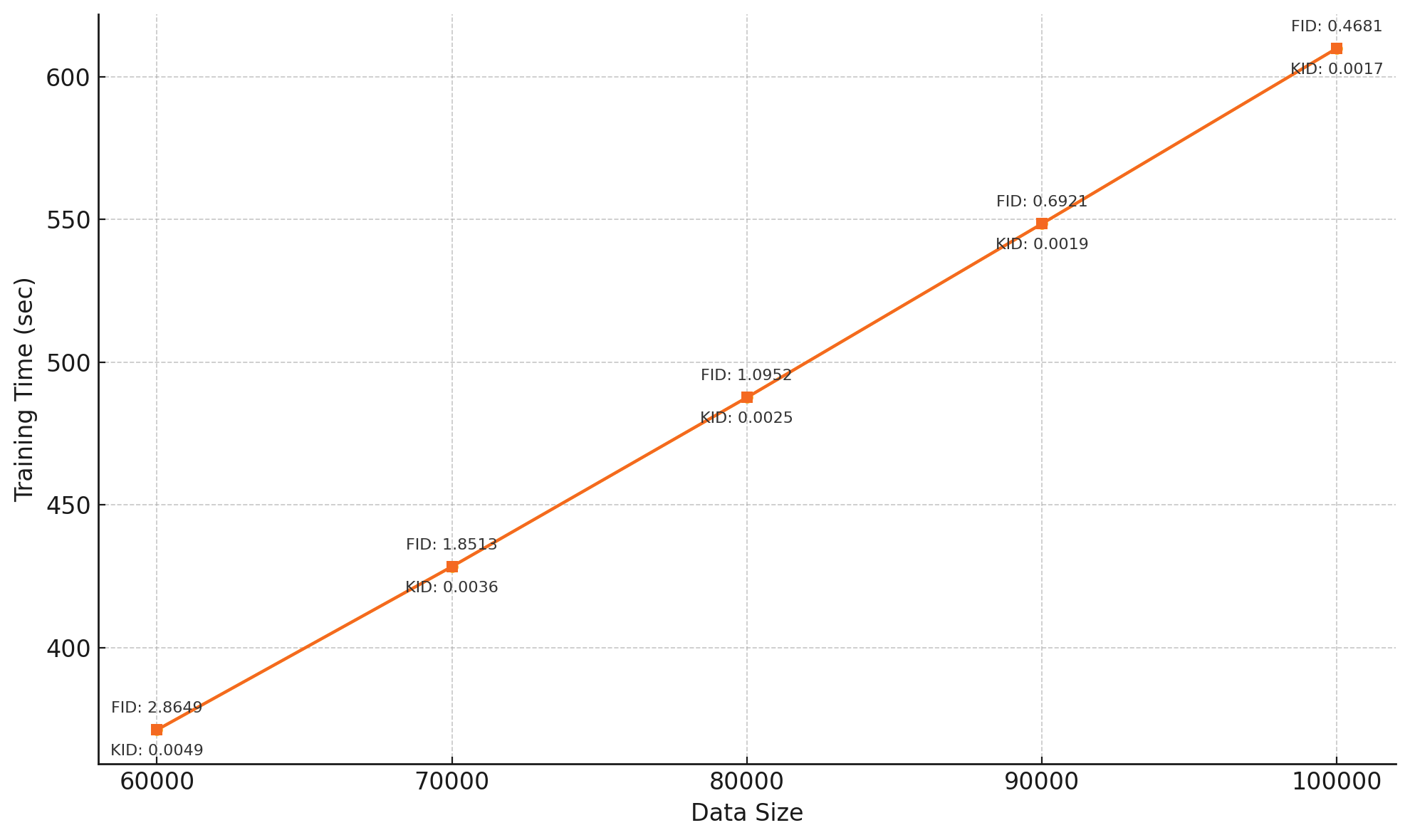}
    \caption{Relationship between dataset size and training time of Vanilla GAN.}
    \label{fig:fid_kid_vs_data}
\end{figure}

As shown in Figure~\ref{fig:fid_kid_vs_data}, we found that \textit{DiffGAN} scales linearly with increasing dataset size, with a training time ranging from 371 to 607 seconds. We note that this increase remains practically acceptable, as GAN training is a one-time process per dataset, and differential testing is neither frequent nor a real-time task. Thus, training costs do not significantly impact the feasibility of using \textit{DiffGAN} for large-scale differential testing.
%It is important to highlight that our study focused solely on GAN training time, as it is the most computationally expensive step in our approach. For example the training time of GAN with the MNIST dataset was NNN while it was NNN with Cifar-10.  
%We did not consider ImageNet, which is a large dataset, in our experiments to evaluate our approach  as we can use existing pretrained GANs~\cite{grigoryevand,zhao2020leveraging} on such type of datasets. 

\Revv{Additionally, it is important to clarify that \textit{DiffGAN} does not strictly require training a GAN from scratch, as it treats GANs (or any other generative model) as black-box components. This means that, for large-scale datasets such as ImageNet, \textit{DiffGAN} can leverage pretrained GANs or generators that have already been trained on such datasets~\cite{grigoryevand,zhao2020leveraging}. This would reduce the computational burden of training new GANs while maintaining the ability of \textit{DiffGAN} to generate diverse and valid test cases. We consider incorporating pretrained GANs and evaluating their effect on triggering input quality and computational savings as future work.}

%Additionally, it is important to clarify that \textit{DiffGAN} does not strictly require training a GAN from scratch, as it treats GANs (or any other generative model) as black-box components. This means that, for large-scale datasets such as ImageNet, \textit{DiffGAN} can leverage pretrained GANs or generators that have already been trained on such datasets~\cite{grigoryevand,zhao2020leveraging}, significantly reducing the computational burden while maintaining its ability to generate diverse and valid test cases. Existing GAN models trained on large datasets can be seamlessly integrated into \textit{DiffGAN} without requiring additional training, making our approach even more scalable and applicable beyond the datasets tested in our current study. 

\section{Threats to Validity} \label{sec:Threats}

In this section, we discuss the potential threats to the validity of our study and describe the measures we took to mitigate them.

\textbf{Internal threats to validity} concern the causal relationship between the treatment and the outcome. One such threat in our study is the inherent randomness in the experimental setup (RQ1-3), particularly the stochastic nature of \textit{DiffGAN} and \textit{DRfuzz}. To mitigate this, we executed each experiment three times, and we observed that the variance across these runs was minimal, leading to consistent conclusions in our RQs. As a result, we concluded that further repetitions, which would come at a high experimental cost, were unnecessary to draw reliable conclusions.

Another potential internal threat is related to the bias in the estimation of the validity of the generated inputs. To address this, we relied on three independent evaluators to assess the validity of the generated inputs manually. By requiring consensus among at least two of the three evaluators for determining validity, we minimized the influence of subjective bias, ensuring a reliable and consistent evaluation process. Additionally, the high level of agreement among evaluators, as measured by Cohen's Kappa, further confirms the reliability of the validity assessments.

\textbf{Construct threats to validity}  concern the relationship between the theory and the observations made. A potential threat to construct validity in our work could arise from the generation of invalid inputs by \textit{DiffGAN}. To mitigate this, we implemented a filtration process using two SOTA techniques: the SSIM score and the GAN's discriminator. We experimented with different SSIM score thresholds to find a near-optimal value for filtering out invalid images, manually reviewing all images in a set to verify the adequacy of this threshold. While we acknowledge that this threshold might potentially filter out a few valid images, it effectively removes the majority of invalid inputs. We also recognize that while this process significantly reduces the occurrence of invalid inputs, it does not entirely eliminate them. However, it minimizes their impact, enhancing the overall validity of our results.
Another potential threat relates to the construct validity of diversity measurements for the generated input sets. To mitigate this, we used two SOTA diversity metrics: Shannon Diversity and Geometric Diversity, each capturing different aspects of inputs' variability. The consistent results across both metrics further reinforce our confidence in our diversity assessment of the generated inputs.   

% \textbf{Construct threats to validity concern} the relation between the theory and the observations made. A construct threat to validity might be due to the generation of invalid inputs by \textit{DiffGAN}. To mitigate this, we applied a filtration process during input generation using two SOTA techniques: the SSIM score and the GAN's discriminator. We acknowledge that while this process significantly reduces the occurrence of invalid inputs, it does not entirely eliminate them. However, it minimizes their impact, enhancing the overall validity of our results.

%Generation of Invalid inputs --> use of SOTA Validity metrics to filter out invalid inputs

\Revv{\textbf{External threats to validity} concern the generalizability of our study. We considered in our experiments two datasets and eight pairs of DNN models, which may limit the generalizability of our study. While we recognize the potential of adding more datasets, such as ImageNet or Tiny ImageNet, incorporating them would require a substantial increase in manual effort to validate the generated images, which was already extensive. For instance, Tiny ImageNet involves 200 classes, making manual validation extremely challenging. Accurately classifying each image among so many visually similar categories is not only time-consuming but also difficult to perform reliably without expert annotators. This is precisely why we chose datasets where such validation, though still effort-intensive, remained feasible. 
To further mitigate the generalizability threat of our study, we included in our experiments 16 models across diverse datasets (MNIST and CIFAR-10), multiple domains, and various architectures (e.g., LeNet, ResNet, Inception), all of which are widely used in the literature. Furthermore, these combinations of models feature diverse model pairs varying in training data, optimizers, and hyperparameters.  Our evaluation includes both highly accurate model pairs and relatively less accurate ones to enhance the generalizability of our results. 
%We mitigate this threat by considering eight different combinations of widely used models and datasets. These combinations feature diverse model pairs varying in architecture, training data, optimizers, and hyperparameters. 
Additionally, we explored a wide range of testing budgets (execution times) in our experiments and compared our results against a SOTA baseline for the differential testing of DNN models.}
%\Rev{One potential external threat to the validity of our work is the generalizability of \textit{DiffGAN} to tasks beyond image classification, such as regression, multi-modal inputs, or time-series data. While \textit{DiffGAN} demonstrates strong performance in image classification tasks, its reliance on GANs, which are designed in this paper for image generation, limits its applicability to other domains without significant modifications. For instance, tasks involving multi-modal inputs (e.g., combining text and images) or time-series data would require alternative generative architectures and tailored fitness functions to define and optimize the generation of triggering inputs. This limitation suggests that the current framework may not directly translate to other deep learning tasks, and further research is needed to adapt \textit{DiffGAN} for broader applications. The inclusion of such tasks are therefore left as future work.}

\section{Related Work}
\label{Sec:RW}

% differential testing of DNN models can be categorized into black-box or white-box methods, depending on the level of access to the internal mechanisms of the models being tested. A few black-box approaches for DNNs have been introduced in the literature.

%Xie \textit{et al.}~\cite{xie2019diffchaser} proposed \textit{DiffChaser}, a search-based test generation approach designed to uncover behavioral disagreements between a DNN and its quantized version. This process begins with the selection of seed images, which are subsequently mutated and recombined iteratively. \textit{DiffChaser} leverages a k-Uncertainty fitness function~\cite{xie2019diffchaser} to select the fittest individuals that will be included in the generation loop aiming to find discrepancies between a model and its quantized version during the testing process.

Differential testing approaches for DNN models can be characterized as black-box or white-box, depending on their access requirements to the internals of the DNN models under test. 
A few black-box differential testing approaches for DNNs have been proposed in the literature. 

Xie \textit{et al.}~\cite{xie2019diffchaser} proposed \textit{DiffChaser}, a black-box, search-based test generation approach aiming to find behavioral disagreements between DNN models and their quantized versions. They initiate the generation process by selecting seed images. Through iterative mutations and crossovers of these seed images, \textit{DiffChaser} applies a k-Uncertainty~\cite{xie2019diffchaser} fitness function to guide the generation process, aiming to find discrepancies between a model and its quantized version during testing.

Tian \textit{et al.}~\cite{tian2023finding} proposed \textit{DFlare}, a search-based test generation approach that iteratively applies a series of mutation operations to a given seed image until a triggering input is found. They rely on \textit{Markov Chains} and a divergence-based fitness function to guide the selection of mutation operators~\cite{tian2023finding}. The fitness function is designed to prioritize mutated inputs that (1) maximize differences between the output probability vectors of two models or (2) trigger new, previously unobserved probability vectors in the models under test. 

You \textit{et al.}~\cite{you2023regression} introduced \textit{DRfuzz}, a differential testing method that detects regression faults in updated versions of DNN models. A regression fault occurs when an input correctly predicted by an earlier model version results in an incorrect prediction post-update. Similar to previous approaches, \textit{DRfuzz} uses seed inputs and mutation rules to generate new inputs that trigger regression faults. It employs a customized reward function to select mutation rules, favoring those that produce more fault-triggering inputs. \textit{DRfuzz} also leverages a GAN's discriminator to filter invalid inputs.

The above test generation approaches heavily rely on mutation operations applied to seed images, thus limiting the process to mutated versions of existing inputs. This dependence significantly narrows the range of test scenarios, as it prevents the introduction of entirely new features (i.e, features that are not present in the initial testing dataset) into the testing process. In contrast, \textit{DiffGAN} enhances its differential testing effectiveness by generating diverse and new test input scenarios.

%The above test generation approaches heavily depend on the availability and quality of seed data, limiting the process to mutated versions of existing inputs. This dependence significantly narrows the range of test scenarios, as it prevents the introduction of entirely new features (i.e, features that are not present in the initial testing dataset) into the testing process---a capability that \textit{DiffGAN} notably possesses, enhancing its effectiveness in differential testing by generating diverse and new test input scenarios.

A few white-box differential testing approaches have been proposed as well. 
\textit{DeepExplore}~\cite{pei2017deepxplore} is the first white-box differential testing approach for DNN models. It formulates the image generation problem as an optimization problem, which uses gradient-based search techniques to find images that maximize neuron coverage and the number of behavioural disagreements between the DNN models under test.
%\textit{DeepHunter} is another white-box test generation approach for DNNs that relies on mutation operators applied in seed inputs and coverage metrics to guide the generation of triggering inputs. The authors investigate NNN seed selection strategies and five coverage metrics to guide the generation process. The evaluation results on quantized DNNs revealed that there is no coverage metric that consistently 

\textit{DeepHunter}~\cite{xie2019deephunter} is a white-box test generation approach for DNNs that uses mutation operators on seed inputs along with coverage metrics to guide the generation of triggering inputs. The approach explores different seed selection strategies, incorporating both diversity-based and recency-based methods, to optimize the seed selection process. \textit{DeepHunter} supports five different coverage metrics to guide the generation process. However, evaluations on quantized DNNs indicate that no single coverage metric consistently outperforms the others across different quantized DNN models, in terms of generating more triggering inputs. 

\textit{Diverget}~\cite{yahmed2022diverget} and \textit{DeepEvolution}~\cite{braiek2019deepevolution} are two white-box differential testing approaches for quantized DNN models. They leverage multi-objective search algorithms and domain-specific mutation operators to generate triggering inputs. Two fitness functions guide the generation. The former fitness function maximizes the divergence in output probability vectors between the models under test, while the latter aims to maximize the coverage of neuron activations within those models. While \textit{DeepEvolution} is designed to test both original and quantized DNN models for image classification tasks, \textit{Diverget} specifically targets DNN models used for hyperspectral image classification~\cite{yahmed2022diverget}.

The use of white-box differential testing approaches is hindered by the required access to the internal structure of the DNN models. This can be a significant limitation in practice, particularly when the models under test are proprietary or provided by a third party~\cite{aghababaeyan2021black}. 
Additionally, several white-box approaches~\cite{yahmed2022diverget,braiek2019deepevolution} rely on comparing neuron activation values or model gradients of DNN models, thus assuming structural similarity among the models being tested. These requirements do not apply to \textit{DiffGAN}, which, unlike existing white-box differential testing approaches, supports testing across any pair of DNN models without requiring structural similarity or the need for internal access.

\Revv{ Several generative model-based approaches for test generation of DNNs have been proposed in the literature~\cite{dola2024cit4dnn,zhang2018deeproad,li2021testing}. For instance, Dola \textit{et al.}~\cite{dola2024cit4dnn} introduced a black-box GAN-based method aimed at generating test inputs that are both feature-diverse and include rare cases. Their approach leverages a generative model and applies combinatorial interaction testing~\cite{cohen1997aetg} to the model's latent space to enhance the diversity of the generated test inputs.  Also, Zhang \textit{et al.}~\cite{zhang2018deeproad} proposed DeepRoad, a GAN-based approach for testing DNN-based autonomous driving systems (ADS).  DeepRoad generates driving scenes with different weather conditions and applies metamorphic testing to detect mispredictions. Metamorphic relations are defined such that driving behavior remains consistent, regardless of how scenes are altered to simulate different weather conditions. Similarly, Li \textit{et al.}~\cite{li2021testing} proposed TACTIC, a test generation approach for DNN-based ADS. TACTIC leverages a GAN-based image-to-image translation framework to identify critical weather conditions that increase the likelihood of erroneous system behavior. 
While these approaches leverage various types of GANs to generate test inputs for DNNs, they focus on identifying mispredictions in individual DNN models. In contrast, our approach is specifically designed for differential testing between pairs of DNN models. We generate triggering inputs that expose behavioral disagreements between models, rather than targeting mispredictions in a single model.}

\section{Conclusion}
\label{Sec:Conclusion}
We presented \textit{DiffGAN}, a novel black-box test generation framework that leverages GANs and NSGA-II for differential testing of deep neural networks. \textit{DiffGAN} uses customized fitness functions in NSGA-II to guide the search process, generating diverse and valid triggering inputs that expose behavioral differences between models. To the best of our knowledge, this is the first work addressing black-box differential testing aimed at model selection. %Model selection is crucial because we frequently encounter different versions of models, which may come from various providers, be trained on different datasets, or be optimized with different hyperparameters. While these models often perform similarly on available test data, their behavior can vary significantly in operation. \textit{DiffGAN} helps systematically generate non-adversarial images that highlight the difference between pairs of models.
We show that \textit{DiffGAN} is capable of revealing behavioural differences across models by exploring large input spaces through GAN and NSGA-II’s diversity-driven fitness functions. %By identifying these differences, \textit{DiffGAN} enables dynamic model selection, offering a tool to decide which model should be deployed depending on the input features.
Our empirical evaluation on eight pairs of image classification models trained on two widely-used datasets demonstrated that \textit{DiffGAN} significantly outperforms the SOTA differential testing baseline, \textit{DRfuzz}, by generating on average four times more triggering inputs. These inputs are not only more diverse but also more likely to be valid, allowing for a deeper analysis of the models' behavior. Furthermore, we showed how these inputs can be applied in a machine learning-based voting mechanism, which dynamically selects the most accurate model based on input characteristics, resulting in improved classification accuracy compared to the baseline approach.
%Our experiments on image classification models trained on two SOTA image classification datasets demonstrate that \textit{DiffGAN} significantly outperforms \textit{DRfuzz}, the SOTA baseline. \textit{DiffGAN} generates a larger number of diverse and valid triggering inputs, which reveal nuanced performance differences that are critical for ensuring robust decision-making. Additionally, by using the triggering inputs generated by \textit{DiffGAN}, machine learning-based model selection becomes more accurate, allowing for informed decisions without the need for real-time evaluation of multiple models. \textit{DiffGAN}’s robustness across multiple runs underscores its reliability in diverse conditions.

For future work, we aim to enhance the generalizability of \textit{DiffGAN} by extending it to support a wider range of models, machine learning tasks, and input types. We also aim to explore \textit{DiffGAN} in dynamic environments to assess the side effects of updating a single ML component on the overall system performance. Rather than comparing pairs of DNN models, we will compare the original system with the updated version. This approach will provide valuable insights into maintaining system reliability and adaptability after updates, particularly in critical areas like autonomous driving and healthcare. 

\section*{Acknowledgments }
\label{Sec:ack}
This research was supported by a grant from General Motors, as well as the Canada Research Chair and Discovery Grant programs of the Natural Sciences and Engineering Research Council of Canada (NSERC). Lionel Briand’s contribution was partially funded by the Science Foundation Ireland grant 13/RC/2094-2. We would also like to thank Amirhossein Zolfagharian for his valuable assistance in evaluating the validity of the generated inputs in our experiments.

\bibliographystyle{IEEEtran}
\bibliography{main.bib} 

\newpage

\twocolumn[\section*{Appendix I }]

\section*{Diversity Comparison}

We evaluated the input diversity generated by \textit{DFlare}, \textit{DRfuzz}, and our approach on the MNIST dataset using four model pairs. Following the methodology of RQ3, we employed Geometric Diversity and Shannon Entropy as metrics, with a fixed generation time of one hour for each method.
As shown in Table~\ref{Tab:Div_dflare}, the results indicate that \textit{DFlare} produced highly redundant inputs compared to both \textit{DRfuzz} and \textit{DiffGAN}, consistently across all model pairs. 
Due to its poor performance in generating diverse inputs, we excluded \textit{DFlare} from further analysis.

% \begin{table}[h]
% \centering
% \color{blue}
% \caption{Diversity Scores of MNIST (1h Configuration)}
% \begin{tabular}{|c|c|r|c|}
% \hline
% \textbf{Models} & \textbf{Method} & \textbf{GD} & \textbf{Shannon Entropy} \\
% \hline
% M1, M2 & DFlare   & -377.52 & 4.7435 \\
%        & DRFuzz   & 422.03  & 11.36 ($\sim$86,361.86) \\
%        & DiffGAN  & \textbf{530.62}  & \textbf{12.10} ($\sim$178,632.90) \\
% \hline
% M3, M4 & DFlare   & 132.15  & 4.7468 \\
%        & DRFuzz   & 438.42  & 11.39 ($\sim$89,882.98) \\
%        & DiffGAN  & \textbf{727.20}  & \textbf{12.21} ($\sim$200,195.46) \\
% \hline
% M5, M6 & DFlare   & -72.37  & 4.7666 \\
%        & DRFuzz   & 410.67  & 11.50 ($\sim$98,610.81) \\
%        & DiffGAN  & \textbf{712.76}  & \textbf{12.27} ($\sim$213,402.74) \\
% \hline
% M7, M8 & DFlare   & 36.42   & 4.7855 \\
%        & DRFuzz   & 116.72  & 11.40 ($\sim$90,159.89) \\
%        & DiffGAN  & \textbf{481.94}  & \textbf{12.18} ($\sim$178,751.64) \\
% \hline
% \end{tabular}
% \label{Tab:Div_dflare}
% \end{table}

\begin{table}[h]

\centering
\captionsetup{justification=centering}
\caption[]{\centering Diversity scores of MNIST (1h Configuration)\par}
\label{Tab:Div_dflare}
\resizebox{\textwidth}{!}{%
\begin{tabular}{|c|c|cc|cc|cc|}
\hline
\textbf{}               & \textbf{} & \multicolumn{2}{c|}{\textbf{DFlare}} & \multicolumn{2}{c|}{\textbf{DRfuzz}}     & \multicolumn{2}{c|}{\textbf{DiffGAN}}                      \\ \hline
Models                  & Configs   & \multicolumn{1}{c|}{GD} & Exponential Shannon ($e^{H}$) & \multicolumn{1}{c|}{GD}       & Exponential Shannon ($e^{H}$)     & \multicolumn{1}{c|}{GD}                & Exponential Shannon ($e^{H}$)            \\ \hline
\multirow{1}{*}{M1, M2} & 1h        & \multicolumn{1}{c|}{-377.52} & $e^{4.74} \approx 114.88$   & \multicolumn{1}{c|}{422.03}   & $e^{11.36} \approx 86,361.86$     & \multicolumn{1}{c|}{\textbf{530.62}}   & \textbf{$e^{12.10} \approx 178,632.90$} \\ \hline
\multirow{1}{*}{M3, M4} & 1h        & \multicolumn{1}{c|}{132.15}  & $e^{4.75} \approx 115.07$   & \multicolumn{1}{c|}{438.42}   & $e^{11.39} \approx 89,882.98$     & \multicolumn{1}{c|}{\textbf{727.20}}   & \textbf{$e^{12.21} \approx 200,195.46$} \\ \hline
\multirow{1}{*}{M5, M6} & 1h        & \multicolumn{1}{c|}{-72.37}  & $e^{4.77} \approx 117.88$   & \multicolumn{1}{c|}{410.67}   & $e^{11.50} \approx 98,610.81$     & \multicolumn{1}{c|}{\textbf{712.76}}   & \textbf{$e^{12.27} \approx 213,402.74$} \\ \hline
\multirow{1}{*}{M7, M8} & 1h        & \multicolumn{1}{c|}{36.42}   & $e^{4.79} \approx 120.37$   & \multicolumn{1}{c|}{116.72}   & $e^{11.40} \approx 90,159.89$     & \multicolumn{1}{c|}{\textbf{481.94}}   & \textbf{$e^{12.18} \approx 178,751.64$} \\ \hline
\end{tabular}%
}
\end{table}

\end{document}